\documentclass[pdflatex,sn-mathphys-num]{sn-jnl}

\usepackage{graphicx}%
\usepackage{multirow}%
\usepackage{subcaption}
\usepackage{caption}
\usepackage{amsmath,amssymb,amsfonts}%
\usepackage{hyperref}
\usepackage{cleveref}

\usepackage{amsthm}%
\usepackage{mathrsfs}%
\usepackage[title]{appendix}%
\usepackage{lmodern}
\usepackage{xcolor}%
\usepackage{textcomp}%
\usepackage{manyfoot}%
\usepackage{booktabs}%
\usepackage{algorithm}%
\usepackage{algorithmicx}%
\usepackage{algpseudocode}%
\usepackage{listings}%
\usepackage[left]{lineno}
\usepackage{xcolor}
\usepackage{comment}

\theoremstyle{thmstyleone}%
\theoremstyle{thmstyletwo}%

\theoremstyle{thmstylethree}%

\begin{document}


\title[Title]{Joint Utilization of Geospatial and census proxies for Autoencoder-Assisted Downscaling (JUGAAD) of socioeconomic indicators in India}


\author*[1]{\fnm{Aditya} \sur{Dutt}}\email{aditya.dutt@ufl.edu}

\author[2]{\fnm{Paul} \sur{Gader}}\email{pgader@ufl.edu}

\author[3]{\fnm{Aditya} \sur{Singh}}\email{aditya.singh@tamuk.edu}

\affil*[1]{\orgdiv{Department of Electrical and Computer Engineering}, \orgname{University of Florida}, \orgaddress{ \city{Gainesville}, \postcode{32603}, \state{Florida}, \country{USA}}}

\affil[2]{\orgdiv{Department of Engineering School of Sustainable Infrastructure \& Environment}, \orgname{University of Florida}, \orgaddress{ \city{Gainesville}, \postcode{32603}, \state{Florida}, \country{USA}}}

\affil[3]{\orgdiv{Caesar Kleberg Wildlife Research Institute}, \orgname{Texas A\&M University}, \orgaddress{ \city{Kingsville}, \postcode{78363}, \state{Texas}, \country{USA}}}

\abstract{Monitoring poverty and food security indicators is imperative for addressing socioeconomic challenges in developing nations. A limitation is mismatches in scale between data sources: census data provide geographic coverage, while socioeconomic indicators are derived from infrequently conducted surveys at coarse resolutions, posing a methodological challenge. This study introduces a deep learning framework, JuGAAD, using Indian census and survey data from 2001 and 2011 as a case study. We employ a three-step process: census and geospatial data are averaged into intermediate village-cluster-scale tessellations to reduce noise and regularize administrative boundary changes; an autoencoder compresses high-dimensional National Sample Survey Office (NSSO) data into a low-dimensional latent representation; and a regression model maps upscaled census and geospatial data to this representation. This function is applied to fine-grained census data to generate high-resolution predictions, validated against ground-truth district-level NSSO indicators. Results confirm the methodology predicts socioeconomic indicators at fine scales with strong accuracy.}


\maketitle

\section{Introduction}
\label{sec:Introduction}
Detailed, unit-level demographic and socioeconomic data are the bedrock of effective, evidence-based policymaking. They allow governments to accurately assess population structures, monitor urbanization, and track migration patterns, all essential steps for designing targeted interventions in sectors like healthcare, education, and infrastructure. In developing economies, these data are crucial for evaluating economic stability and pinpointing disparities in income, employment, and resource access, which in turn inform poverty reduction strategies and resource allocation. Furthermore, they are vital for measuring progress toward international targets, such as the Sustainable Development Goals (SDGs), nearly half of which rely on such indicators.\par

Despite their fundamental importance, high-resolution socioeconomic data are frequently unavailable, costly to acquire, or confined to infrequent household surveys. This results in a critical gap between data granularity and policy requirements. For instance, in India, the National Sample Survey Office (NSSO) provides socioeconomic metrics at the district level, a resolution insufficient for localized policy action. \par

Conventional efforts to address this scarcity have centered on Small Area Estimation (SAE) techniques. Prominent SAE methods, including the Elbers, Lanjouw, and Lanjouw (ELL) approach \cite{das2015robust}, Empirical Bayes Prediction (EBP) \cite{molina_small_2010}, and M-quantile (MQ) modeling \cite{chambers_m-quantile_2006}, have become standard tools for integrating survey and census data to generate localized estimates. While each method offers distinct benefits—such as the parametric simplicity of ELL, the non-linearity handling of EBP, and the outlier robustness of MQ—they are uniformly constrained by a reliance on the assumption of spatial homogeneity. This foundational premise posits that relationships between variables are stable across different geographic areas. This assumption often compromises predictive power in developing regions, which are frequently characterized by the very spatial heterogeneity that these models fail to capture.\par

Recognizing the limitations of conventional statistical models in capturing spatial complexity, the research community has increasingly shifted toward remote sensing and machine learning as scalable, data-driven alternatives. Among the earliest and most impactful geospatial indicators is night-time lights (NTL) data, which became publicly available in the 1990s. The underlying principle is that artificial illumination serves as a strong, quantifiable proxy for economic activity \cite{huang2021saturated}. Studies have confirmed robust correlations between NTL intensity and various economic metrics, including wage income in Sweden \cite{mellander2015night} and asset-based wealth across numerous African nations \cite{noor2008using, andersson2019data}. \par

However, a core limitation of NTL-based methods is their difficulty in distinguishing between economically distinct areas with similar luminosity (\textit{e.g.}, densely populated low-income versus sparsely populated high-income regions) \cite{jean2016combining}. This critical shortcoming necessitated an evolution: researchers began augmenting NTL data with high-resolution daytime imagery. The incorporation of granular daytime features, such as building roofing materials and infrastructure, provides the necessary detail for more robust and accurate poverty prediction frameworks \cite{jean2016combining}. To extract meaningful spatial representations from high-resolution satellite imagery, Convolutional Neural Networks (CNNs) serve as the methodological standard \cite{lecun2002gradient, daoud2023using}. A significant obstacle, however, is the scarcity of labeled satellite data, which is particularly acute for the data-hungry deep learning architectures required to achieve generalizable features and high predictive performance.\par

The prevailing strategy to mitigate data scarcity is transfer learning. This paradigm adapts knowledge from models pre-trained on massive, general-purpose image datasets, such as ImageNet, to a more specialized domain. By fine-tuning these powerful, learned feature hierarchies, effective models can be developed even with limited task-specific data. Features from networks like Overfeat \cite{sermanet2013overfeat}, for example, have demonstrated efficacy in generalizing to diverse downstream tasks \cite{sharif2014cnn} and have been applied to identify development proxies (\textit{e.g.}, built structures, vegetation) in satellite imagery from countries including Haiti, Cambodia, and India. \par

A persistent challenge is the domain mismatch between the object-centric, ground-level perspective of ImageNet and the top-down, aerial perspective of satellite imagery. To bridge this gap, a pivotal innovation was the intermediate fine-tuning step proposed by Xie \textit{et al.} \cite{xie2016transfer}. Their multi-stage approach first adapts the ImageNet-trained model to the aerial domain by training it to predict NTL intensity from daytime images—a well-established proxy for economic activity. This intermediate task compels the network to learn spatially relevant representations from an aerial viewpoint. Methodologically, this is often implemented using a fully convolutional architecture to preserve image detail. While powerful, a purely spatial analysis provides only a static snapshot; consequently, recent work has underscored the importance of incorporating the temporal dimension to capture the dynamics of development \cite{pettersson2023time, kakooei2024increasing}.\par

A parallel research stream seeks to augment remote sensing data with other auxiliary sources, such as human mobility patterns derived from mobile phone metadata and environmental conditions from meteorological records \cite{engstrom2015mapping, puttanapong_predicting_2020, xu_poverty_2021}. These approaches often share a deep learning foundation with imagery-based techniques, exploiting spatial patterns in urbanization and land use \cite{tingzon_mapping_2019, ayush2021efficient, xie_transfer_2016}. However, these novel data streams face critical limitations, including uneven spatial coverage, reliance on proprietary data, and significant privacy concerns.\par

In policy-relevant domains, a critical consideration is model interpretability. Deep neural networks, while excelling at modeling complex, high-dimensional data, often function as ``black boxes," obscuring the decision-making process \cite{rudin_stop_2019, zhang_survey_2021}. This presents a fundamental trade-off: traditional transparent models like linear regression lack the capacity to capture the non-linear relationships inherent in geospatial data. Research is actively focused on developing inherently interpretable models for use with satellite data to reconcile the tension between predictive power and explainability \cite{ayush_generating_2020, abitbol2020interpretable, ledesma2020interpretable}. \par

In addition, another foundational challenge is the selection of an appropriate poverty metric, as poverty is inherently multidimensional \cite{alkire2015multidimensional, d2024multidimensional}. Metrics can vary by measurement scale: Absolute (\textit{e.g.}, World Bank's international poverty lines) vs. relative (benchmarked against income distribution) \cite{decerf2022unambiguous}; be of a temporal nature: Transient vs. Persistent \cite{chung2020understanding, dang2019poverty}, or may vary by the unit of Analysis: Individual vs. Place-based deprivation \cite{abascal2022domains}. For geospatial mapping, persistent, place-based poverty is particularly salient, as it identifies regions with structural disadvantages that can be targeted for intervention \cite{hall2022review}. Approaches that eliminate predefining any poverty metric may therefore be attractive simply due to the flexibility in use for downstream users.\par

In India, a significant body of research has applied machine learning frameworks to integrate traditional survey data, such as the National Family Health Survey (NFHS), with a diverse range of geospatial indicators: census records, satellite daytime imagery, NTL, climate data, and points of interest \cite{daoud2021measuring, mehta2025predicting}. This multi-source strategy aims to overcome the limitations of survey data, which, despite their detail, can be biased or statistically insufficient in underrepresented regions (\textit{e.g.}, NSSO samples in some districts contain fewer than 30 observations \cite{mehta2025predicting}).\par

While a consensus is emerging that integrating heterogeneous data sources is crucial for robust poverty estimation in India \cite{arya2025integrating, khare2024eyes}, significant methodological challenges persist.
Firstly, most existing models generate coarse predictions at the state or district level, leaving high-resolution mapping unresolved \cite{subash2018satellite, mehta2025predicting}. Secondly, socioeconomic indicators are typically high-dimensional and exhibit strong multicollinearity. And finally, data misalignment can occur when administrative boundaries (villages, \textit{taluks}, districts) are moved without generating corresponding lookup tables to reconcile databases across space and time. Conventional dimensionality reduction techniques, such as Principal Component Analysis (PCA) \cite{wold_principal_1987}, assume linearity and often fail to capture the complex, nonlinear patterns in socioeconomic data. A clear need exists for a framework that can inherently model non-linear relationships within high-dimensional data to produce granular, high-resolution predictions.\par

We propose \textit{JuGAAD}, a deep learning framework designed to directly address the challenges of high spatial resolution and data complexity in socioeconomic mapping, particularly for downscaling raw socioeconomic indicators. \textit{'JuGAAD'} also happens to be a colloquial Hindi term for finding a novel solution to a complex problem. Our core methodology leverages the complementary strengths of two key datasets: 1) High-Resolution Inputs: geographically comprehensive but thematically sparse village-level Indian Census data augmented with geospatial environmental variables, and 2) Target rich data: Detailed NSSO indicators ($\sim$450, including debt, employment, housing condition, etc.) available only at a coarse quasi-district-level resolution.\par

We first establish a temporally consistent spatial reference scheme based on a hexagonal tessellation (sized by clusters of $\sim$20 villages) that allows data to be tracked across years. We then generate a downscaling function that maps the high-resolution input features to the detailed NSSO indicators. By applying this learned relationship at the sub-district level, \textit{JuGAAD} generates fine-scale predictions of critical socioeconomic indicators where they were previously unavailable. The result is a set of high-resolution, temporally aligned maps that enable analysis of decadal changes in poverty and food security.

To do this, we first employ an autoencoder to compress high-dimensional and multicollinear NSSO data into a dense, low-dimensional latent representation. As an unsupervised neural network, the autoencoder learns to efficiently encode and reconstruct the input, preserving essential patterns while removing statistical redundancy. Unlike linear methods such as PCA, autoencoders excel at modeling the complex, non-linear relationships characteristic of socioeconomic data \cite{zhou_anomaly_2017, chung_audio_2016, pu_variational_2016}. With the compressed representation established, the second stage involves training a predictive regression model to estimate these latent features using readily available high-resolution data. The input feature set comprises fine-grained census indicators and geographical variables. Recognizing that administrative heterogeneity (policy and governance) can lead to divergent poverty levels across similar geographies, we augment this feature set with state-level identifiers. A neural network is then trained to map this comprehensive set of high-resolution inputs to a low-dimensional latent representation of NSSO indicators. This regression model forms the core of our downscaling mechanism, enabling predictions at a granular level. During the inference phase, the trained regression model is deployed using fine-grained census and geographic data to produce sub-district-level socioeconomic estimates. To rigorously validate the framework, high-resolution predictions are aggregated back to the original district level and then compared against actual district level NSSO indicators. Experimental results confirm that this framework successfully downscales socioeconomic indicators with high fidelity, providing a scalable, data-efficient solution to generate granular socioeconomic data in regions where it is otherwise unavailable.

\section{Results}
\label{sec:Results}

This section presents the performance evaluation and analysis of the proposed \textit{JuGAAD} framework. We first provide a quantitative assessment of the downscaled socioeconomic estimates to evaluate downscaled estimates of NSSO indicators. Next, spatial visualizations of the predicted maps are examined to illustrate how the model captures fine-scale variation across regions. Finally, we analyze the resulting high-resolution socioeconomic indicators with a focus on poverty-related measures to demonstrate the framework’s interpretability and policy relevance.

\subsection{Quantitative Evaluation of Downscaling Results}

Following the preprocessing steps, the final analytical dataset derived from the NSSO surveys comprises a total of 475 features distributed across six socioeconomic categories. The feature counts for each category are as follows: Consumer Expenditure (110), Employment (92), Agriculture (76), Housing Conditions (89), Land and Livestock Holdings (45), and Debt and Investment (83). For the downscaling task, each of these categories was encoded into a lower-dimensional latent space. Specifically, the Agriculture and Land and Livestock Holdings categories were mapped to 16-dimensional latent vectors, while the remaining four categories were each mapped to a 20-dimensional latent vector. This results in a consolidated latent representation with a total dimensionality of 112.\par

To validate the dimensionality reduction process, a separate autoencoder was trained for each socioeconomic category using the combined district-level data from 2001 and 2011. The performance of these autoencoders was evaluated based on their ability to reconstruct the original high-dimensional features from the compressed latent representations. Table~\ref{table:ae_metrics} summarizes the reconstruction accuracy, reporting the Mean Squared Error (MSE), Uniform $R^2$, and Weighted $R^2$ for each category.

The trained autoencoders demonstrated strong performance in reconstructing the original data from the compressed latent spaces, with most categories achieving a Weighted $R^2$ value exceeding $0.85$. The Land and Livestock Holdings and Housing Conditions categories yielded the highest reconstruction fidelity (Weighted $R^2$=0.89 and $0.87$, respectively), suggesting a relatively stable and well-defined feature structure. The Consumer Expenditure and Agriculture categories also showed robust performance (Weighted $R^2\approx0.87$). In particular, the model effectively captured the structure of the Agriculture data despite its higher intrinsic variation between districts. The slightly lower reconstruction accuracy for the Debt and Investment, and Employment categories is likely attributable to greater heterogeneity in their constituent variables.

To contextualize these results, we performed a comparative analysis against Principal Component Analysis (PCA), a standard linear dimensionality reduction technique. For a direct comparison, PCA was applied to reduce the dimensionality of each category to match that of its autoencoder counterpart. As detailed in Table~\ref{table:ae_metrics}, the autoencoder consistently outperformed PCA, achieving higher $R^2$ values and lower Mean Squared Error across all categories. This result underscores the advantage of employing a nonlinear model to capture the complex relationships within the socioeconomic data, a departure from prior methodologies that have often relied on PCA.

\begin{figure}[!htb]
        \centering
    
        \begin{subfigure}[!htb]{0.25\linewidth}
            \centering
            \includegraphics[width=\linewidth]{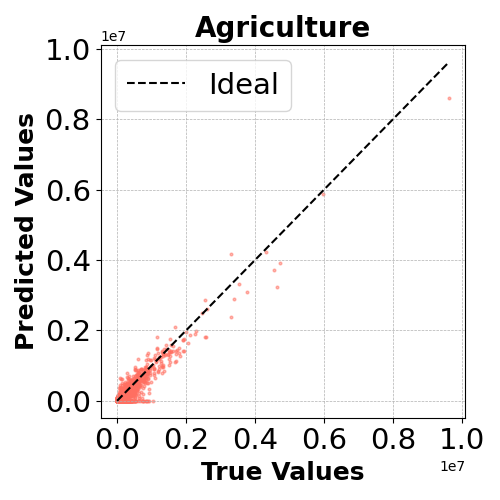}
            \captionsetup{justification=centering}
            \caption{}
        \end{subfigure}
        \begin{subfigure}[!htb]{0.25\linewidth}
            \centering
            \includegraphics[width=\linewidth]{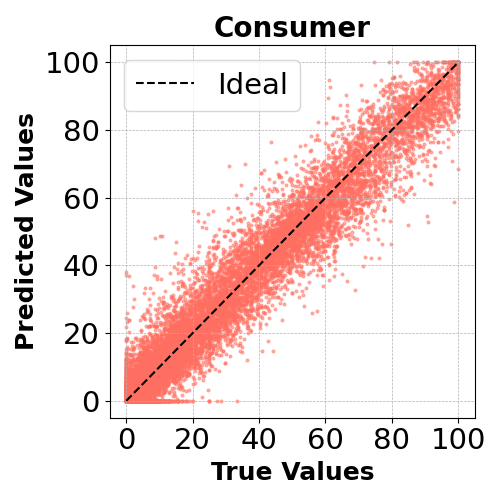}
            \captionsetup{justification=centering}
            \caption{}
        \end{subfigure} 
        \begin{subfigure}[!htb]{0.25\linewidth}
            \centering
            \includegraphics[width=\linewidth]{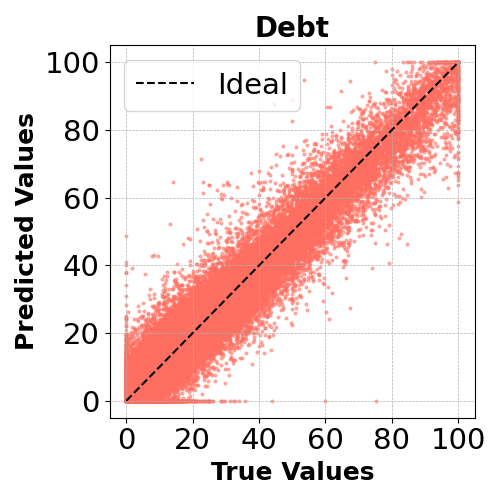}
            \captionsetup{justification=centering}
            \caption{}
        \end{subfigure} 
        \begin{subfigure}[!htb]{0.25\linewidth}
            \centering
            \includegraphics[width=\linewidth]{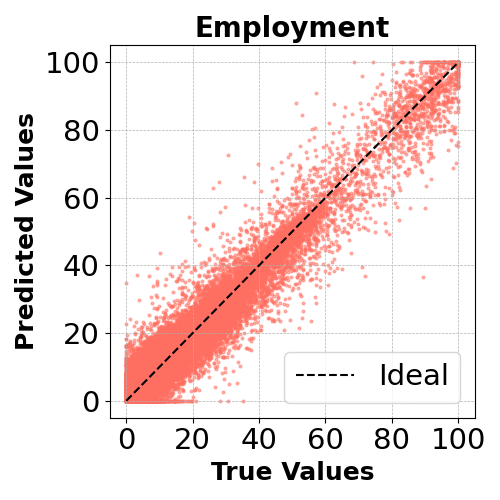}
            \captionsetup{justification=centering}
            \caption{}
        \end{subfigure}
        \begin{subfigure}[!htb]{0.25\linewidth}
            \centering
            \includegraphics[width=\linewidth]{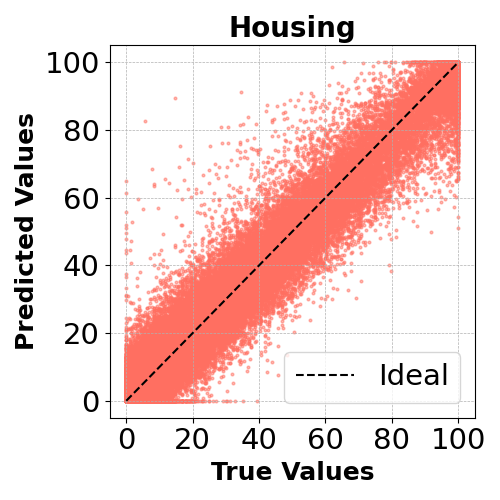}
            \captionsetup{justification=centering}
            \caption{}
        \end{subfigure} 
        \begin{subfigure}[!htb]{0.25\linewidth}
            \centering
            \includegraphics[width=\linewidth]{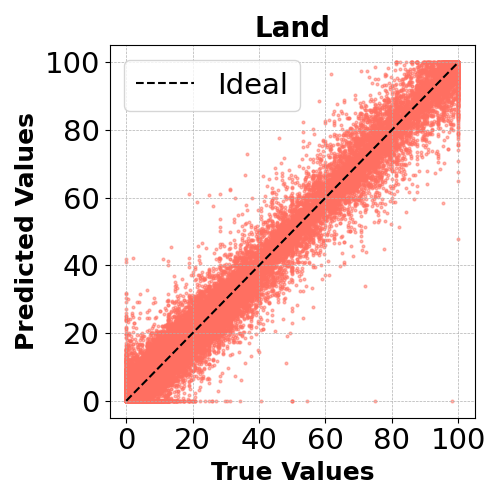}
            \captionsetup{justification=centering}
            \caption{}
        \end{subfigure} 

        \caption[]{Feature Reconstruction Fidelity of the Autoencoder. Autoencoder reconstruction scatter plots on training data compare the true values of the high-dimensional NSSO input features (x-axis) against the values reconstructed from the compressed latent space (y-axis). Plots correspond to (a) Agriculture, (b) Consumer, (c) Debt, (d) Employment, (e) Housing, and (f) Land categories, using the combined $2001$ and $2011$ NSSO datasets. The tight clustering of points along the diagonal line confirms the high fidelity of the autoencoder in capturing the non-linear structure of the data.}
        \label{fig:ae_recon_scatter}
    
    \end{figure}

While the category-level metrics indicate strong overall performance, a more granular, feature-level analysis reveals important variations. The scatter plots presented in Fig. ~\ref{fig:ae_recon_scatter}, which compare original and reconstructed values for individual variables, illustrate this point. Although the majority of features within each category were reconstructed with high fidelity, certain indicators, particularly within the Debt and Investment and Employment categories, exhibit greater variance and reconstruction error. This finding suggests that while the category-specific autoencoders effectively capture the dominant modes of variation, the reconstruction quality for specific, more heterogeneous indicators can vary at the individual level. \par

\begin{table*}[!htb]
    \centering
    \caption{Performance comparison between autoencoder (non-linear) and PCA (linear) for dimensionality reduction of NSSO data. The table compares the fidelity of the reconstructed district-level NSSO data after compression into a low-dimensional latent space. Metrics are presented for Mean Squared Error (MSE), Uniform $R^{2}$, and Weighted $R^{2}$, combining data from 2001 and 2011. The consistently lower MSE and higher $R^{2}$ values validate the autoencoder's capacity to reconstruct the complex, non-linear structure of the socioeconomic data more effectively than the linear PCA.}

    \footnotesize
    \resizebox{\linewidth}{!}{    

    \begin{tabular}{lccccccc}
        \toprule
        \multirow{2}{*}{\textbf{Category}} & 
        \multicolumn{3}{c}{\textbf{Autoencoder}} & 
        \multicolumn{3}{c}{\textbf{PCA}} \\
        \cmidrule(lr){2-4} \cmidrule(lr){5-7}
        & \textbf{MSE} & \textbf{Uniform $R^2$} & \textbf{Weighted $R^2$} 
        & \textbf{MSE} & \textbf{Uniform $R^2$} & \textbf{Weighted $R^2$} \\
        \midrule
        Land         & 3.60  & 0.8600 & 0.8948 & 28.32 & 0.7810 & 0.8397 \\
        Consumer     & 0.92  & 0.8490 & 0.8682 & 8.93  & 0.7307 & 0.7809 \\
        Debt         & 3.94  & 0.7673 & 0.8573 & 28.40 & 0.7199 & 0.7993 \\
        Employment   & 2.09  & 0.8029 & 0.8552 & 11.71 & 0.7035 & 0.7900 \\
        Agriculture  & 6759.88 & 0.8452 & 0.8694 & 1.85$\times10^9$ & 0.6754 & 0.6799 \\
        Housing      & 5.85  & 0.8601 & 0.8688 & 58.99 & 0.7857 & 0.8118 \\
        \bottomrule
    \end{tabular}}
    \label{table:ae_metrics}
\end{table*}

Following the dimensionality reduction step, a regression model was trained to predict the category-specific latent NSSO encodings at the district level. The model's inputs consisted of district-level Census data augmented with auxiliary geographic features and one-hot encoded state identifiers. Separate regression models were trained for the 2001 and 2011 datasets, and their respective performance statistics are presented in Table~\ref{table:regression_2001_2011_comparison}.\par

The results indicate strong predictive performance for most categories, with Weighted $R^2$ values generally exceeding 0.65. The Consumer Expenditure and Housing Conditions categories demonstrated the highest correspondence between predicted and ground-truth encodings. In contrast, the Agriculture and Employment categories proved more challenging to predict, exhibiting lower $R^2$ values and higher MSE. This is likely attributable to the presence of more idiosyncratic, region-specific patterns or greater noise within these particular variables. \par

\begin{table*}[!htb]
    \centering
    \caption{Regression model performance for predicting latent NSSO encodings from census data (2001, 2011). Mean Squared Error (MSE) and $R^2$ during training for predicting NSSO encodings from Census data across socioeconomic categories, evaluated against ground truth NSSO encodings for $2001$ and $2011$. The high $R^2$ values indicate the model successfully learned the non-linear relationship between census data and highly compressed, low-dimensional feature space.}
    \footnotesize
    \begin{tabular}{lcccc}
        \toprule
        \multirow{2}{*}{\textbf{Category}} & \multicolumn{2}{c}{\textbf{2001}} & \multicolumn{2}{c}{\textbf{2011}} \\
        \cmidrule(lr){2-3} \cmidrule(lr){4-5}
        & \textbf{MSE} & \textbf{$R^2$} & \textbf{MSE} & \textbf{$R^2$} \\
        \midrule
        Land         & $6.11 \times 10^{-5}$   & 0.9937 & $1.28 \times 10^{-4}$ & 0.9880 \\
        Consumer     & $1.64 \times 10^{-4}$   & 0.9917 & $9.10 \times 10^{-5}$ & 0.9937 \\
        Debt         & $7.27 \times 10^{-5}$   & 0.9923 & $1.13 \times 10^{-4}$ & 0.9694 \\
        Employment   & $1.03 \times 10^{-4}$   & 0.9950 & $1.90 \times 10^{-4}$ & 0.9717 \\
        Agriculture  & $8.73 \times 10^{-5}$   & 0.9913 & $2.68 \times 10^{-4}$ & 0.9753 \\
        Housing      & $1.09 \times 10^{-4}$   & 0.9954 & $1.75 \times 10^{-4}$ & 0.9869 \\
        \bottomrule
    \end{tabular}
    \label{table:regression_2001_2011_comparison}
\end{table*}

\begin{figure}[!htb]
        \centering
    
        \begin{subfigure}[!htb]{0.28\linewidth}
            \centering
            \includegraphics[width=\linewidth]{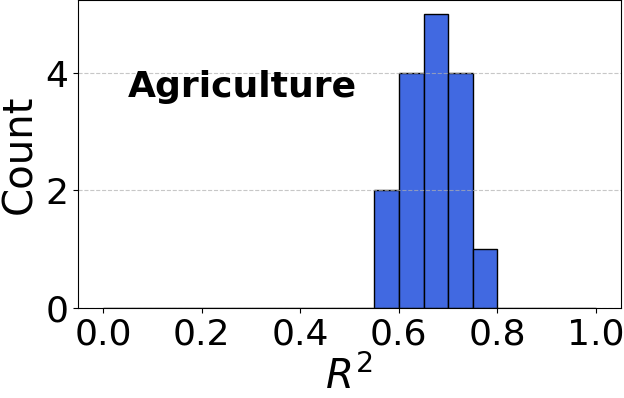}
            \captionsetup{justification=centering}
            \caption{}
        \end{subfigure}
        \begin{subfigure}[!htb]{0.28\linewidth}
            \centering
            \includegraphics[width=\linewidth]{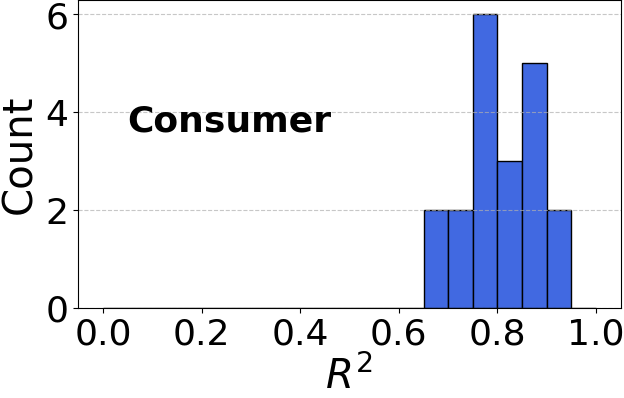}
            \captionsetup{justification=centering}
            \caption{}
        \end{subfigure} 
        \begin{subfigure}[!htb]{0.28\linewidth}
            \centering
            \includegraphics[width=\linewidth]{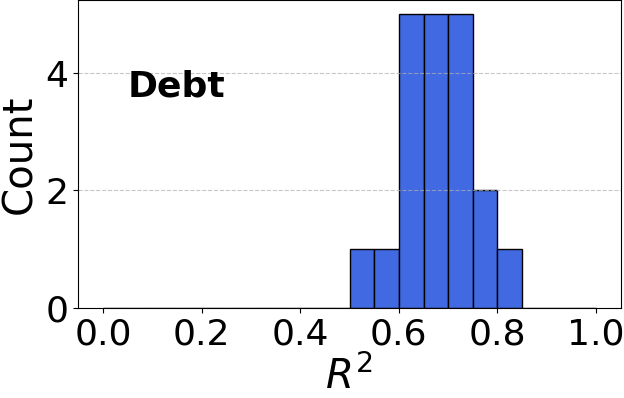}
            \captionsetup{justification=centering}
            \caption{}
        \end{subfigure} 
        \begin{subfigure}[!htb]{0.28\linewidth}
            \centering
            \includegraphics[width=\linewidth]{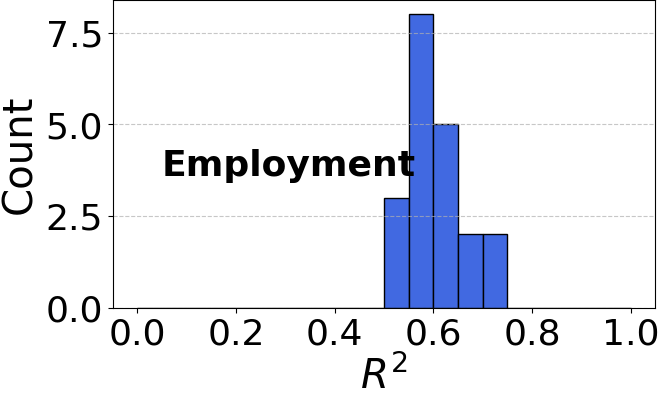}
            \captionsetup{justification=centering}
            \caption{}
        \end{subfigure}
        \begin{subfigure}[!htb]{0.28\linewidth}
            \centering
            \includegraphics[width=\linewidth]{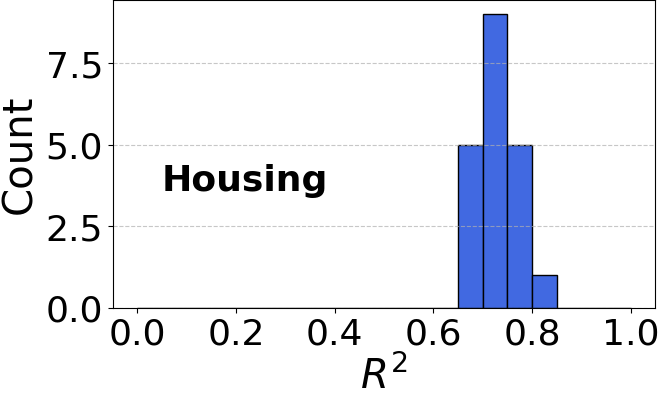}
            \captionsetup{justification=centering}
            \caption{}
        \end{subfigure} 
        \begin{subfigure}[!htb]{0.28\linewidth}
            \centering
            \includegraphics[width=\linewidth]{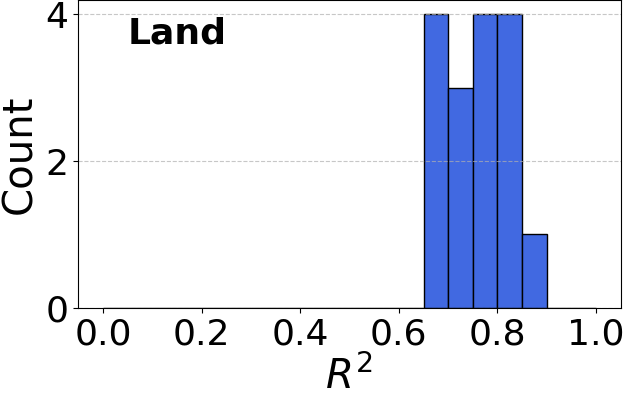}
            \captionsetup{justification=centering}
            \caption{}
        \end{subfigure} 

        \caption[]{Histograms of $R^2$ scores at the district level comparing aggregated cluster-based encodings predictions with ground truth encodings, shown separately for each category. The data is from 2001 for (a) Agriculture, (b) Consumer, (c) Debt, (d) Employment, (e) Housing, and (f) Land categories.}
        \label{fig:reg_enc2001_r2}
    
    \end{figure}

\begin{figure}[!htb]
        \centering
    
        \begin{subfigure}[!htb]{0.28\linewidth}
            \centering
            \includegraphics[width=\linewidth]{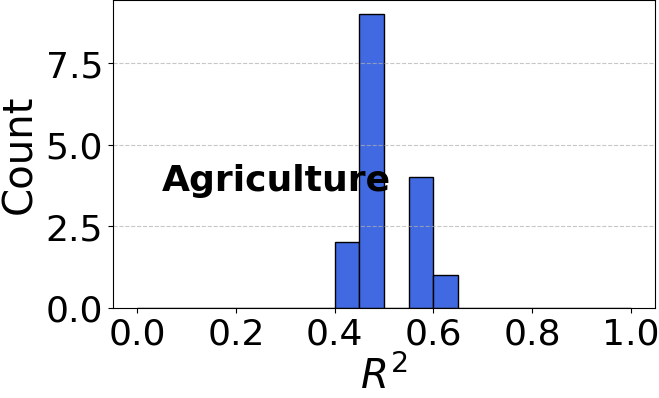}
            \captionsetup{justification=centering}
            \caption{}
        \end{subfigure}
        \begin{subfigure}[!htb]{0.28\linewidth}
            \centering
            \includegraphics[width=\linewidth]{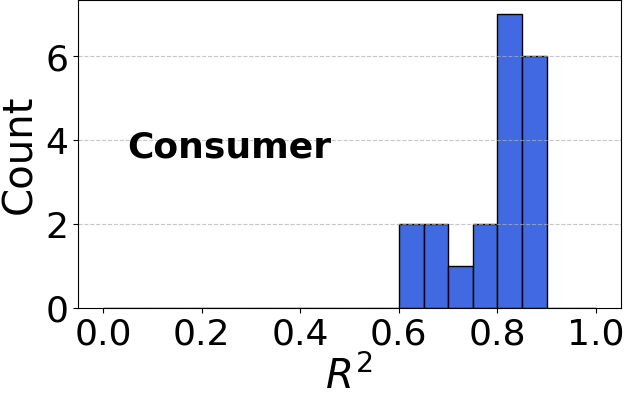}
            \captionsetup{justification=centering}
            \caption{}
        \end{subfigure} 
        \begin{subfigure}[!htb]{0.28\linewidth}
            \centering
            \includegraphics[width=\linewidth]{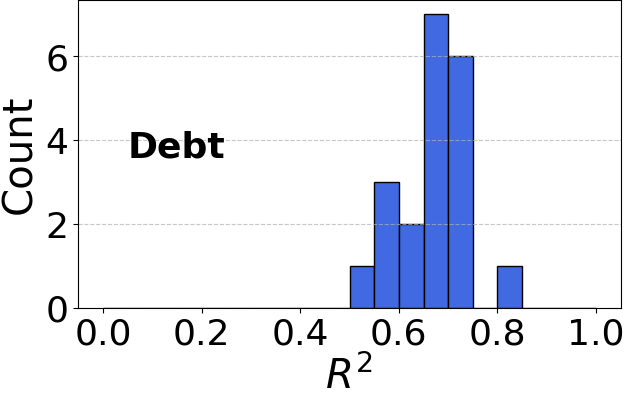}
            \captionsetup{justification=centering}
            \caption{}
        \end{subfigure} 
        \begin{subfigure}[!htb]{0.28\linewidth}
            \centering
            \includegraphics[width=\linewidth]{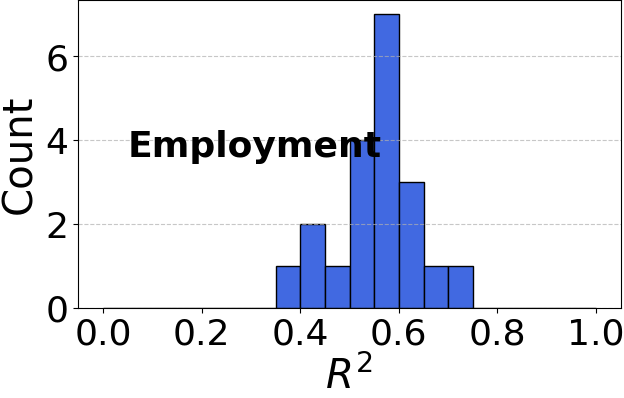}
            \captionsetup{justification=centering}
            \caption{}
        \end{subfigure}
        \begin{subfigure}[!htb]{0.28\linewidth}
            \centering
            \includegraphics[width=\linewidth]{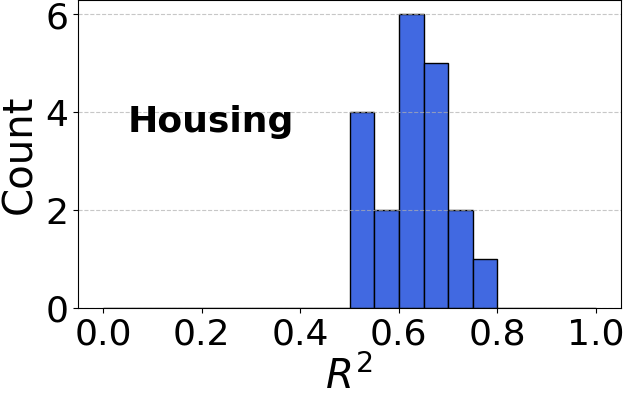}
            \captionsetup{justification=centering}
            \caption{}
        \end{subfigure} 
        \begin{subfigure}[!htb]{0.28\linewidth}
            \centering
            \includegraphics[width=\linewidth]{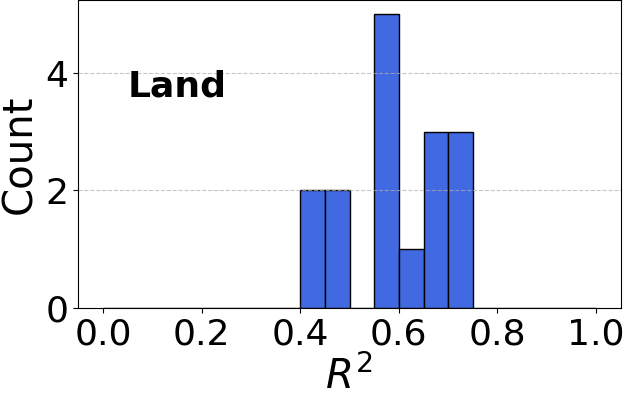}
            \captionsetup{justification=centering}
            \caption{}
        \end{subfigure} 

        \caption[]{Histograms of $R^2$ scores at the district level comparing aggregated cluster-based encodings predictions with ground truth encodings, shown separately for each category. The data is from 2011 for (a) Agriculture, (b) Consumer, (c) Debt, (d) Employment, (e) Housing, and (f) Land categories.}
        \label{fig:reg_enc2011_r2}
    
    \end{figure}

Following the training phase, regression models were deployed for inference to predict the NSSO latent encodings at the cluster level, using the corresponding high-resolution census and auxiliary features as input. To validate these downscaled predictions, the predicted cluster-level encodings were then aggregated to the district level, enabling a direct comparison with the ground-truth district-level encodings. The summary statistics for this comparison are presented in Table \ref{table:agg_enc_2001_2011_comparison}. Furthermore, Figs. \ref{fig:reg_enc2001_r2} and \ref{fig:reg_enc2011_r2} present histograms of the resulting $R^2$ scores for the years 2001 and 2011, respectively, illustrating the distribution of predictive accuracy across the districts. \par


\begin{table*}[!htb]
    \centering
    \caption{Performance of cluster-level encoding predictions aggregated to the district scale. Summary statistics (MSE, $R^{2}$) quantify the model's accuracy after generating fine-scale cluster-level predictions of latent NSSO encodings, aggregating them to the district level, and comparing against the ground-truth NSSO latent representations for 2001 and 2011.}

    \footnotesize
    \begin{tabular}{lcccccc}
        \toprule
        \multirow{2}{*}{\textbf{Category}} &
        \multicolumn{3}{c}{\textbf{2001}} &
        \multicolumn{3}{c}{\textbf{2011}} \\
        \cmidrule(lr){2-4} \cmidrule(lr){5-7}
        & \textbf{MSE} & \textbf{Uniform $R^2$} & \textbf{Weighted $R^2$}
        & \textbf{MSE} & \textbf{Uniform $R^2$} & \textbf{Weighted $R^2$} \\
        \midrule
        Land         & 0.0487 & 0.7603 & 0.7886 & 0.0657 & 0.5955 & 0.6344 \\
        Consumer     & 0.0621 & 0.8082 & 0.8289 & 0.0537 & 0.7975 & 0.8237 \\
        Debt         & 0.0550 & 0.6805 & 0.6950 & 0.0356 & 0.6771 & 0.6912 \\
        Employment   & 0.0948 & 0.6030 & 0.5930 & 0.0546 & 0.5539 & 0.5658 \\
        Agriculture  & 0.0579 & 0.6674 & 0.6745 & 0.0743 & 0.4999 & 0.5092 \\
        Housing      & 0.0844 & 0.7339 & 0.7480 & 0.0709 & 0.6260 & 0.6408 \\
        \bottomrule
    \end{tabular}
    \label{table:agg_enc_2001_2011_comparison}
\end{table*}

An analysis of the predictive performance across the different socioeconomic categories reveals significant variation by both theme and year. The Consumer Expenditure category consistently exhibited the strongest performance, achieving Weighted $R^2$ values of $0.8289$ (2001) and $0.8237$ (2011), which suggests its underlying latent structure is well-captured by the available predictor variables.

The Land and Livestock Holdings and Housing Conditions categories also demonstrated high predictive accuracy in 2001 (Weighted $R^2>$0.74); however, both experienced a notable decline in performance in 2011, with the Land category showing a substantial drop from $0.7886$ to $0.6344$. In contrast, the Debt and Investment and Employment categories showed more consistent, albeit moderate, performance across the two periods, with $R^2$ values typically ranging between $0.55$ and $0.70$.

The Agriculture category was consistently the most challenging to predict, with its Weighted $R^2$ falling to $0.5092$ in 2011. This lower performance likely reflects a high degree of variability and noise inherent in agricultural indicators that are not fully explained by the model's inputs. Collectively, these results indicate that while the latent NSSO encodings can be effectively approximated from high-resolution data, the model's accuracy is heterogeneous, varying significantly by both socioeconomic category and year.

\begin{figure}[!htb]
        \centering
    
        \begin{subfigure}[!htb]{0.28\linewidth}
            \centering
            \includegraphics[width=\linewidth]{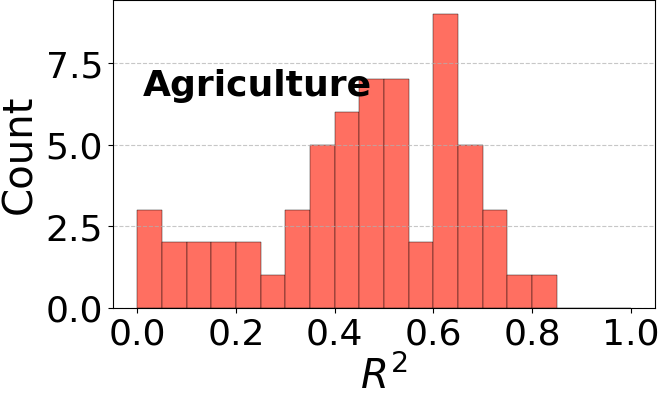}
            \captionsetup{justification=centering}
            \caption{}
        \end{subfigure}
        \begin{subfigure}[!htb]{0.28\linewidth}
            \centering
            \includegraphics[width=\linewidth]{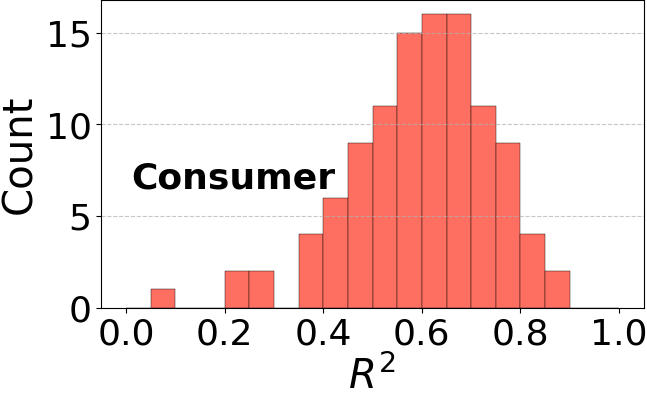}
            \captionsetup{justification=centering}
            \caption{}
        \end{subfigure} 
        \begin{subfigure}[!htb]{0.28\linewidth}
            \centering
            \includegraphics[width=\linewidth]{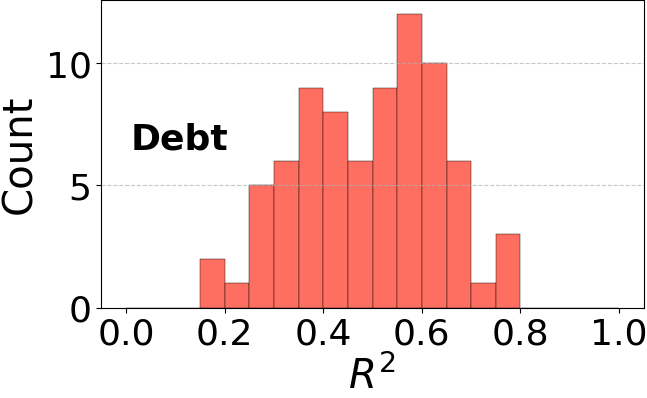}
            \captionsetup{justification=centering}
            \caption{}
        \end{subfigure} 
        \begin{subfigure}[!htb]{0.28\linewidth}
            \centering
            \includegraphics[width=\linewidth]{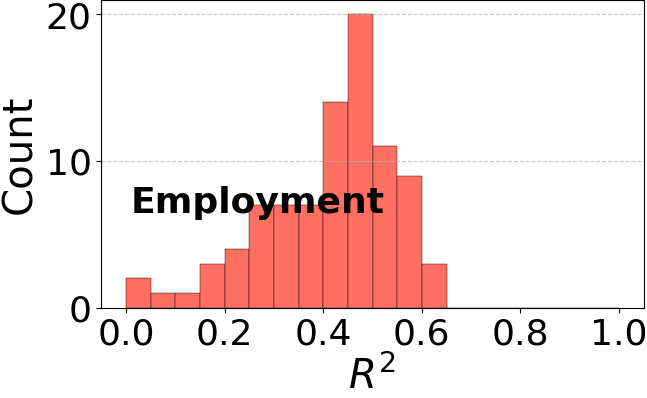}
            \captionsetup{justification=centering}
            \caption{}
        \end{subfigure}
        \begin{subfigure}[!htb]{0.28\linewidth}
            \centering
            \includegraphics[width=\linewidth]{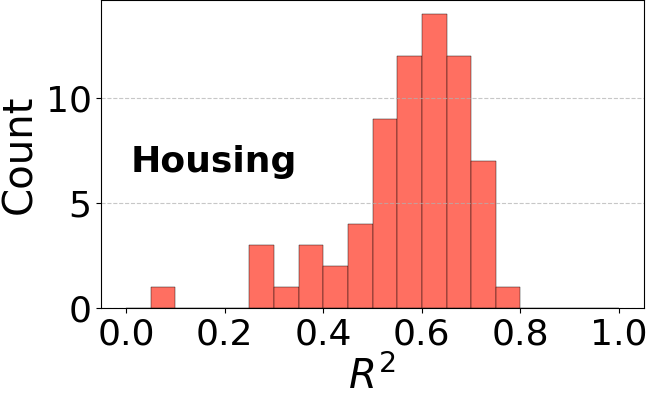}
            \captionsetup{justification=centering}
            \caption{}
        \end{subfigure} 
        \begin{subfigure}[!htb]{0.28\linewidth}
            \centering
            \includegraphics[width=\linewidth]{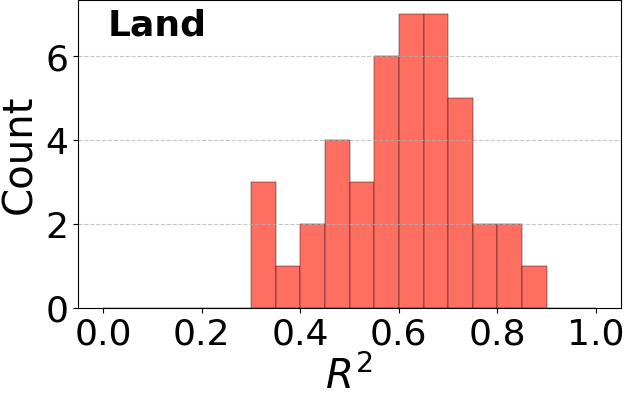}
            \captionsetup{justification=centering}
            \caption{}
        \end{subfigure} 

        \caption[]{Histograms of $R^2$ scores at the district level comparing aggregated cluster-based full-scale decoded predictions with ground truth NSSO data, shown separately for each category. The data is from 2001 for (a) Agriculture, (b) Consumer, (c) Debt, (d) Employment, (e) Housing, and (f) Land categories.}
        \label{fig:reg_dec2001_r2}
    
    \end{figure}

\begin{figure}[!htb]
        \centering
    
        \begin{subfigure}[!htb]{0.28\linewidth}
            \centering
            \includegraphics[width=\linewidth]{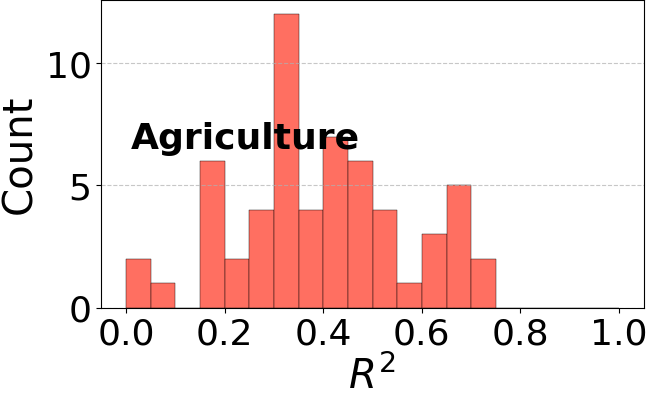}
            \captionsetup{justification=centering}
            \caption{}
        \end{subfigure}
        \begin{subfigure}[!htb]{0.28\linewidth}
            \centering
            \includegraphics[width=\linewidth]{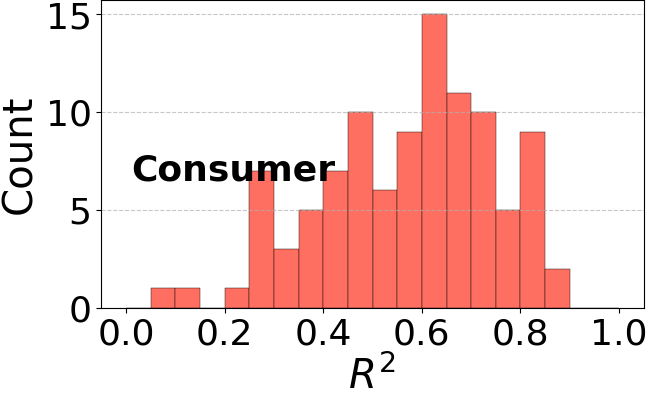}
            \captionsetup{justification=centering}
            \caption{}
        \end{subfigure} 
        \begin{subfigure}[!htb]{0.28\linewidth}
            \centering
            \includegraphics[width=\linewidth]{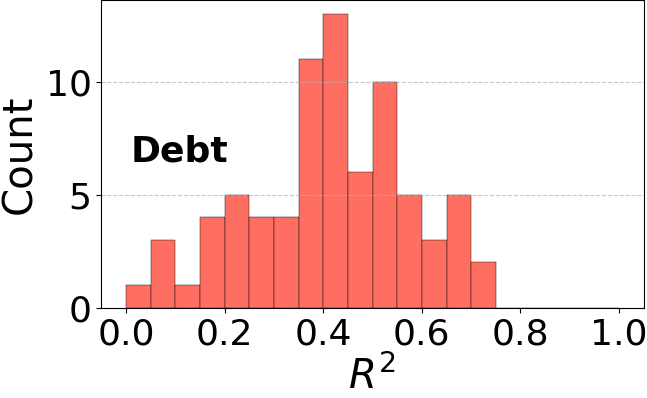}
            \captionsetup{justification=centering}
            \caption{}
        \end{subfigure} 
        \begin{subfigure}[!htb]{0.28\linewidth}
            \centering
            \includegraphics[width=\linewidth]{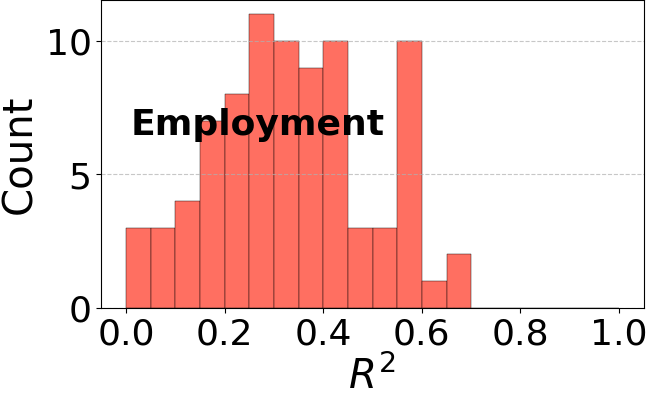}
            \captionsetup{justification=centering}
            \caption{}
        \end{subfigure}
        \begin{subfigure}[!htb]{0.28\linewidth}
            \centering
            \includegraphics[width=\linewidth]{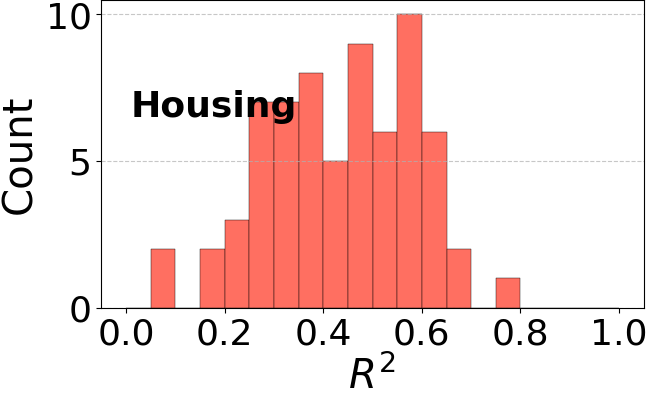}
            \captionsetup{justification=centering}
            \caption{}
        \end{subfigure} 
        \begin{subfigure}[!htb]{0.28\linewidth}
            \centering
            \includegraphics[width=\linewidth]{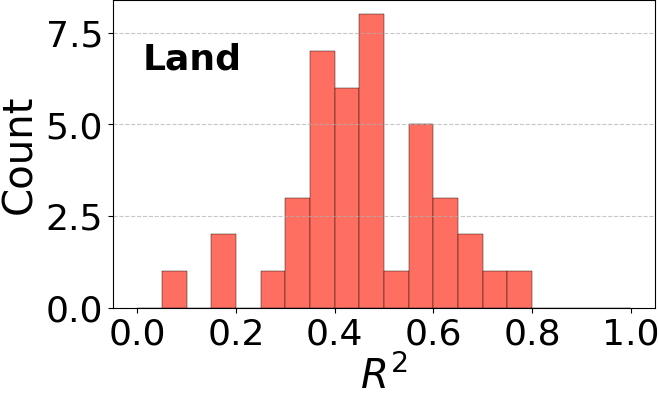}
            \captionsetup{justification=centering}
            \caption{}
        \end{subfigure} 

        \caption[]{Histograms of $R^2$ scores at the district level comparing aggregated cluster-based full-scale decoded predictions with ground truth NSSO data, shown separately for each category. The data is from 2011 for (a) Agriculture, (b) Consumer, (c) Debt, (d) Employment, (e) Housing, and (f) Land categories.}
        \label{fig:reg_dec2011_r2}
    
    \end{figure}

The final stage of the methodology involves generating the full-scale, high-resolution NSSO indicators. This was accomplished by passing the predicted cluster-level latent encodings through their respective pre-trained, category-specific decoders. To quantitatively validate these final reconstructed indicators, they were first spatially aggregated to the district level by averaging the predictions of all clusters within each district boundary. This aggregation allows for a direct comparison against the ground-truth district-level NSSO data. Table \ref{table:agg_dec_2001_2011_res} summarizes the performance of this final reconstruction, reporting the Mean Squared Error (MSE), the overall $R^2$, and the number of individual indicators achieving an $R^2\geq0.5$ for both years. Additionally, the distributions of the $R^2$ scores are visualized in the histograms presented in Figs. \ref{fig:reg_dec2001_r2} and \ref{fig:reg_dec2011_r2}.

The final evaluation assessed the correspondence between the fully reconstructed high-resolution indicators (aggregated to the district level) and the ground-truth NSSO data. While the decoding process introduced an expected degree of information loss compared to the latent space prediction, the final indicators for most categories demonstrated a satisfactory correspondence with the ground truth.

The Consumer Expenditure and Land and Livestock Holdings categories were the most accurately reconstructed, yielding Weighted $R^2$ values of approximately $0.69$ and $0.68$, respectively, in 2001, with only a modest decline in performance observed in 2011. The robustness of the reconstruction for the Consumer Expenditure category is further evidenced by the fact that over $85\%$ of its constituent variables consistently achieved an $R^2\geq 0.5$ in both years, indicating a reliable recovery of the primary underlying patterns.
In contrast, categories like Agriculture and Employment showed lower $R^2$ scores and higher MSE, particularly in 2011. For example, Agriculture’s Weighted $R^2$ dropped to $0.2887$, with fewer than half of the variables meeting the $R^2 \geq 0.5$ threshold. These results suggest that some fine-grained details may be lost during encoding, prediction, and decoding—especially in categories with more variable or nonlinear distributions. \par

In summary, the results demonstrate that the end-to-end pipeline can successfully reconstruct coherent socioeconomic patterns from the predicted latent encodings, although performance varies by category and year. Crucially, this validates the central premise of our approach: that it is feasible to generate meaningful, high-resolution estimates of complex survey-based indicators using only census and geographic features as inputs during the inference stage, without recourse to the target survey data itself.\par

\begin{table*}[!htb]
    \centering
    \caption{End-to-end predictive performance of downscaled socioeconomic indicators after full reconstruction. The table shows the performance metrics (MSE, $R^{2}$) for the final, high-resolution cluster-level predictions after they have been aggregated back to the district level and compared against the original ground-truth NSSO data for 2001 and 2011. The column ``\textit{Vars with $R^{2} \ge 0.5$}" indicates the percentage of individual variables demonstrating adequate predictive correspondence. Note the negative $R^{2}$ values for Agriculture, suggesting low predictive power for highly heterogeneous indicators after the full downscaling process.}
    \resizebox{\linewidth}{!}{    
    \begin{tabular}{lcccccccr}
        \toprule
        \multirow{2}{*}{\textbf{Category}} &
        \multicolumn{4}{c}{\textbf{2001}} &
        \multicolumn{4}{c}{\textbf{2011}} \\
        \cmidrule(lr){2-5} \cmidrule(lr){6-9}
        & \textbf{MSE} & \shortstack{\textbf{Uniform} \\ \boldmath{$R^2$}} & \shortstack{\textbf{Weighted} \\ \boldmath{$R^2$}} & \shortstack{\textbf{Vars with} \\ \boldmath{$R^2 \geq 0.5$}} 
        & \textbf{MSE} & \shortstack{\textbf{Uniform} \\ \boldmath{$R^2$}} & \shortstack{\textbf{Weighted} \\ \boldmath{$R^2$}} & \shortstack{\textbf{Vars with} \\ \boldmath{$R^2 \geq 0.5$}} \\
        \midrule
        Land         & 5.3006   & 0.5615  & 0.6843 & 40 / 45  & 6.3567   & 0.4128  & 0.5580 & 29 / 45 \\
        Consumer     & 1.4001   & 0.5384  & 0.6902 & 103 / 110 & 1.3205   & 0.5038  & 0.6581 & 87 / 110 \\
        Debt         & 6.5405   & 0.4402  & 0.5663 & 56 / 83  & 4.9339   & 0.2946  & 0.5182 & 47 / 83 \\
        Employment   & 4.0608   & 0.4301  & 0.4962 & 62 / 92  & 2.7764   & 0.2839  & 0.5045 & 30 / 92 \\
        Agriculture  & 11875.6300 & -0.2405 & 0.4537 & 46 / 76  & 12962.6700 & -8.1636 & 0.2887 & 30 / 76 \\
        Housing      & 10.5353  & 0.5791  & 0.6318 & 63 / 69  & 9.8129   & 0.4086  & 0.4942 & 40 / 69 \\
        \midrule
        \textbf{Total} & -- & -- & -- & \textbf{370 / 475} & -- & -- & -- & \textbf{263 / 475} \\
        \bottomrule
    \end{tabular}}
    \label{table:agg_dec_2001_2011_res}
\end{table*}

\subsection{Spatial Visualization of Downscaled Predictions}
Beyond the statistical performance metrics, a crucial evaluation of the downscaling model involves analyzing the spatial patterns of its predictions. Visualizing the downscaled indicators on a map offers qualitative insights into regional trends that may not be apparent from aggregate statistics alone. This approach enables a direct visual comparison between the fine-grained predicted patterns and the coarse-grained ground-truth data, providing an assessment of the model's capacity to generate spatially coherent results.\par

\subsubsection{Encoded variables visualization}
The spatial analysis begins with a visualization of the predicted latent encodings. While a detailed semantic interpretation of these latent dimensions is reserved for future investigation, their geographic distributions offer insight into the model's learned representations. We present these spatial distributions at both the high-resolution cluster level and the aggregated district level.

\begin{figure}[!htb]
        \centering
        \begin{subfigure}[!htb]{0.315\linewidth}
            \centering
            \includegraphics[width=\linewidth]{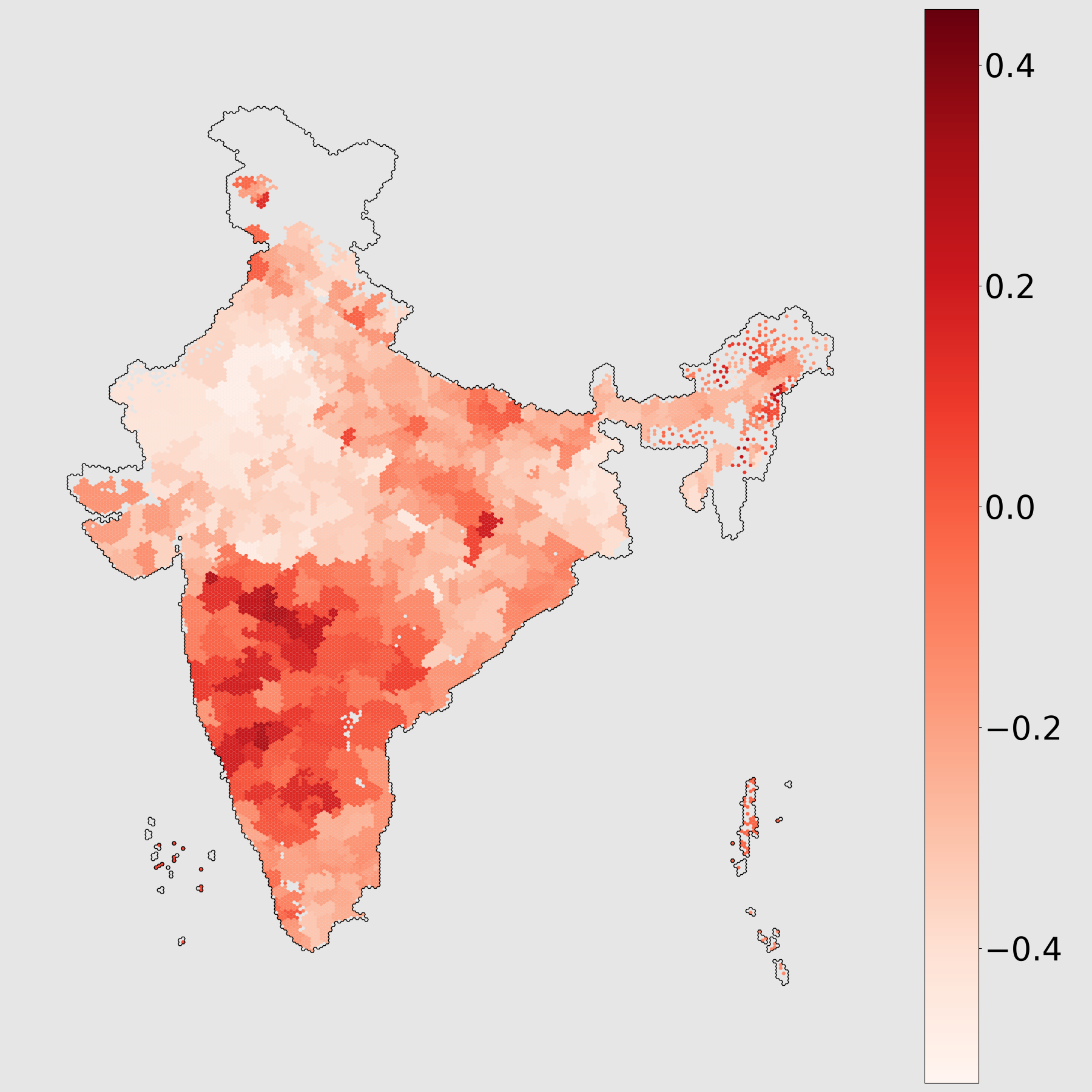}
            \captionsetup{justification=centering}
            \caption{}
        \end{subfigure} 
        \begin{subfigure}[!htb]{0.315\linewidth}
            \centering
            \includegraphics[width=\linewidth]{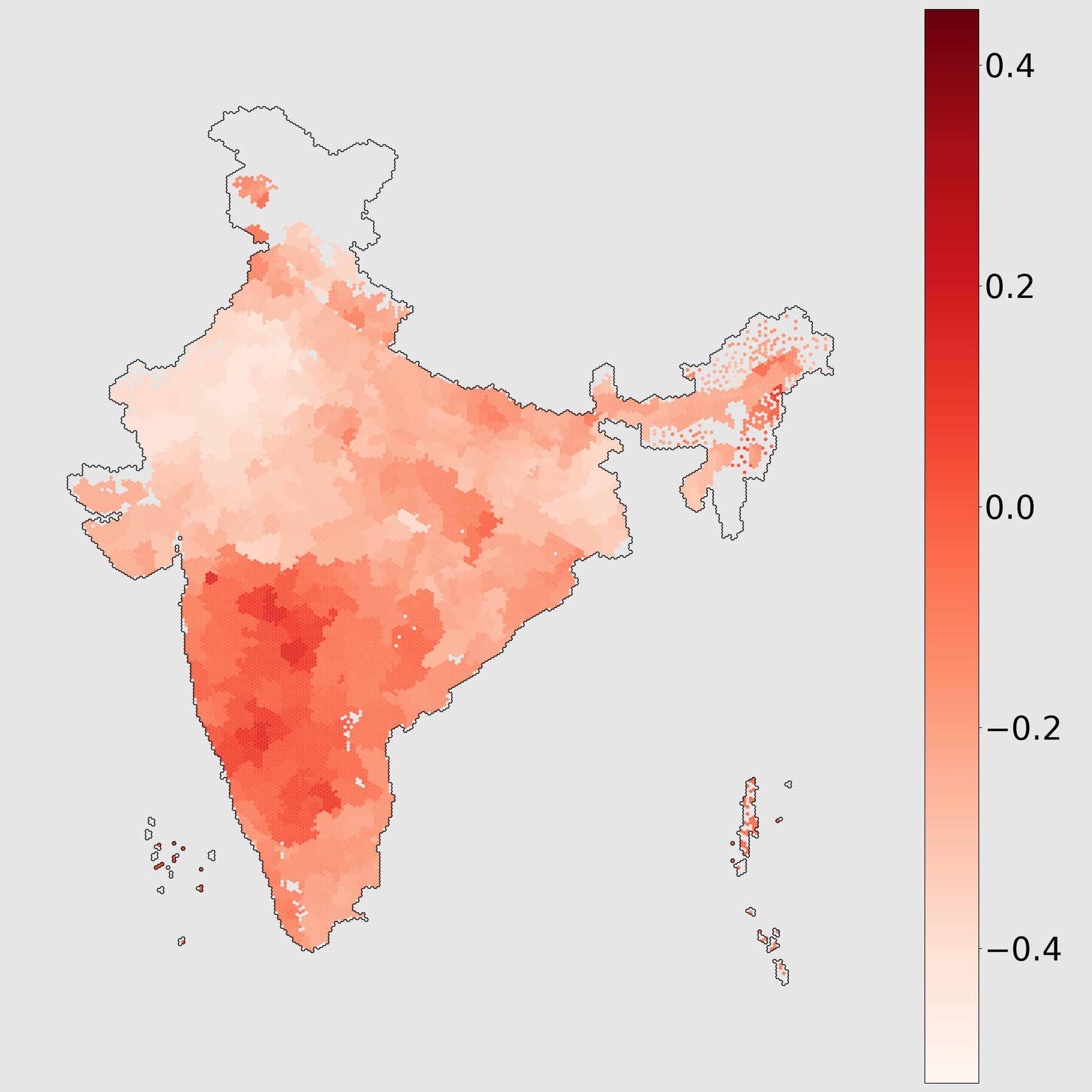}
            \captionsetup{justification=centering}
            \caption{}
        \end{subfigure} 
        \begin{subfigure}[!htb]{0.315\linewidth}
            \centering
            \includegraphics[width=\linewidth]{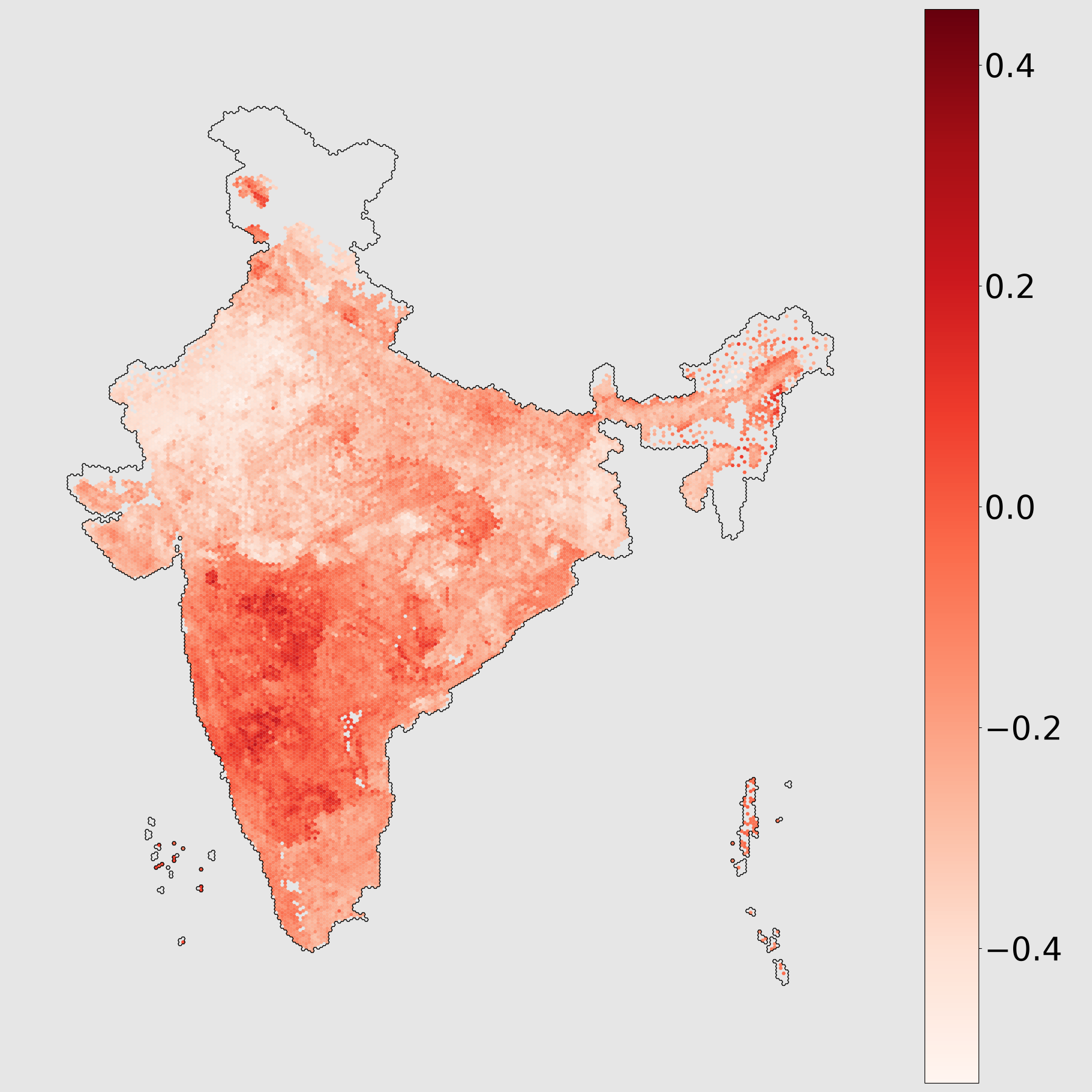}
            \captionsetup{justification=centering}
            \caption{}
        \end{subfigure} \\

        \begin{subfigure}[!htb]{0.315\linewidth}
            \centering
            \includegraphics[width=\linewidth]{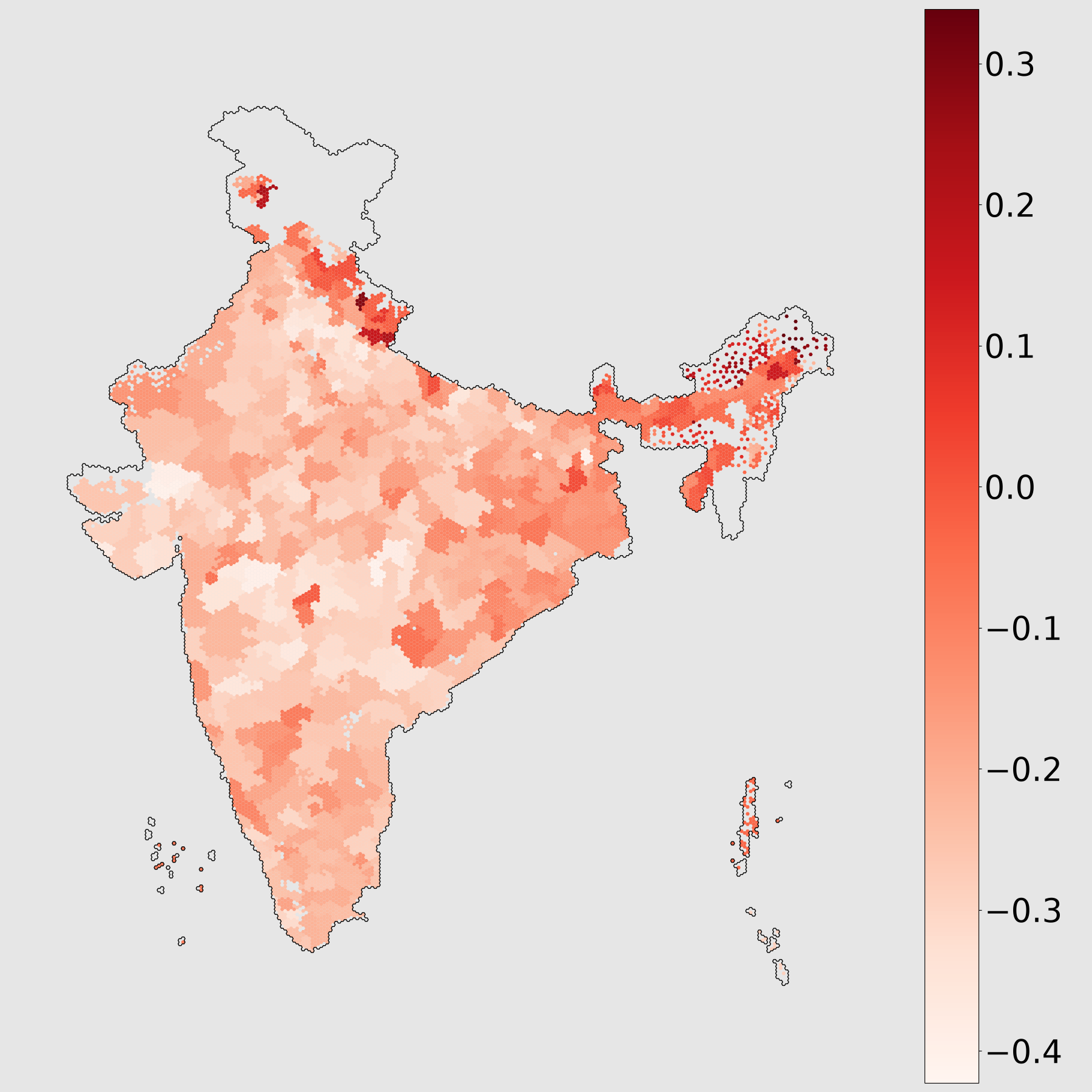}

            \captionsetup{justification=centering}
            \caption{}
        \end{subfigure} 
\begin{subfigure}[!htb]{0.315\linewidth}
            \centering
            \includegraphics[width=\linewidth]{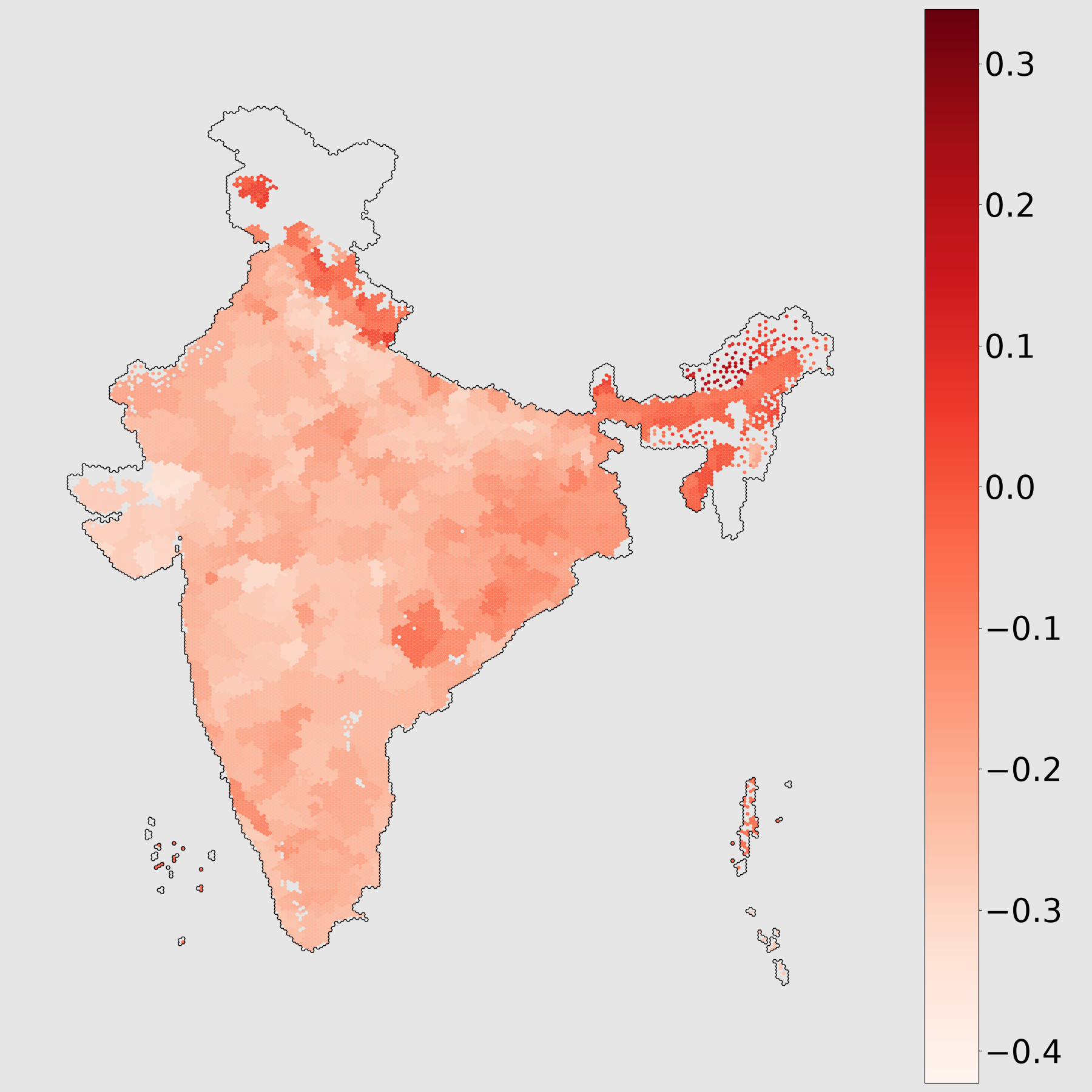}

            \captionsetup{justification=centering}
            \caption{}
        \end{subfigure} 
\begin{subfigure}[!htb]{0.315\linewidth}
            \centering
            \includegraphics[width=\linewidth]{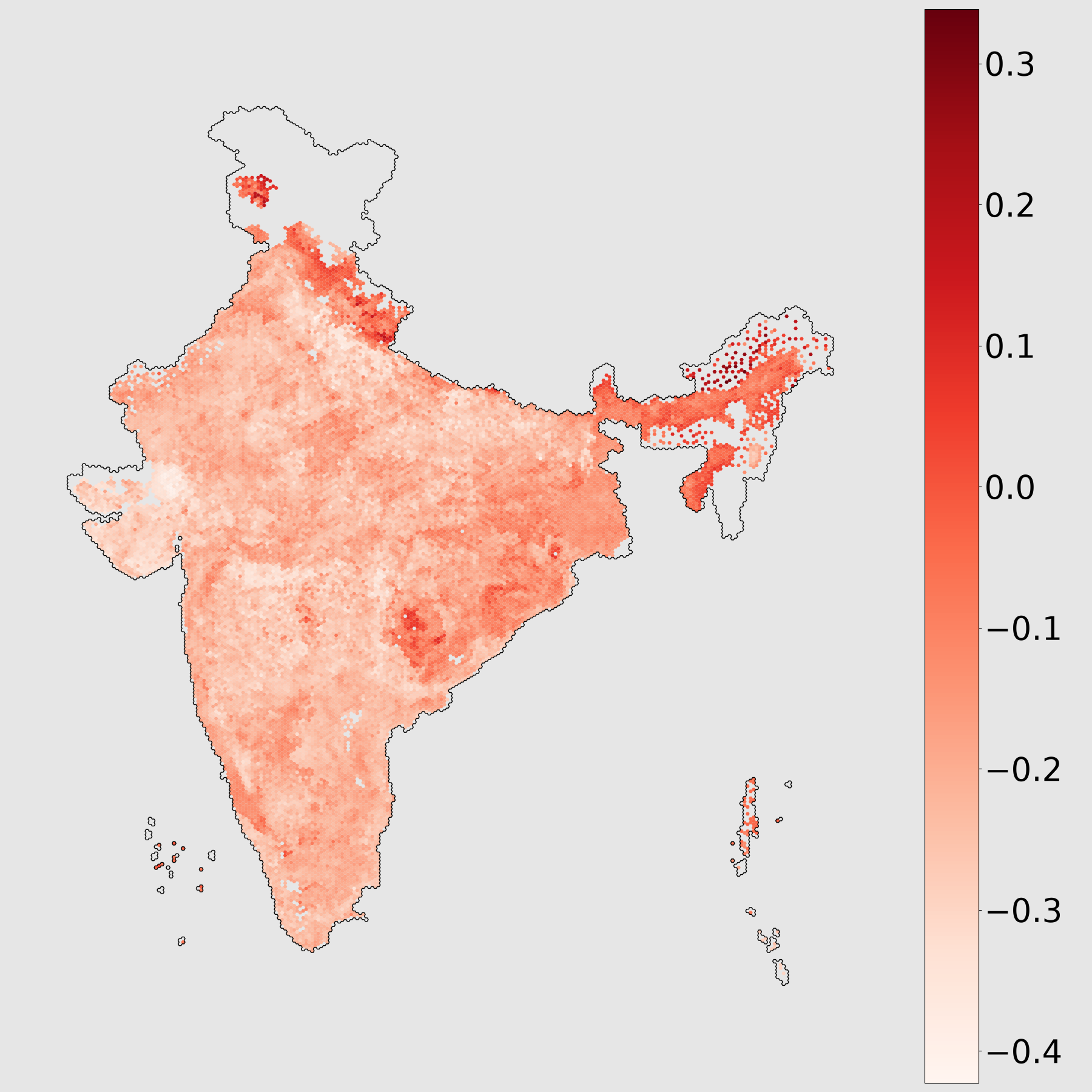}

            \captionsetup{justification=centering}
            \caption{}
        \end{subfigure} \\





        \begin{subfigure}[!htb]{0.315\linewidth}
            \centering
            \includegraphics[width=\linewidth]{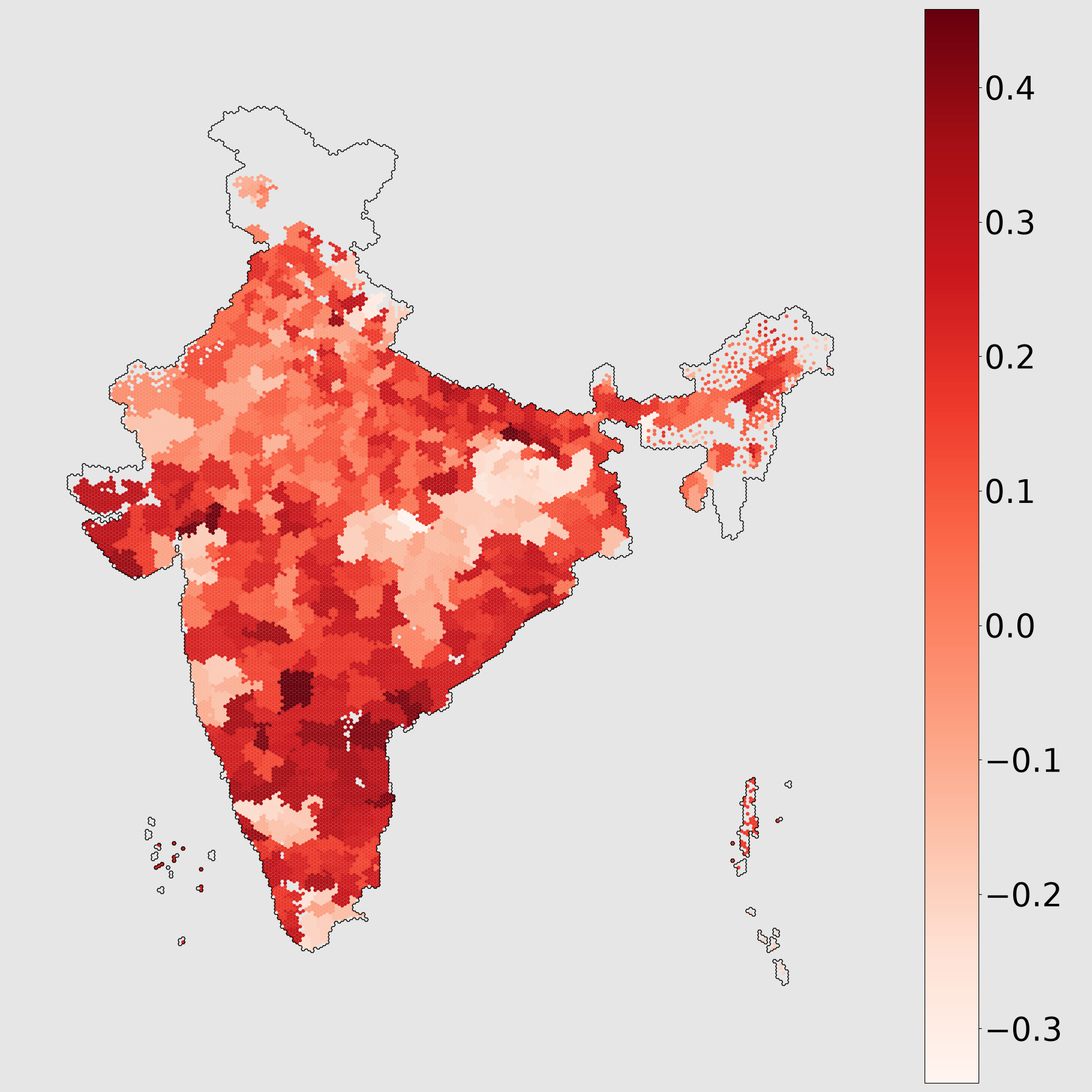}

            \captionsetup{justification=centering}
            \caption{}
        \end{subfigure}
        \begin{subfigure}[!htb]{0.315\linewidth}
            \centering
            \includegraphics[width=\linewidth]{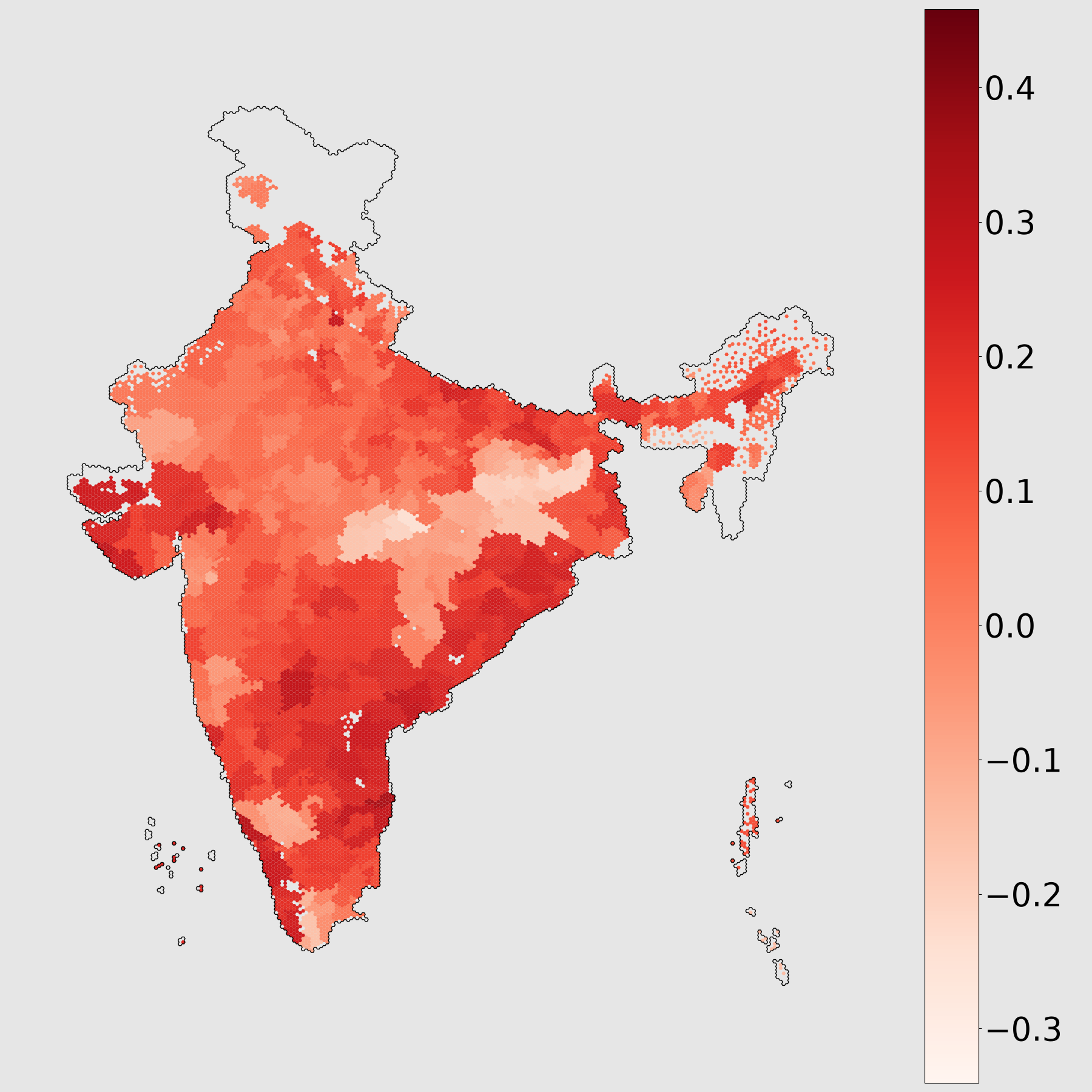}

            \captionsetup{justification=centering}
            \caption{}
        \end{subfigure}
        \begin{subfigure}[!htb]{0.315\linewidth}
            \centering
            \includegraphics[width=\linewidth]{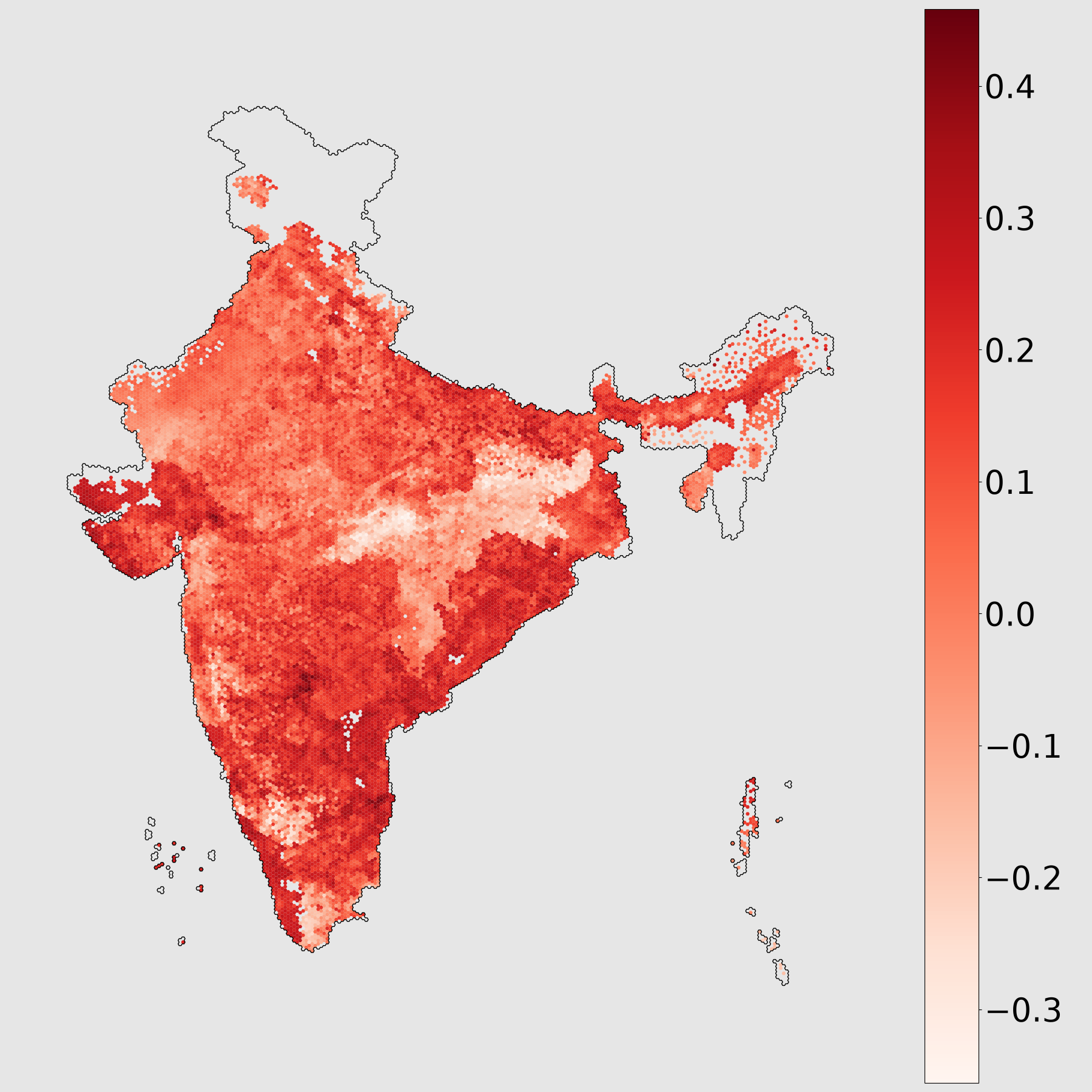}

            \captionsetup{justification=centering}
            \caption{}
        \end{subfigure}

        \caption[]{Spatial Coherence of Predicted Latent Socioeconomic Structures (2001). Each row represents one latent dimension of the compressed NSSO space. Columns: Left (a, d, g) show ground-truth district-level values; Center (b, e, h) show predicted cluster values averaged to the district level; Right (c, f, i) show final high-resolution cluster-level predictions. The visual alignment between the Center and Left columns illustrates the model's qualitative agreement with the ground truth, while the Right column reveals plausible fine-grained, intra-district variations absent in the coarse input.}
        \label{fig:enc_map_2001}

    \end{figure}

\begin{figure}[!htb]
        \centering

        \begin{subfigure}[!htb]{0.315\linewidth}
            \centering
            \includegraphics[width=\linewidth]{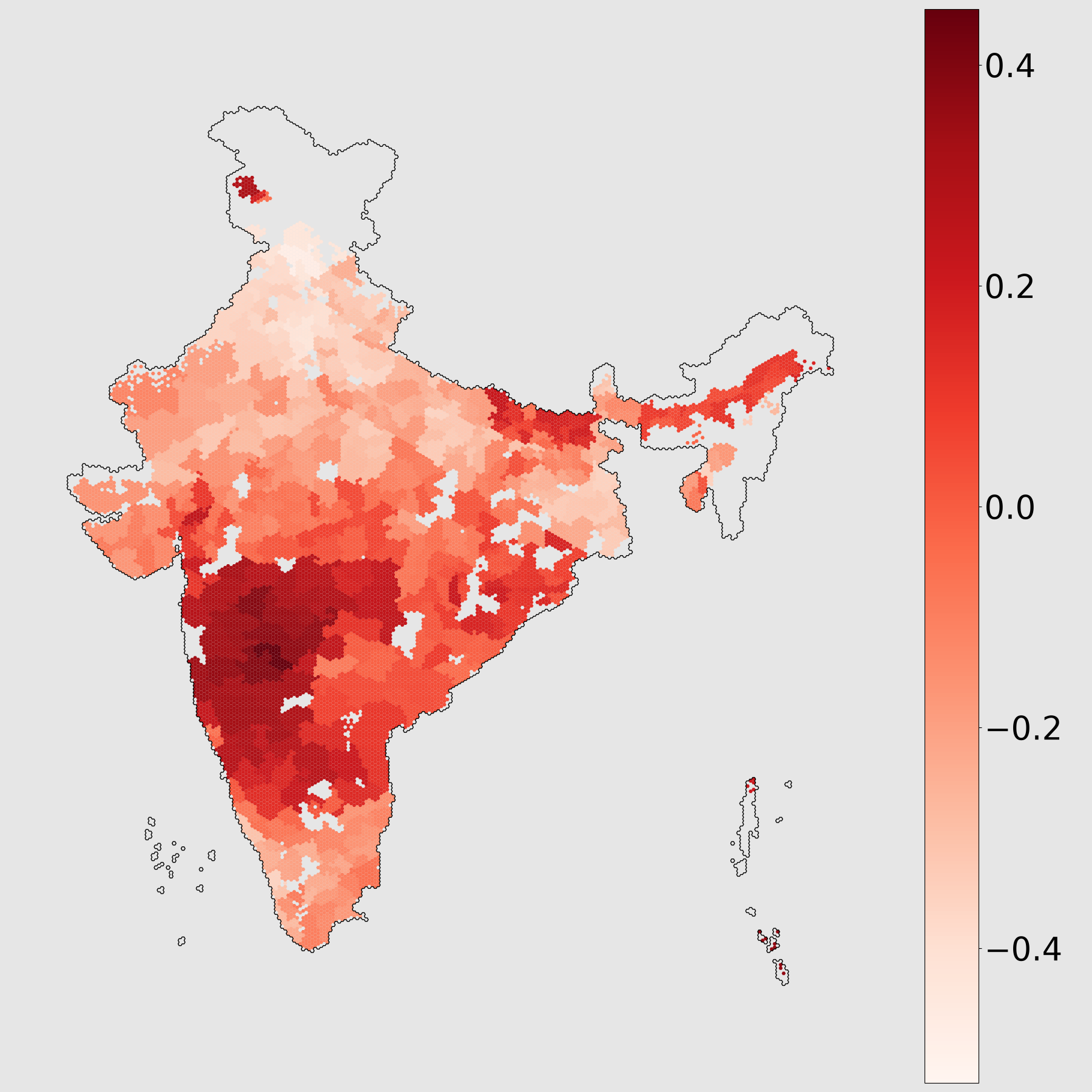}
            \captionsetup{justification=centering}
            \caption{}
        \end{subfigure} 
        \begin{subfigure}[!htb]{0.315\linewidth}
            \centering
            \includegraphics[width=\linewidth]{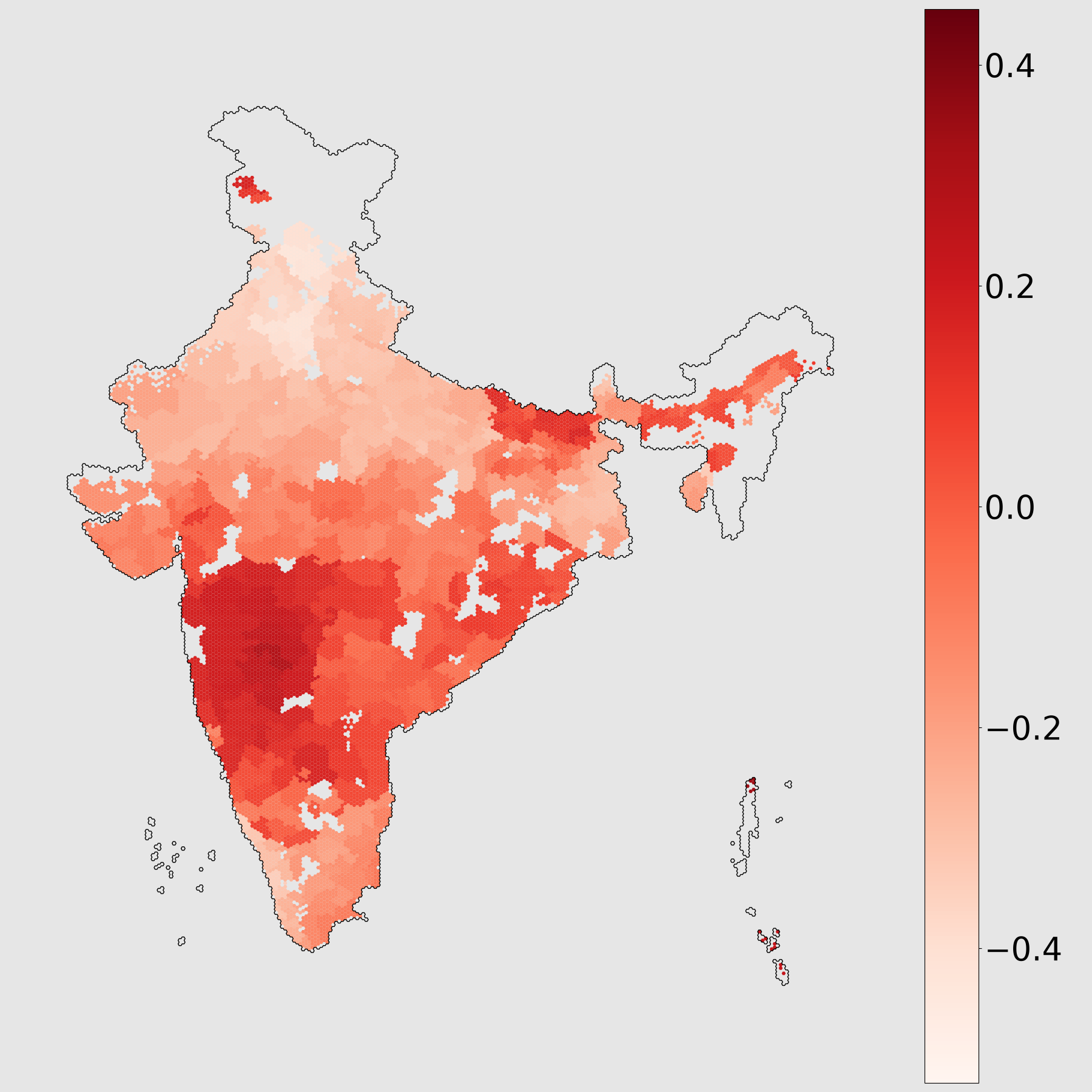}
            \captionsetup{justification=centering}
            \caption{}
        \end{subfigure} 
        \begin{subfigure}[!htb]{0.315\linewidth}
            \centering
            \includegraphics[width=\linewidth]{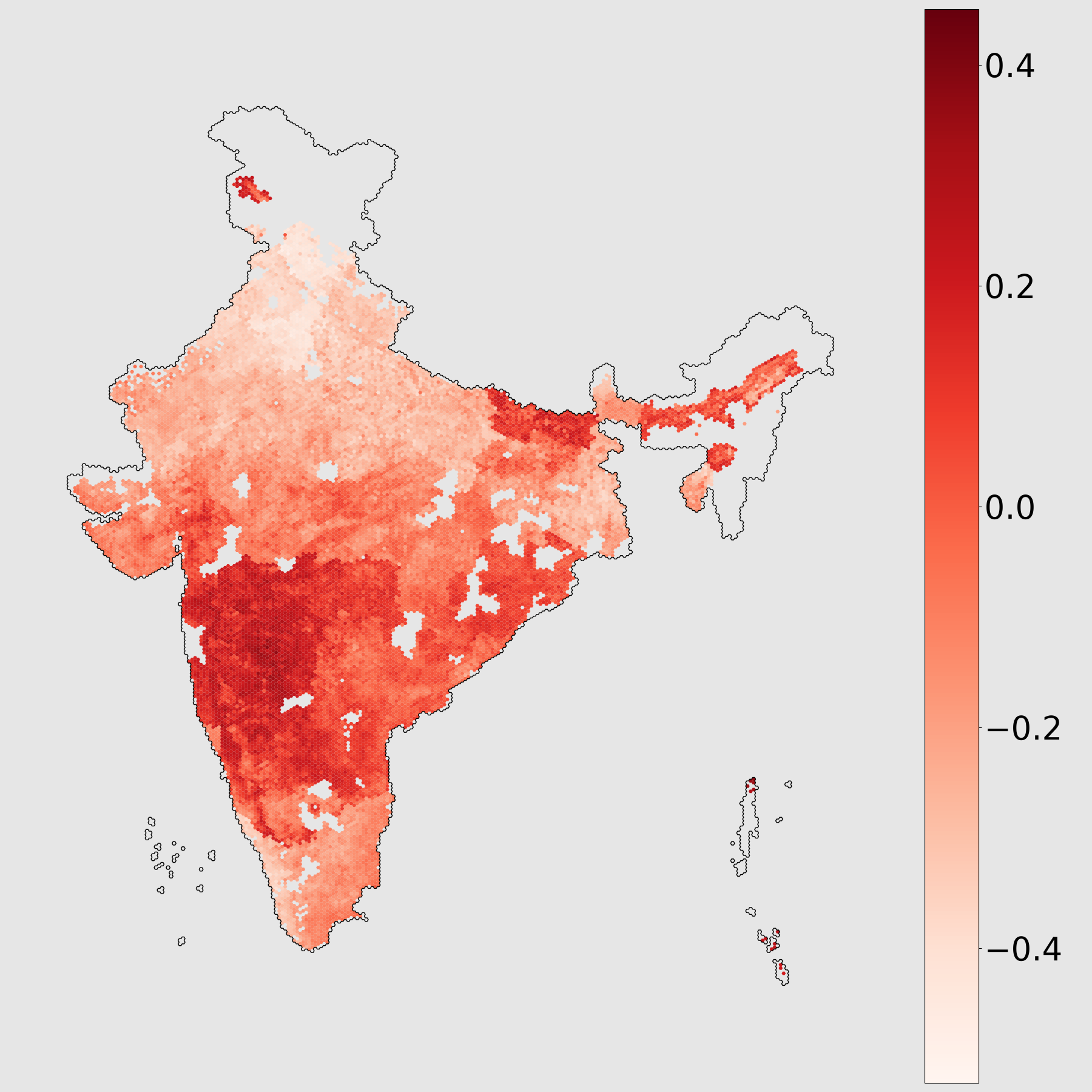}
            \captionsetup{justification=centering}
            \caption{}
        \end{subfigure} \\

        \begin{subfigure}[!htb]{0.315\linewidth}
            \centering
            \includegraphics[width=\linewidth]{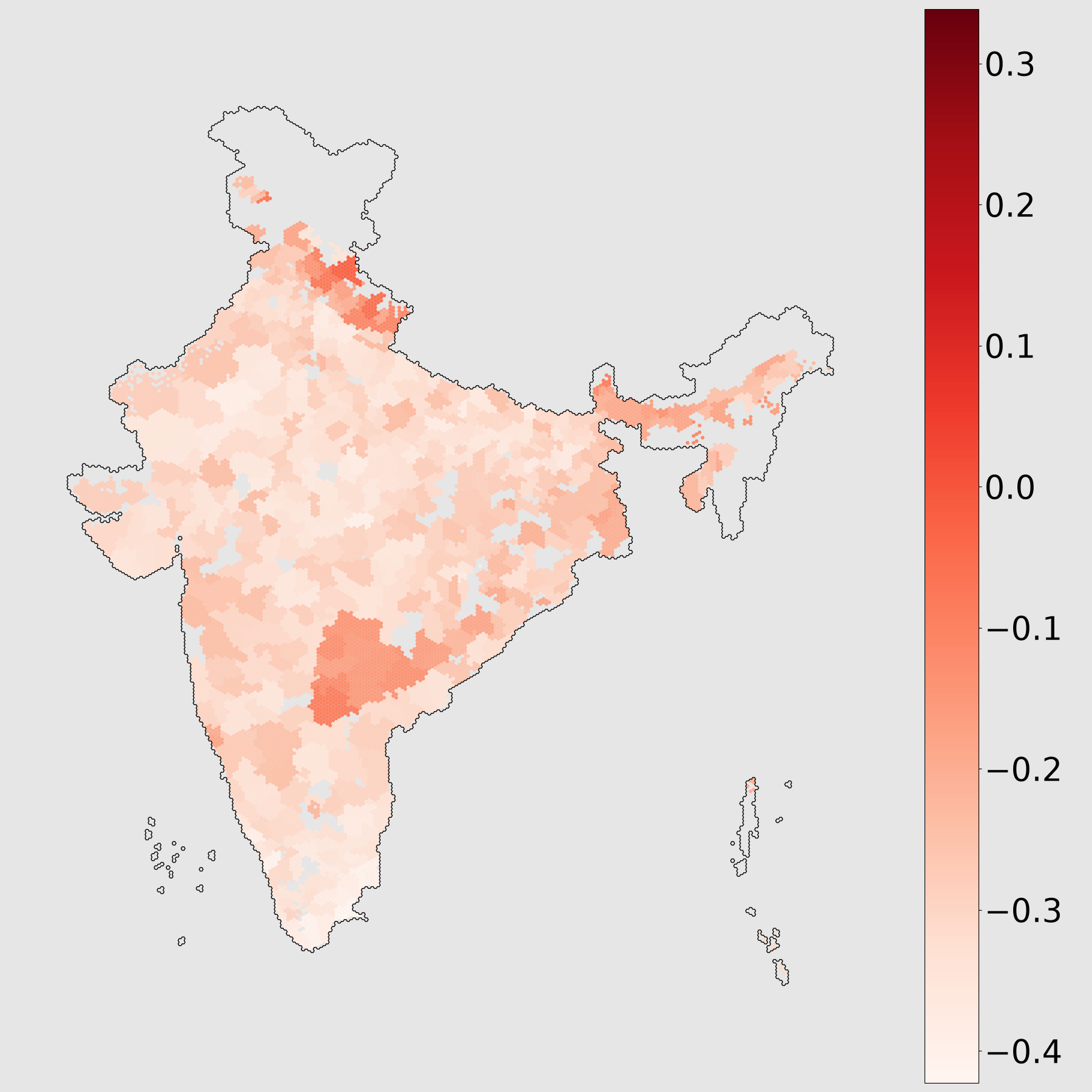}

            \captionsetup{justification=centering}
            \caption{}
        \end{subfigure} 
\begin{subfigure}[!htb]{0.315\linewidth}
            \centering
            \includegraphics[width=\linewidth]{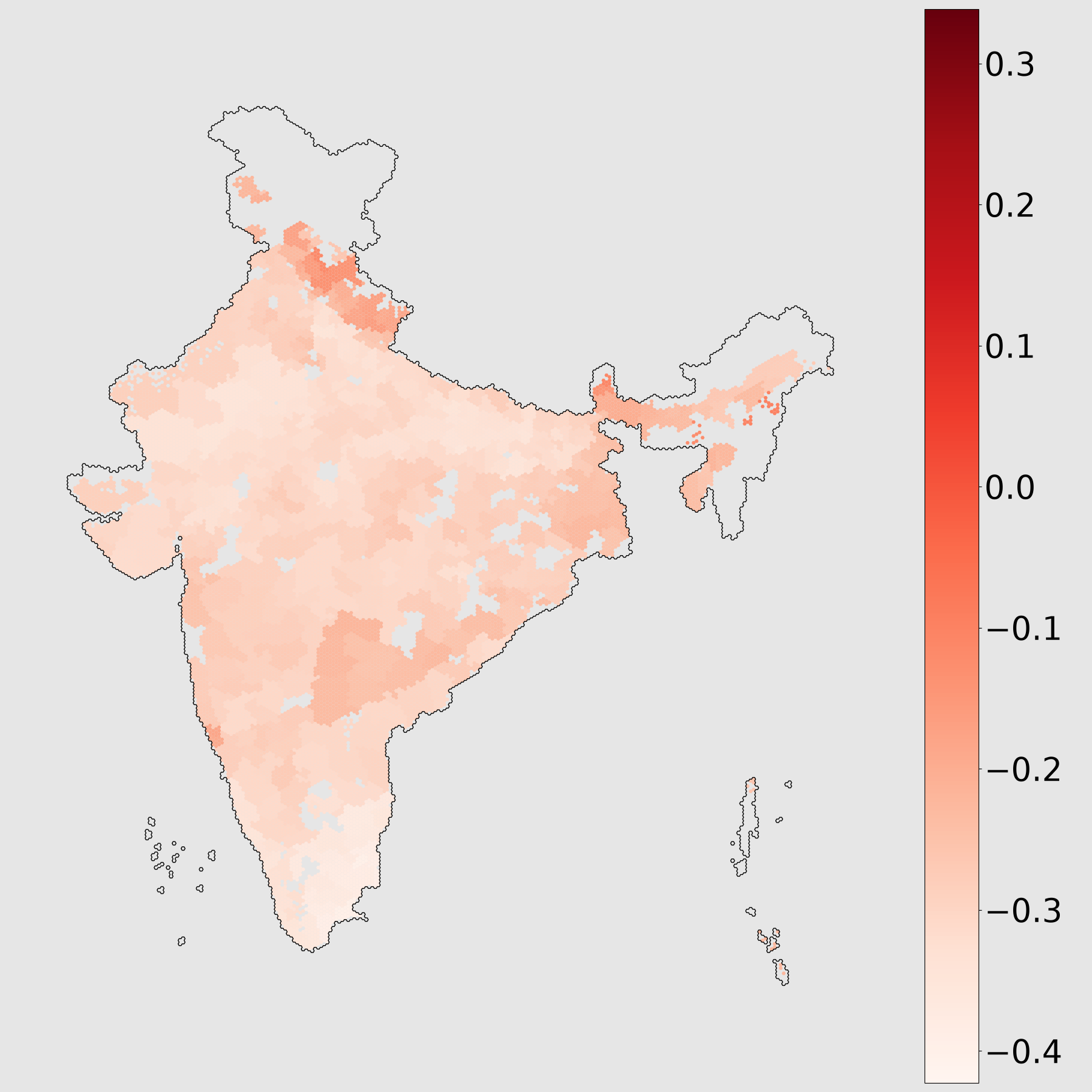}

            \captionsetup{justification=centering}
            \caption{}
        \end{subfigure} 
\begin{subfigure}[!htb]{0.315\linewidth}
            \centering
            \includegraphics[width=\linewidth]{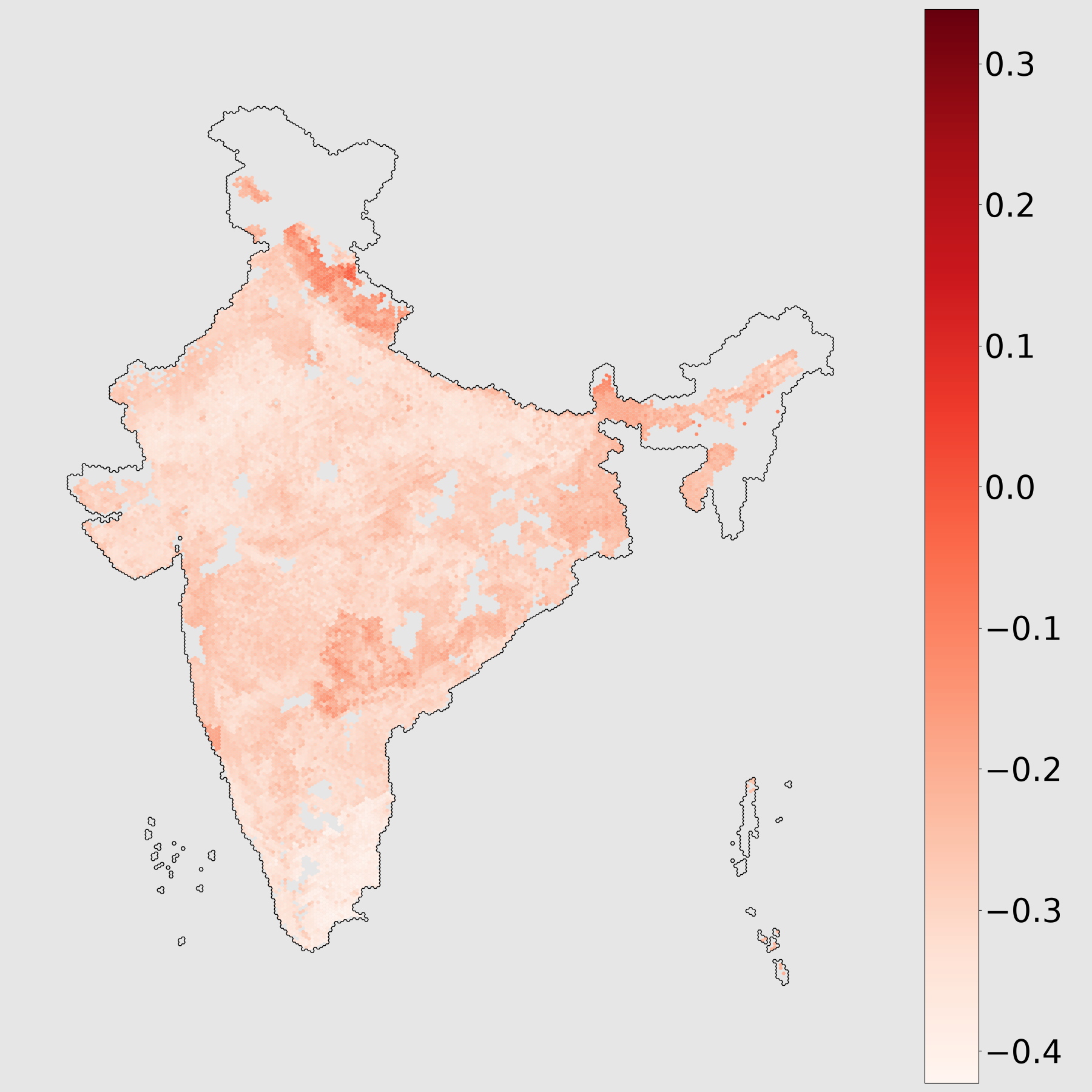}

            \captionsetup{justification=centering}
            \caption{}
        \end{subfigure} \\





        \begin{subfigure}[!htb]{0.315\linewidth}
            \centering
            \includegraphics[width=\linewidth]{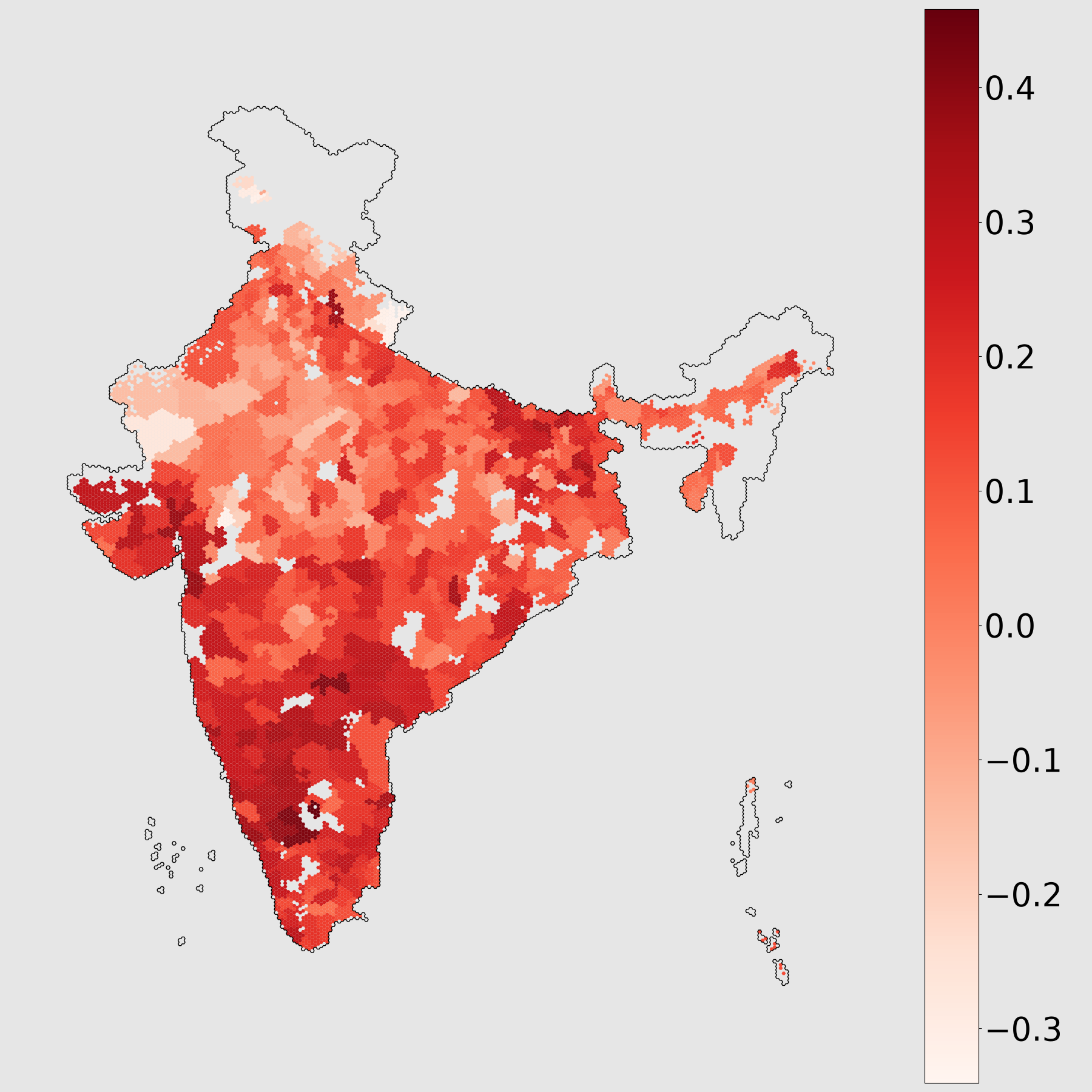}

            \captionsetup{justification=centering}
            \caption{}
        \end{subfigure}
        \begin{subfigure}[!htb]{0.315\linewidth}
            \centering
            \includegraphics[width=\linewidth]{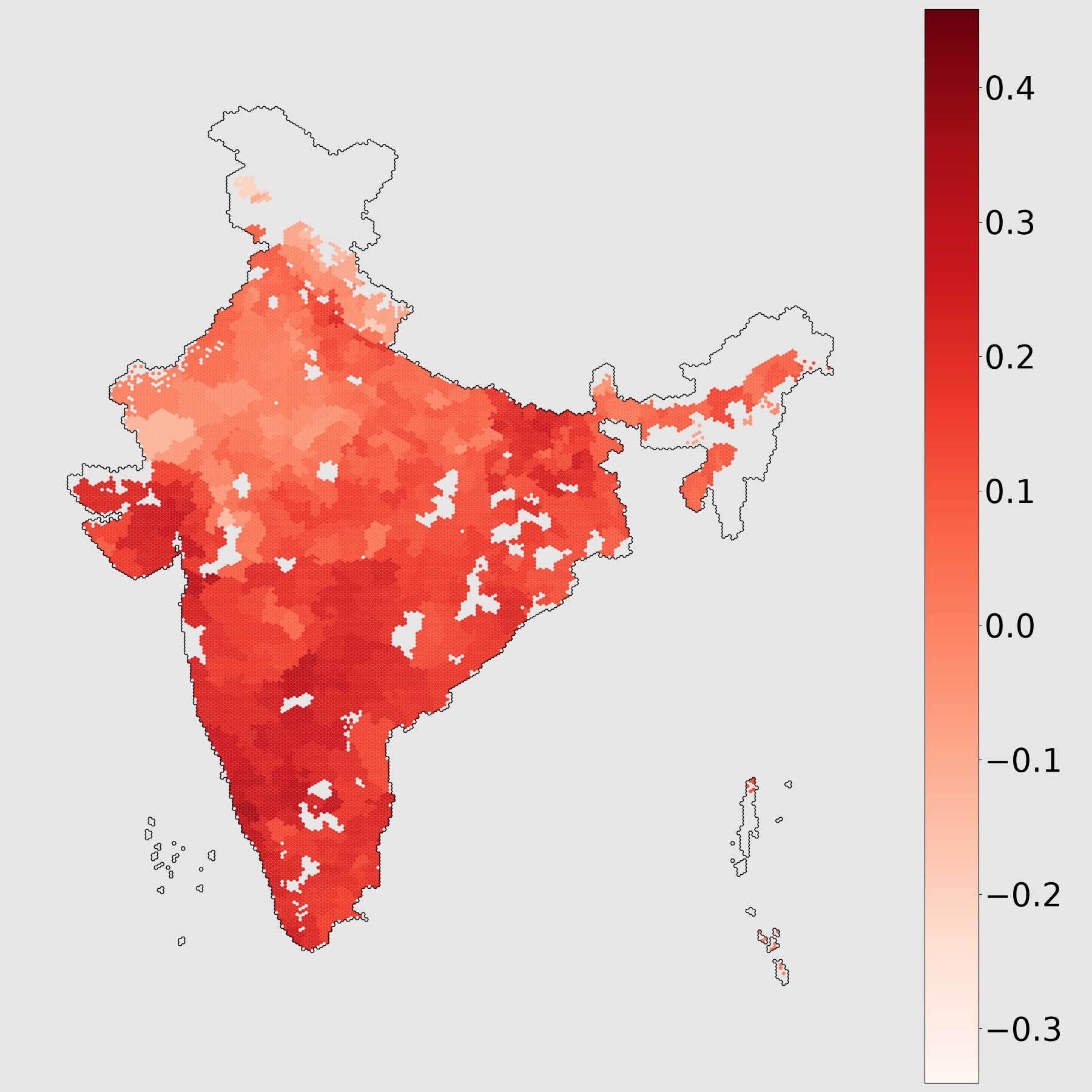}

            \captionsetup{justification=centering}
            \caption{}
        \end{subfigure}
        \begin{subfigure}[!htb]{0.315\linewidth}
            \centering
            \includegraphics[width=\linewidth]{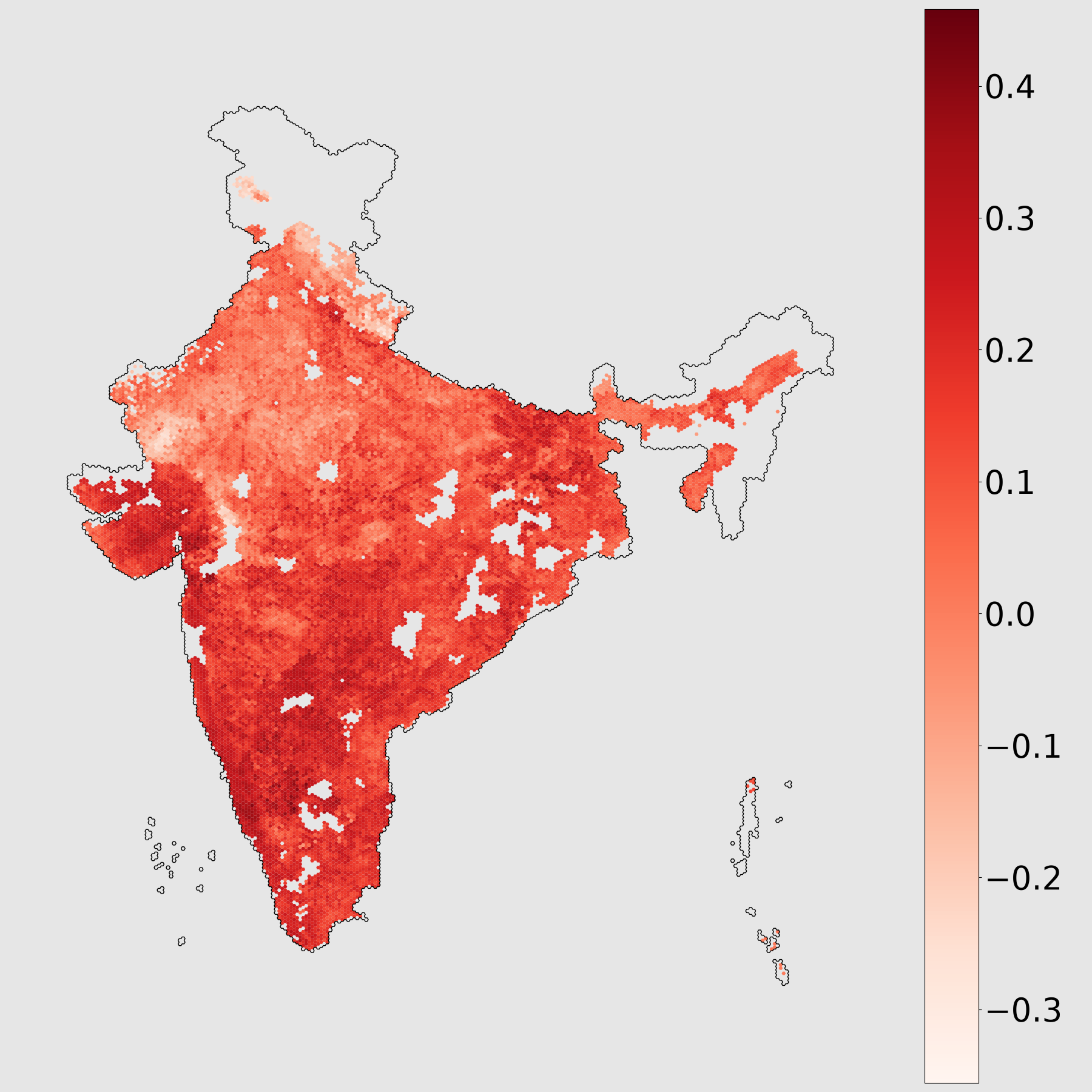}

            \captionsetup{justification=centering}
            \caption{}
        \end{subfigure}
        

        \caption[]{Spatial Coherence of Predicted Latent Socioeconomic Structures (2011). Each row represents one latent dimension of the compressed NSSO space. Columns: Left (a, d, g) show ground-truth district-level values; Center (b, e, h) show predicted cluster-scale values averaged to the district level; Right (c, f, i) show final high-resolution cluster-scale predictions. The visual alignment between the Center and Left columns illustrates the model's qualitative agreement with the ground truth, while the Right column reveals plausible fine-grained, intra-district variations absent in the coarse input.}
        \label{fig:enc_map_2011}
    
    \end{figure}

This dual visualization serves two purposes: first, it allows for a qualitative comparison between the district-averaged predictions and the corresponding ground-truth encodings; second, it enables an assessment of the spatial coherence between the fine-grained and coarse-grained predictions. Figures \ref{fig:enc_map_2001} and \ref{fig:enc_map_2011} display these maps for the years 2001 and 2011, respectively, illustrating the patterns with representative latent vectors from the Consumer Expenditure, Debt and Investment, and Housing Conditions categories. \par

A key characteristic of the autoencoder's latent space is that it represents a complex, nonlinear combination of the original input indicators. A direct consequence of this is that a singular, human-interpretable meaning—such as `income' or `expenditure'—cannot be ascribed to any individual latent dimension. However, despite this lack of direct semantic interpretability, a qualitative inspection of the mapped dimensions reveals that some components do appear to encode distinct geographic signatures. These often manifest as an emphasis on particular regional patterns, such as concentrations in eastern or southern districts, alongside other more complex, mixed spatial signals.\par

A central finding of this spatial analysis is that the high-resolution cluster-level predictions exhibit a strong spatial coherence with the district-level ground-truth data. The overall geographic patterns observed in the predicted maps correspond closely to those in the original district data, while simultaneously revealing plausible, fine-grained variations within each district that are absent in the coarse summaries.

This alignment is confirmed both qualitatively through visual inspection and quantitatively by aggregating the cluster-scale predictions to the district scale for direct comparison. The model's demonstrated ability to preserve both broad regional trends and localized patterns indicates that the latent encodings—despite their lack of direct semantic interpretability—successfully capture meaningful socioeconomic structures.\par

The predicted high-resolution maps effectively capture the temporal evolution of socioeconomic indicators between 2001 and 2011, revealing significant shifts in their spatial distributions over the decade. The model successfully resolves these year-to-year dynamics while simultaneously preserving the broader, underlying regional patterns, demonstrating a capacity to model both change and stability. \par

\subsubsection{Decoded (full-scale) variables visualization}
Following the analysis of the learned latent representations, the examination now turns to the final, full-scale indicators produced by the decoder. These decoded variables represent the model's best estimate of the original NSSO features and provide the basis for the subsequent quantitative error analysis. \par

 \begin{figure}[!htb]
        \centering

        \begin{subfigure}[!htb]{0.325\linewidth}
            \centering
            \includegraphics[width=\linewidth]{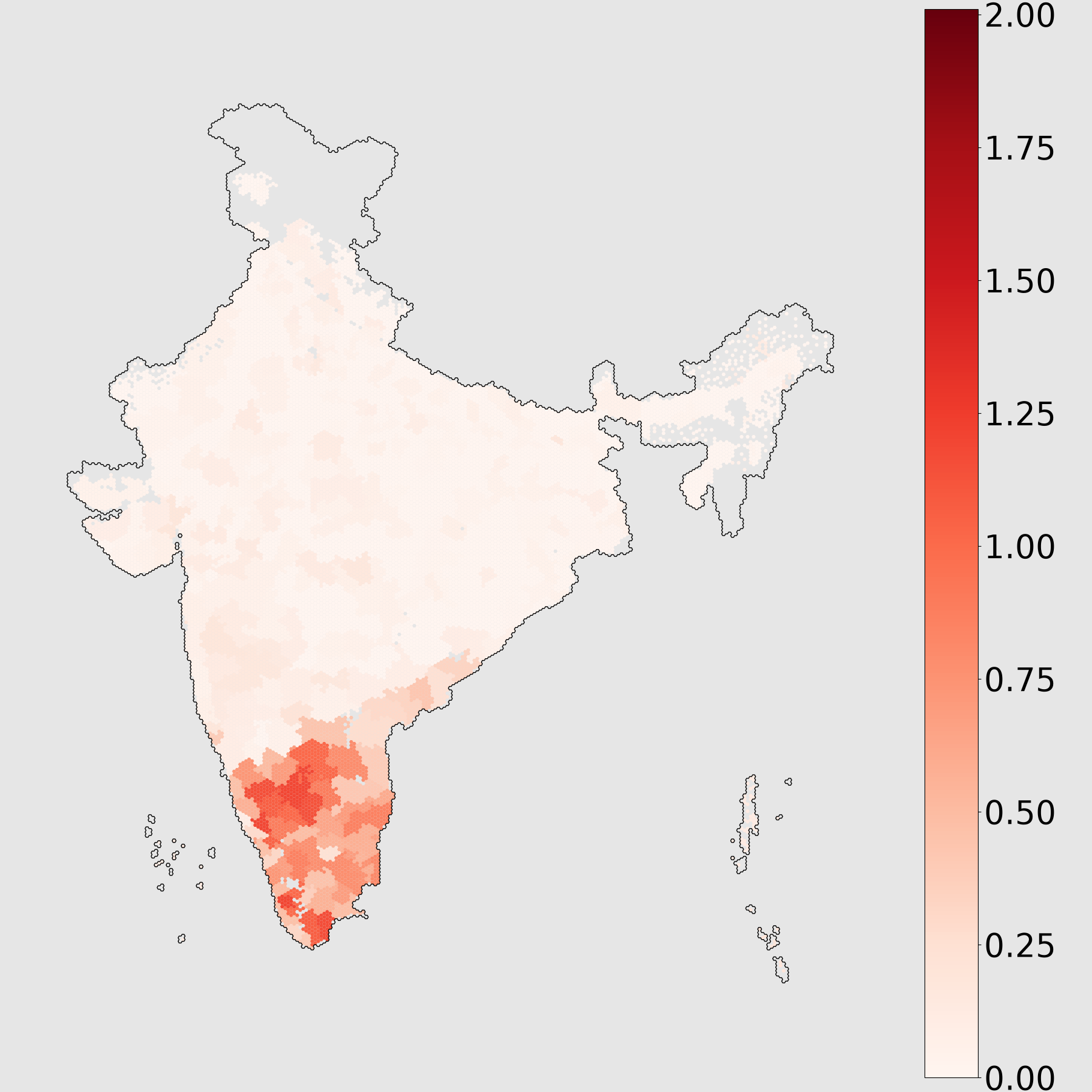}
            \captionsetup{justification=centering}
            \caption{}
            \label{fig:coffee_org_2001}
        \end{subfigure} 
        \begin{subfigure}[!htb]{0.325\linewidth}
            \centering
            \includegraphics[width=\linewidth]{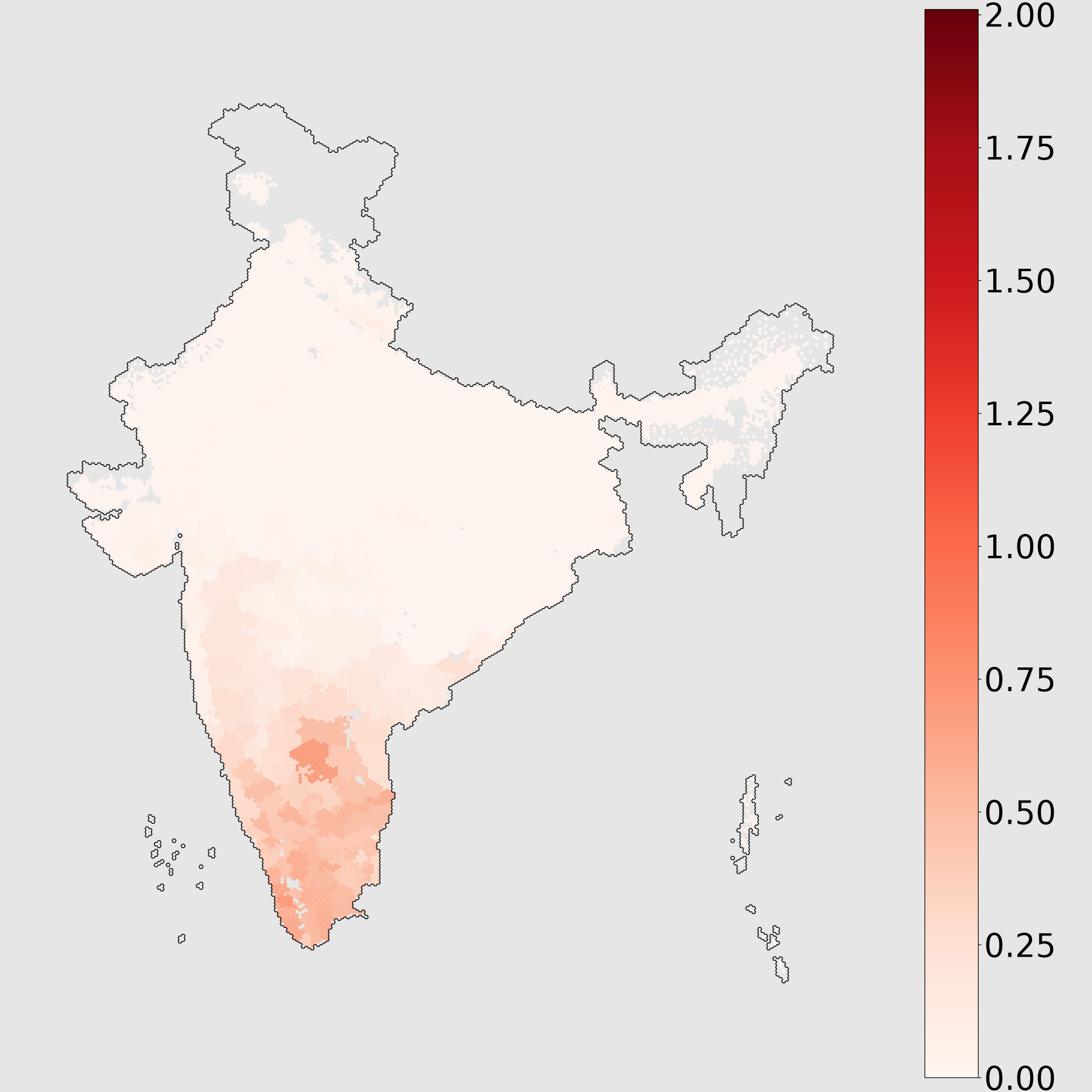}
            \captionsetup{justification=centering}
            \caption{}
            \label{fig:coffee_agg_2001}
        \end{subfigure} 
        \begin{subfigure}[!htb]{0.325\linewidth}
            \centering
            \includegraphics[width=\linewidth]{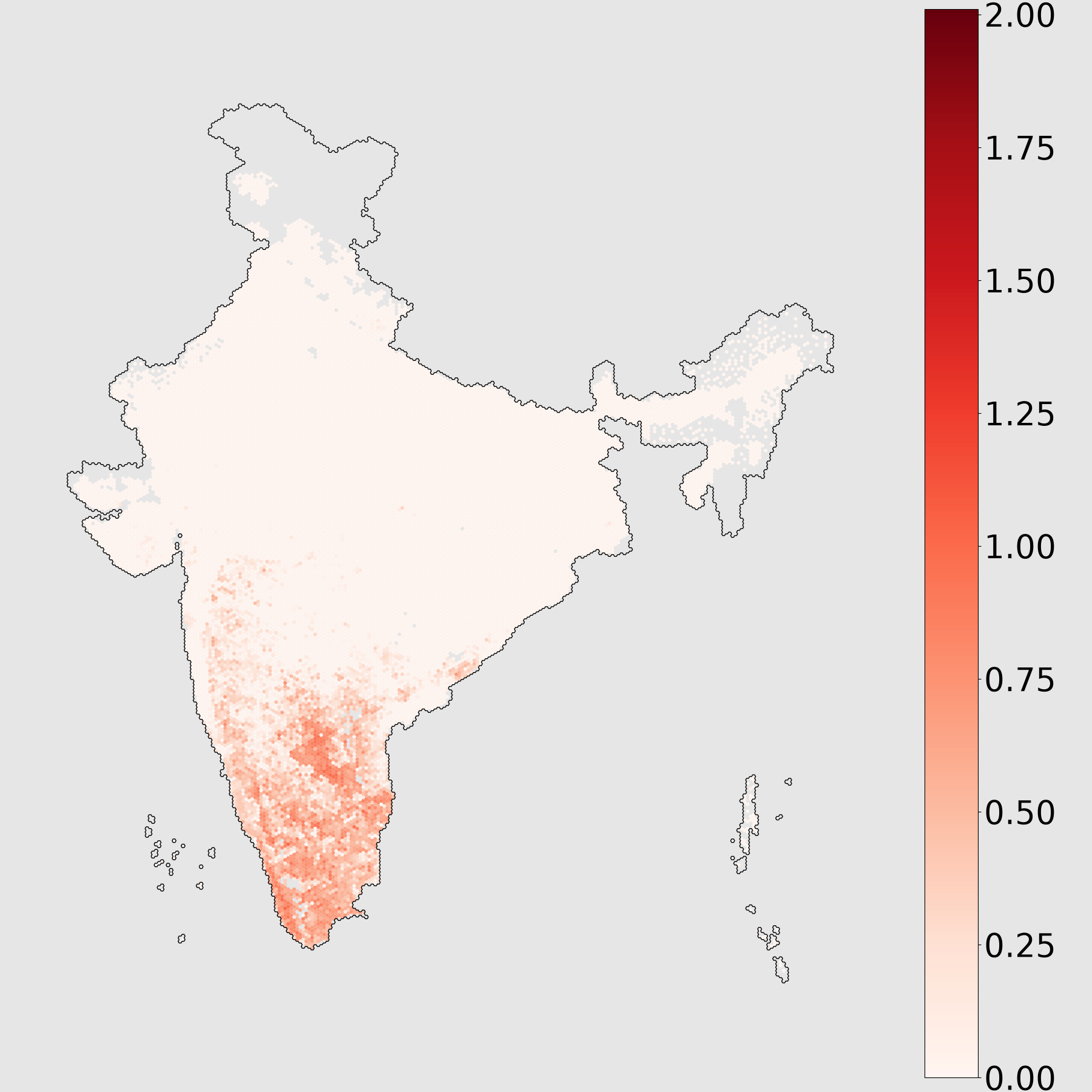}
            \captionsetup{justification=centering}
            \caption{}
            \label{fig:coffee_hex_2001}
        \end{subfigure} \\

        \begin{subfigure}[!htb]{0.325\linewidth}
            \centering
            \includegraphics[width=\linewidth]{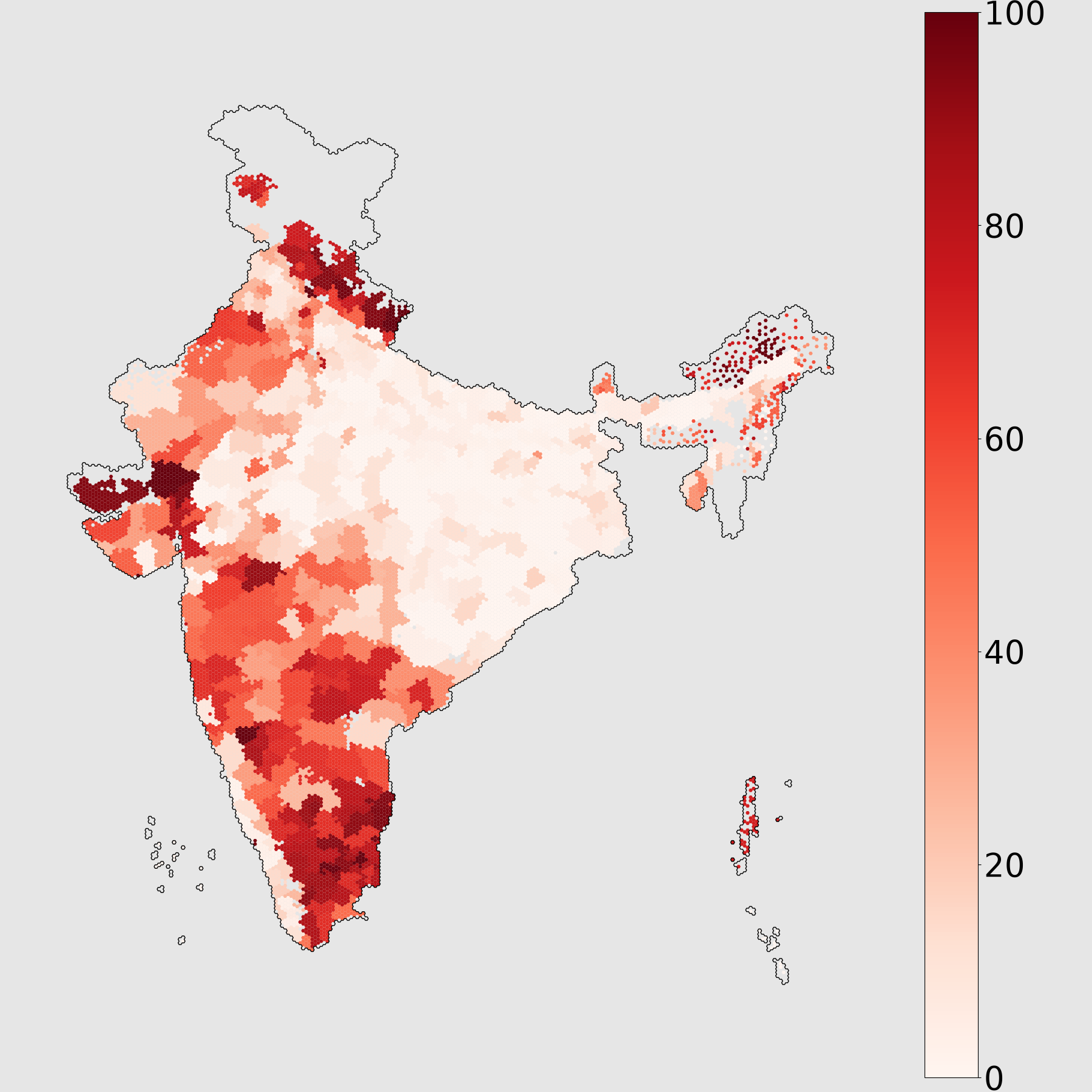}

            \captionsetup{justification=centering}
            \caption{}
            \label{fig:tap_org_2001}
        \end{subfigure} 
\begin{subfigure}[!htb]{0.325\linewidth}
            \centering
            \includegraphics[width=\linewidth]{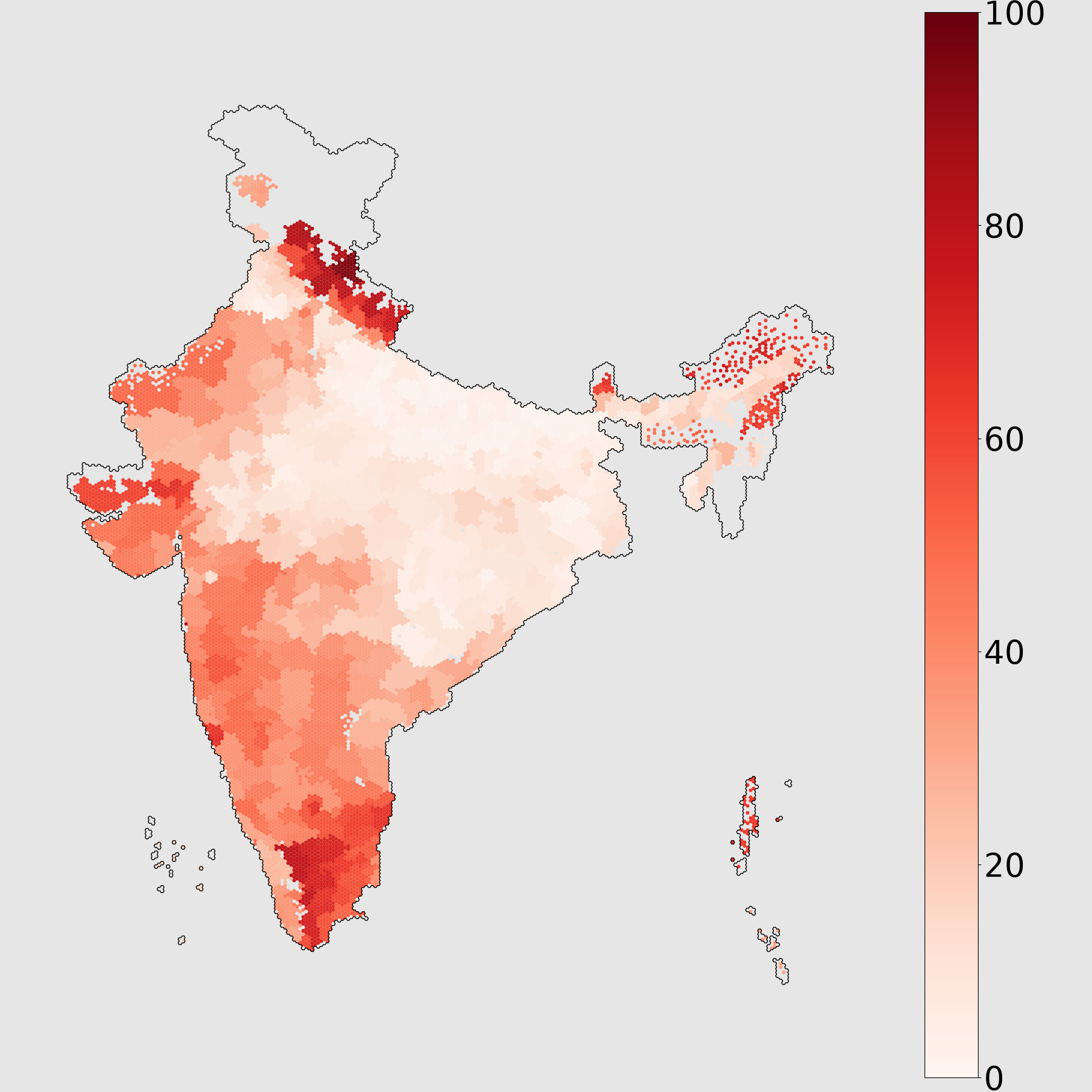}

            \captionsetup{justification=centering}
            \caption{}
            \label{fig:tap_agg_2001}
        \end{subfigure} 
\begin{subfigure}[!htb]{0.325\linewidth}
            \centering
            \includegraphics[width=\linewidth]{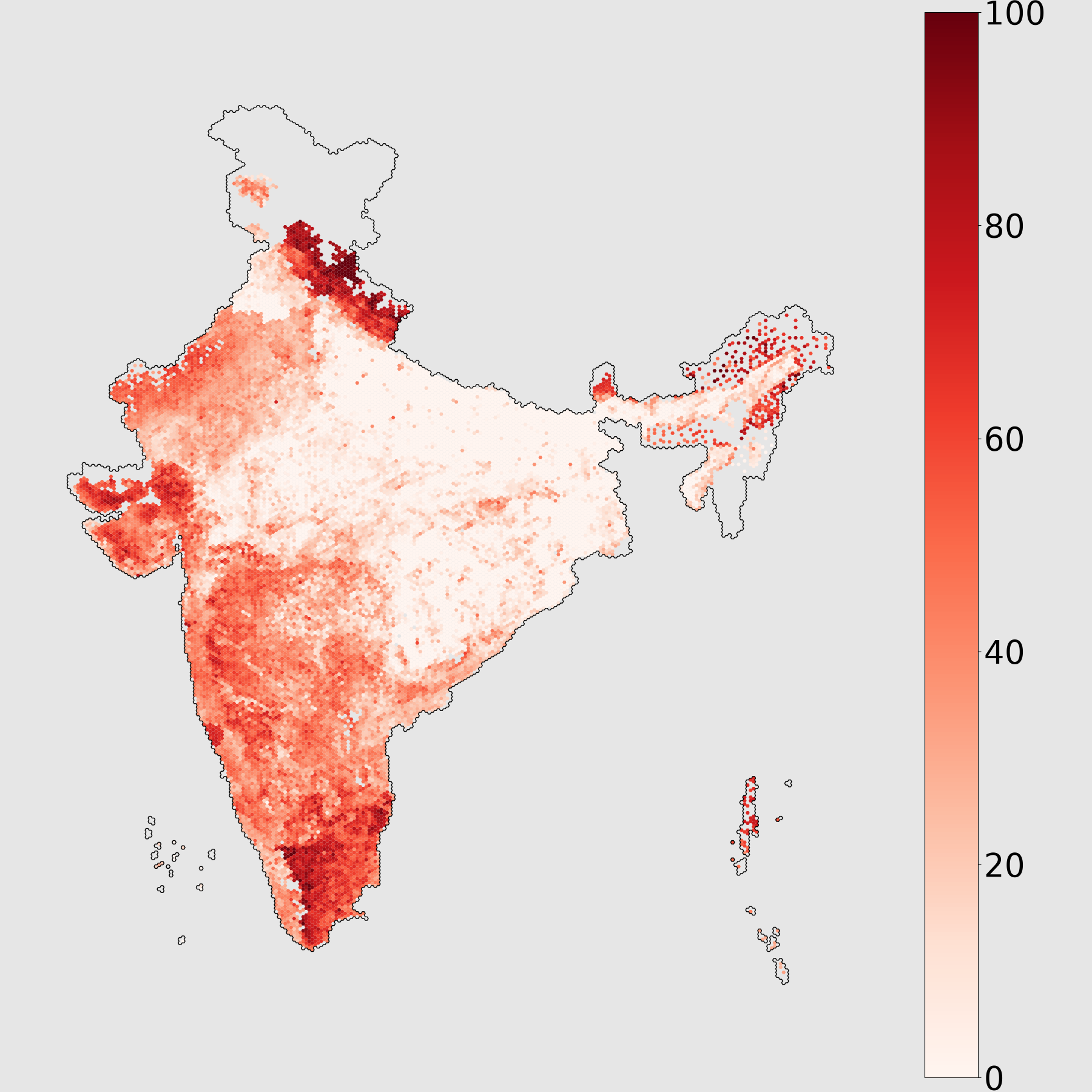}

            \captionsetup{justification=centering}
            \caption{}

            \label{fig:tap_hex_2001}
        \end{subfigure} \\

        \begin{subfigure}[!htb]{0.325\linewidth}
            \centering
            \includegraphics[width=\linewidth]{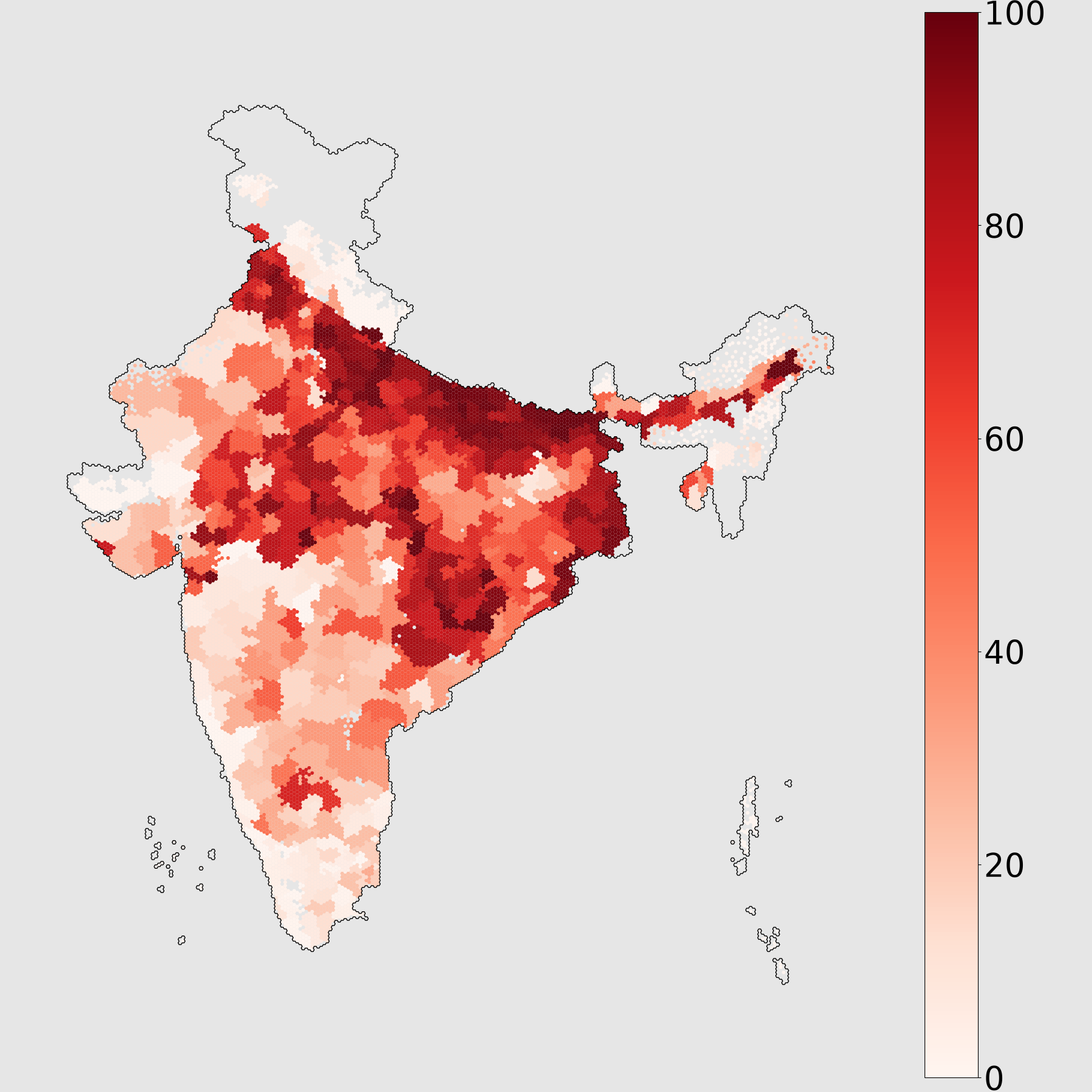}

            \captionsetup{justification=centering}
            \caption{}
            \label{fig:tube_org_2001}
        \end{subfigure}
\begin{subfigure}[!htb]{0.325\linewidth}
            \centering
            \includegraphics[width=\linewidth]{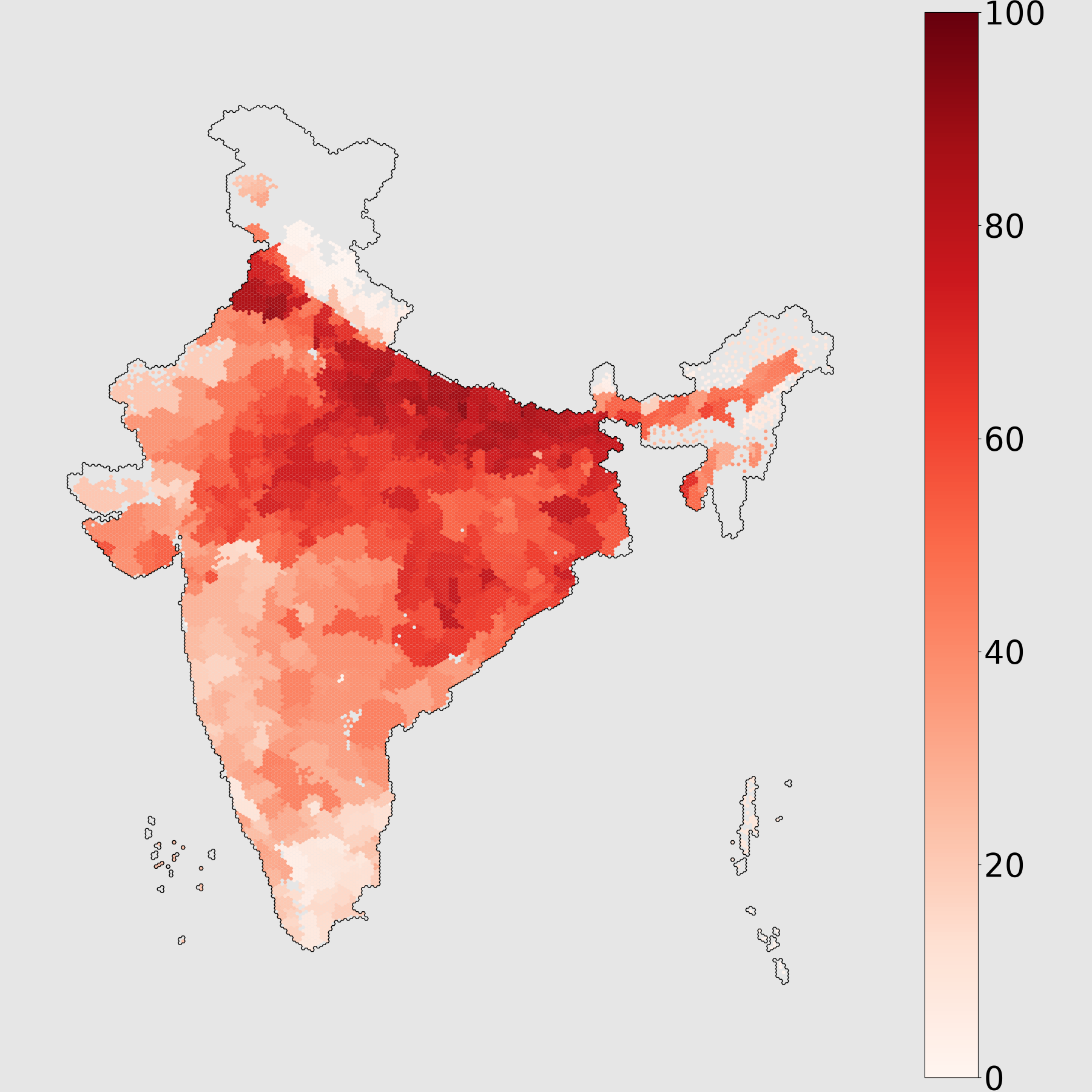}

            \captionsetup{justification=centering}
            \caption{}

            \label{fig:tube_agg_2001}
        \end{subfigure}
\begin{subfigure}[!htb]{0.325\linewidth}
            \centering
            \includegraphics[width=\linewidth]{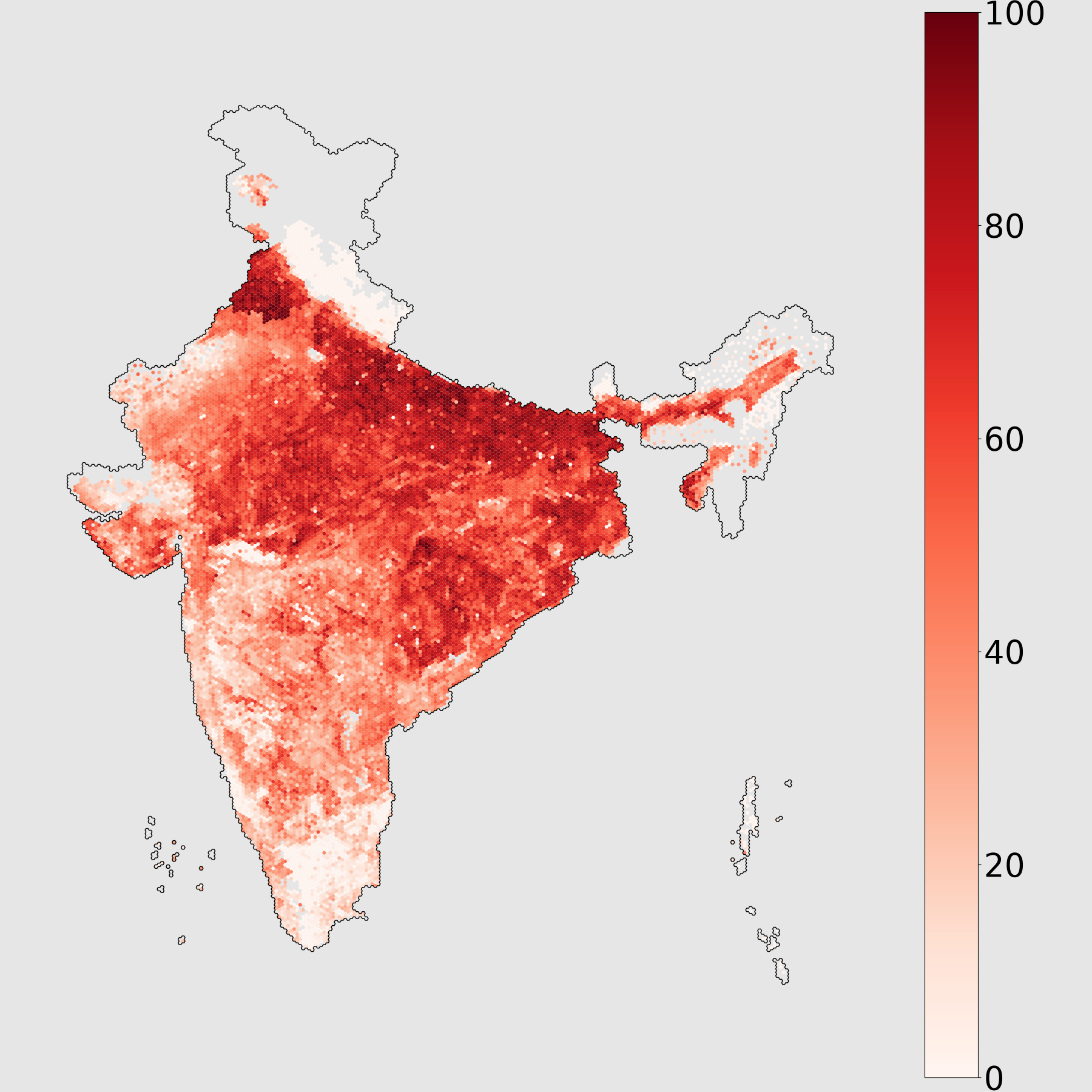}

            \captionsetup{justification=centering}
            \caption{}

            \label{fig:tube_hex_2001}
        \end{subfigure}


        \caption[]{Visualization of Decoded Full-Scale Socioeconomic Indicators (2001). The figure highlights the model's ability to reproduce broad spatial patterns while capturing fine-scale detail. Rows: (a–c) Coffee Expenditure, concentrated in southern states; (d–f) Tap Water Access, showing regional infrastructure gaps; and (g–i) Tube Well Usage, inversely related to tap water, indicating groundwater reliance in northern regions. Columns: Left—ground-truth district values; Center—predicted cluster-scale values averaged to districts; Right—final high-resolution predictions.}
        \label{fig:dec_map_2001}
    
    \end{figure}

An examination of the decoded variables reveals that their spatial distributions often correspond to established regional characteristics. A clear example is the expenditure on coffee powder, as depicted in the original district-level data (Fig. \ref{fig:coffee_org_2001}). The map indicates that expenditure is highest in the southern states of Kerala, Tamil Nadu, and Karnataka, a pattern consistent with their status as India's primary coffee-producing and consuming regions due to both cultural and agricultural factors.
A visual comparison with the model's high-resolution cluster-level prediction (Fig. \ref{fig:coffee_hex_2001}) and the aggregated district-level prediction (Fig. \ref{fig:coffee_agg_2001}) confirms that our framework successfully reproduces this broad geographic trend. However, while the primary regional pattern is preserved, a loss of some finer-scale detail is evident in the aggregated map. This attenuation is likely attributable to factors such as the limited number of training samples and the inherent uncertainty in the source NSSO data, which required imputation for missing indicators. Nevertheless, the strong overall correspondence demonstrates the promise of this approach and provides a foundation for future methodological refinements. \par

\begin{figure}[!htb]
        \centering

    
        \begin{subfigure}[!htb]{0.325\linewidth}
            \centering
            \includegraphics[width=\linewidth]{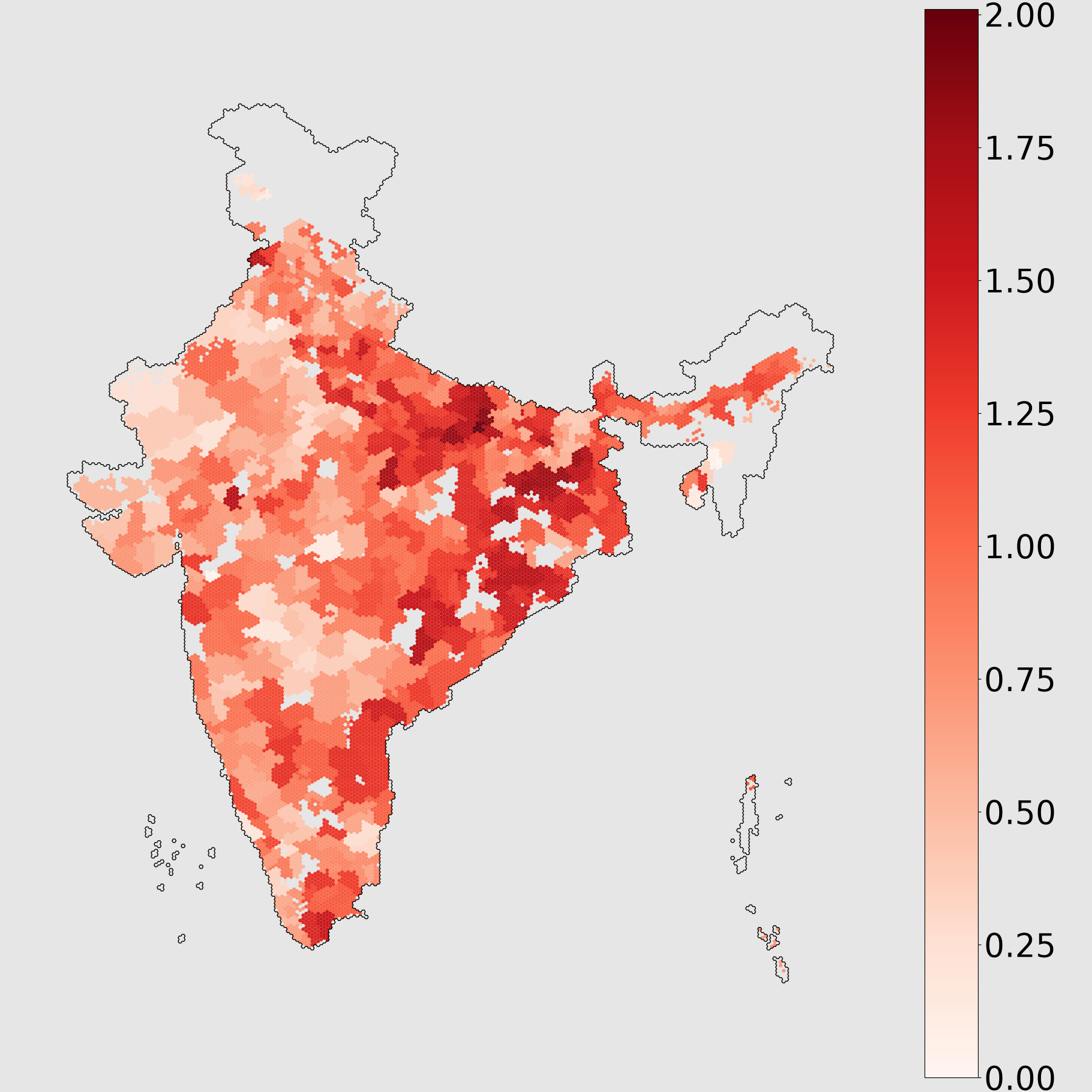}
            \captionsetup{justification=centering}
            \caption{}
            \label{fig:coffee_org_2011}

        \end{subfigure} 
        \begin{subfigure}[!htb]{0.325\linewidth}
            \centering
            \includegraphics[width=\linewidth]{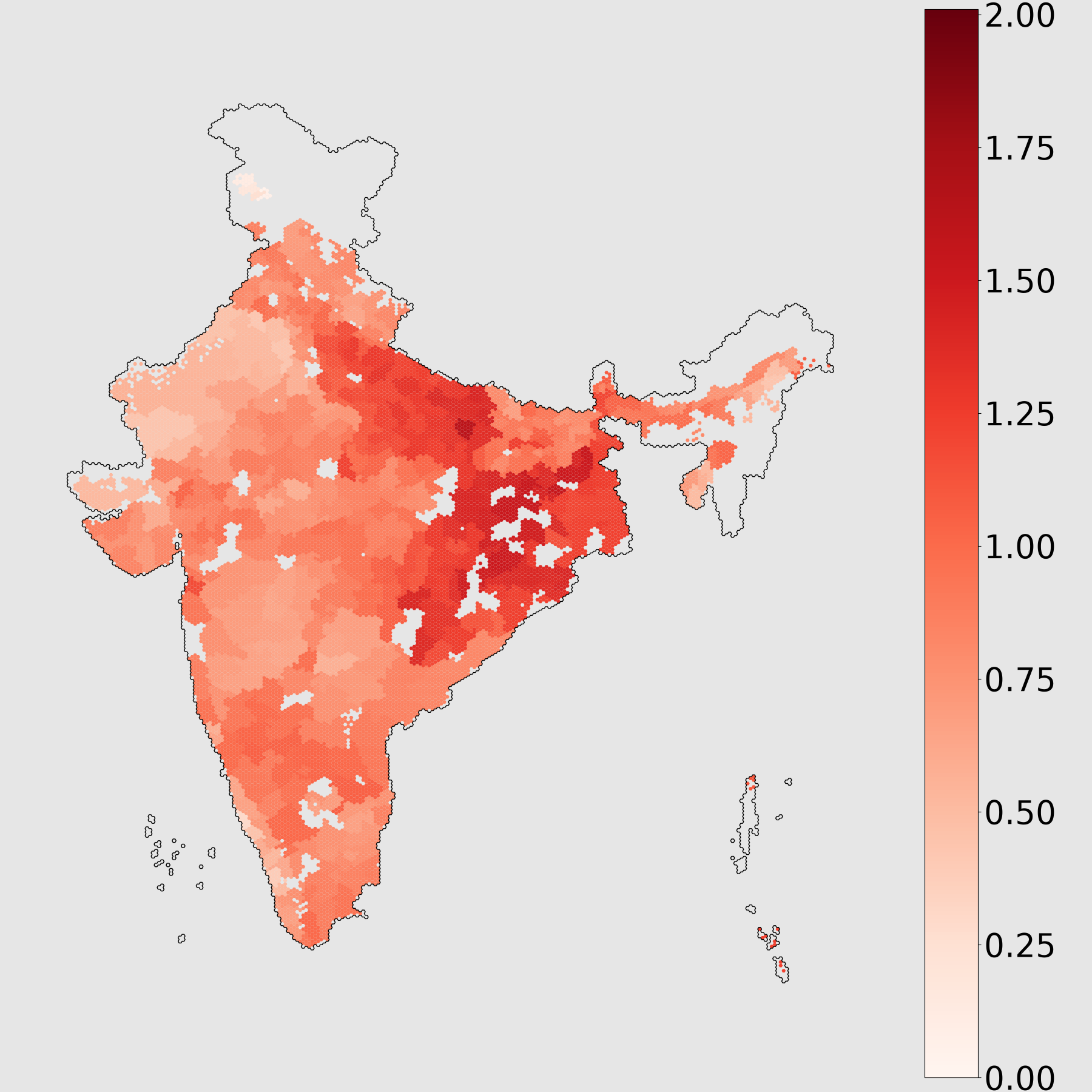}
            \captionsetup{justification=centering}
            \caption{}
            \label{fig:coffee_agg_2011}

        \end{subfigure} 
        \begin{subfigure}[!htb]{0.325\linewidth}
            \centering
            \includegraphics[width=\linewidth]{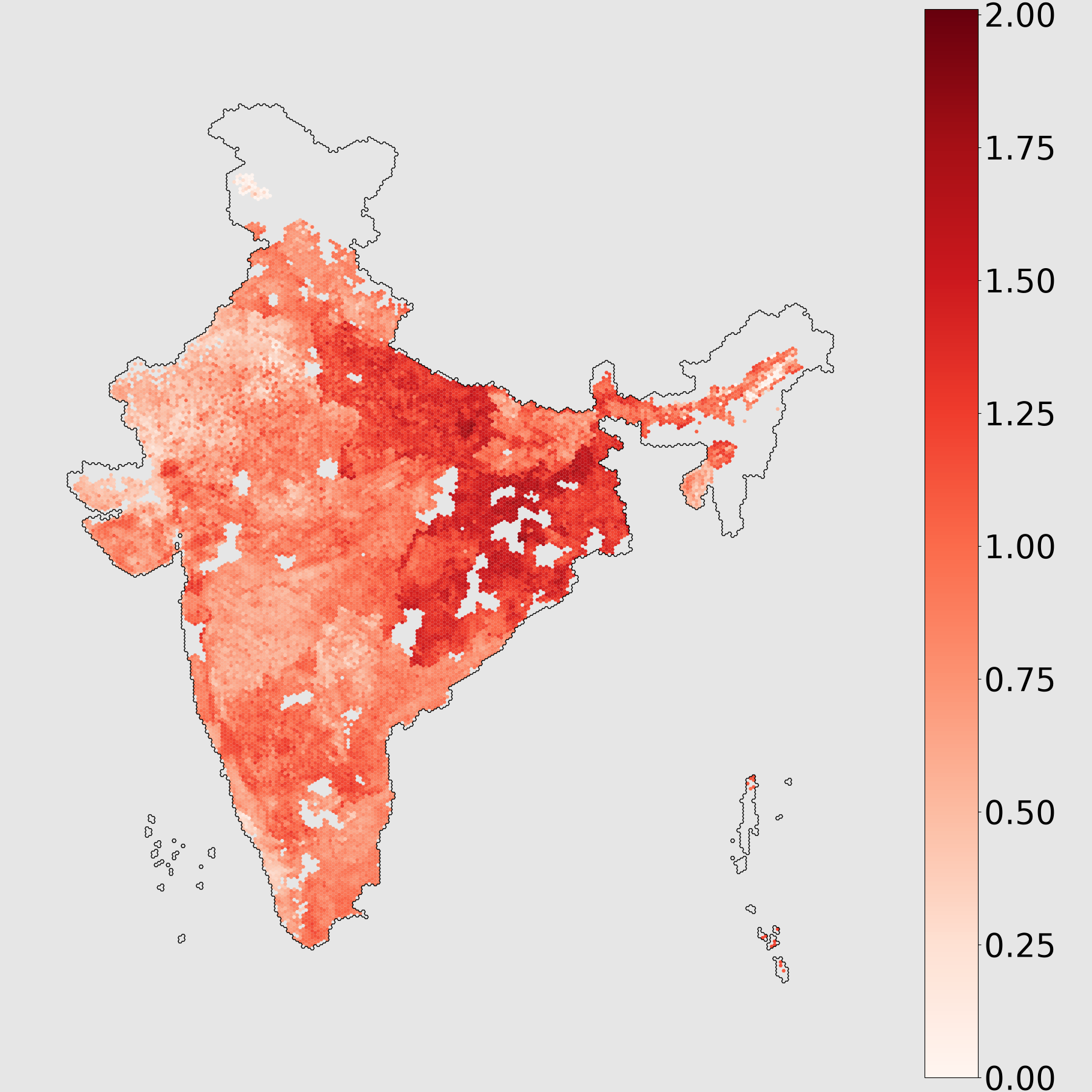}
            \captionsetup{justification=centering}
            \caption{}
            \label{fig:coffee_hex_2011}

        \end{subfigure} \\

        \begin{subfigure}[!htb]{0.325\linewidth}
            \centering
            \includegraphics[width=\linewidth]{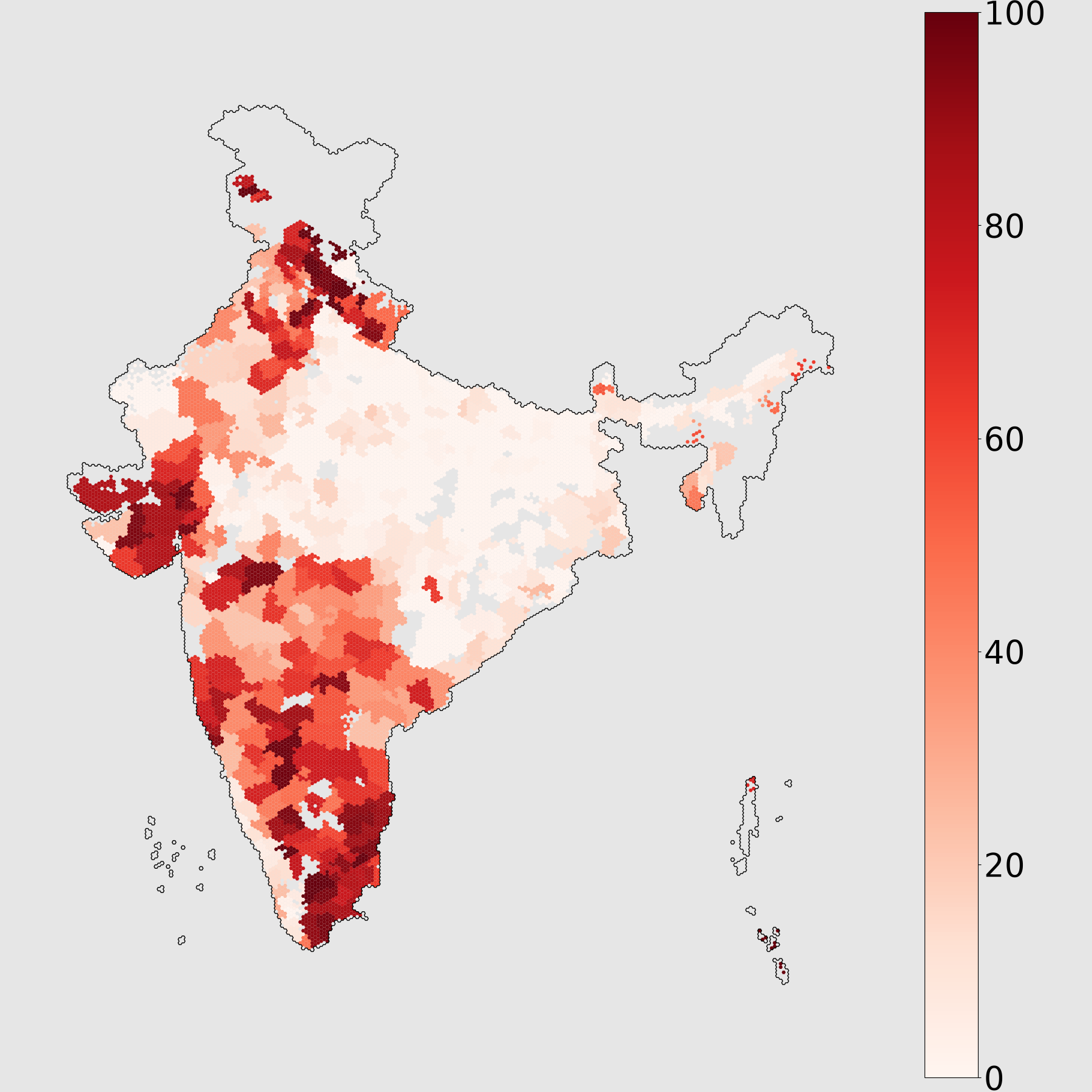}

            \captionsetup{justification=centering}
            \caption{}
            \label{fig:tap_org_2011}

        \end{subfigure} 
\begin{subfigure}[!htb]{0.325\linewidth}
            \centering
            \includegraphics[width=\linewidth]{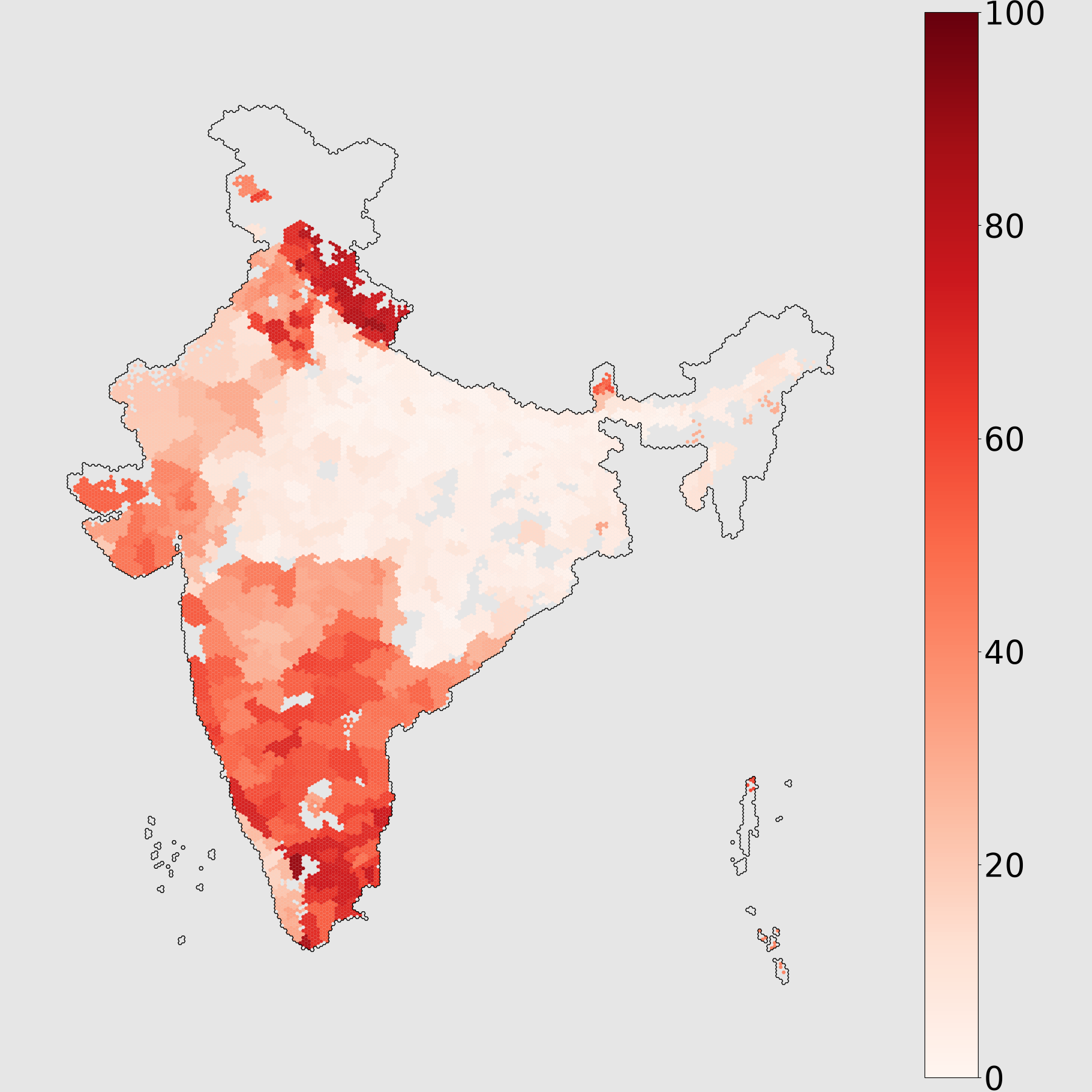}

            \captionsetup{justification=centering}
            \caption{}
            \label{fig:tap_agg_2011}

        \end{subfigure} 
\begin{subfigure}[!htb]{0.325\linewidth}
            \centering
            \includegraphics[width=\linewidth]{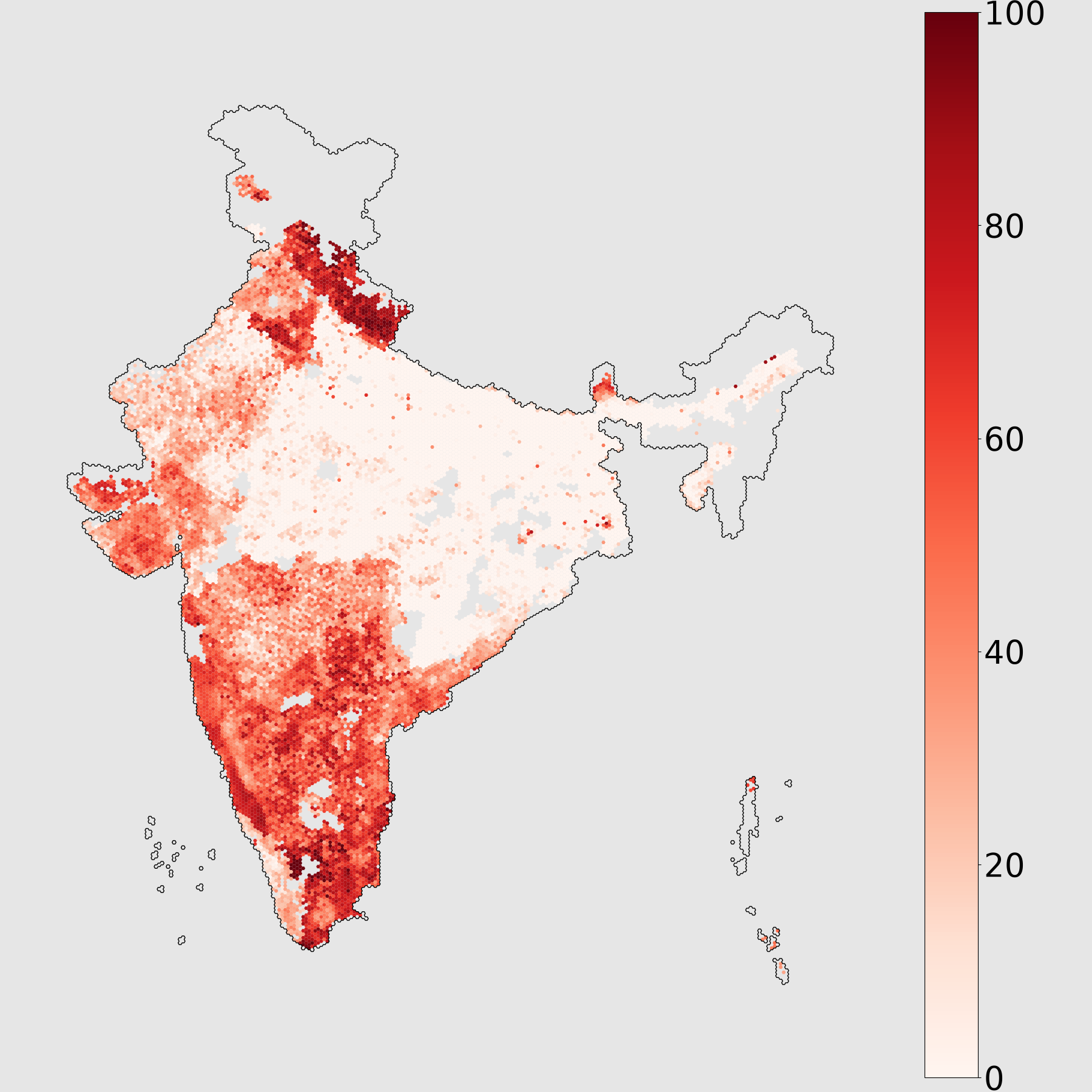}

            \captionsetup{justification=centering}
            \caption{}
            \label{fig:tap_hex_2011}

        \end{subfigure} \\

        \begin{subfigure}[!htb]{0.325\linewidth}
            \centering
            \includegraphics[width=\linewidth]{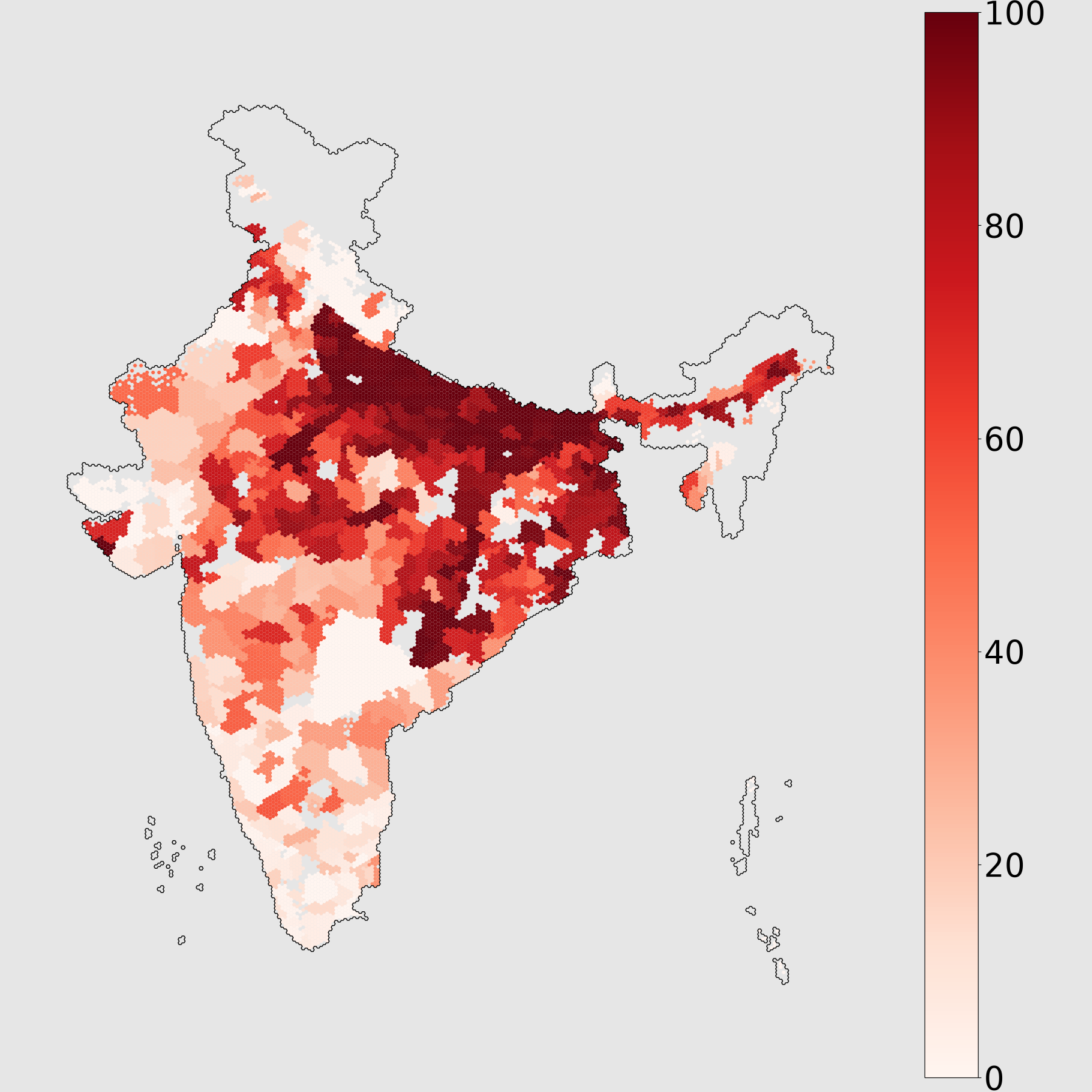}

            \captionsetup{justification=centering}
            \caption{}
            \label{fig:tube_org_2011}

        \end{subfigure}
\begin{subfigure}[!htb]{0.325\linewidth}
            \centering
            \includegraphics[width=\linewidth]{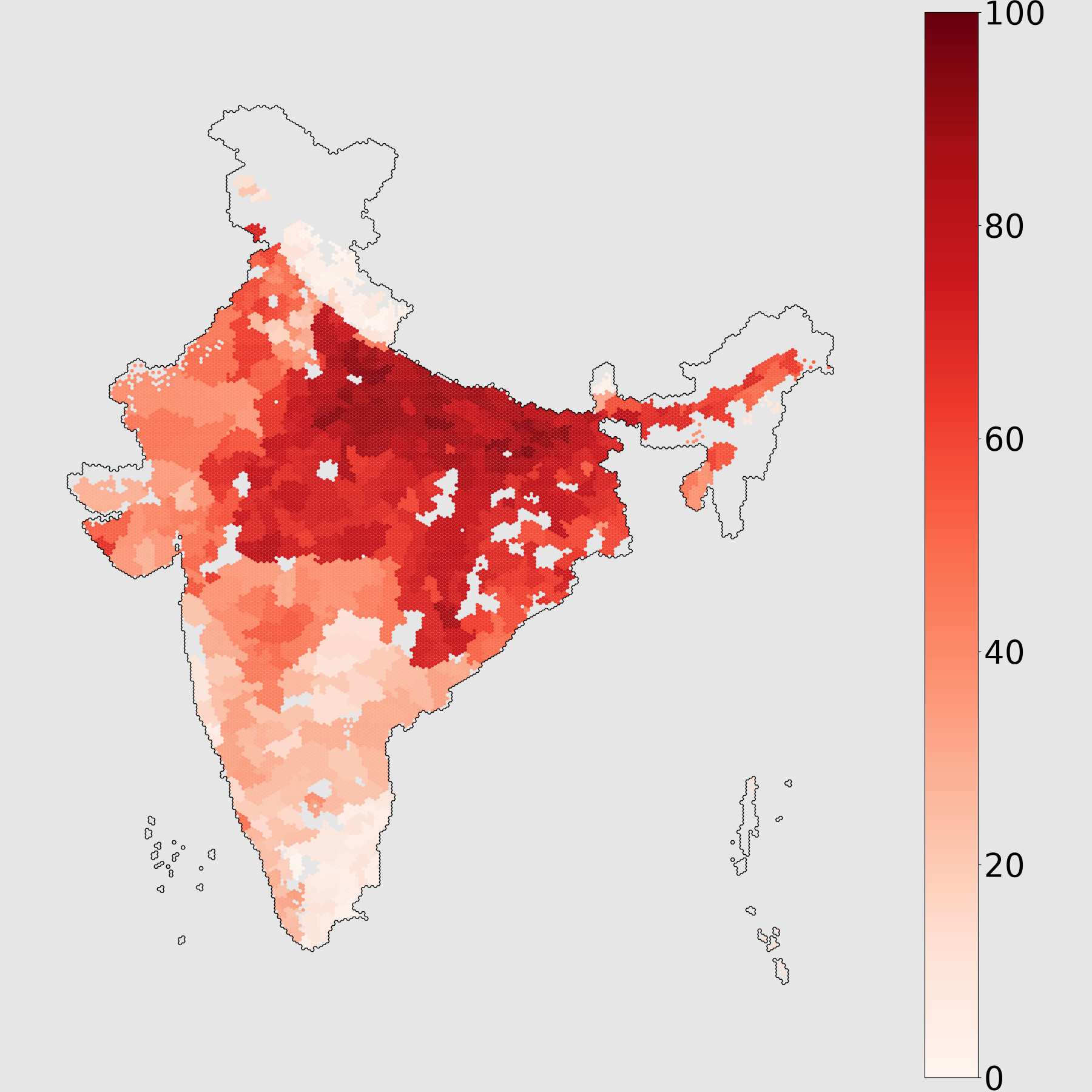}

            \captionsetup{justification=centering}
            \caption{}
            \label{fig:tube_agg_2011}

        \end{subfigure}
\begin{subfigure}[!htb]{0.325\linewidth}
            \centering
            \includegraphics[width=\linewidth]{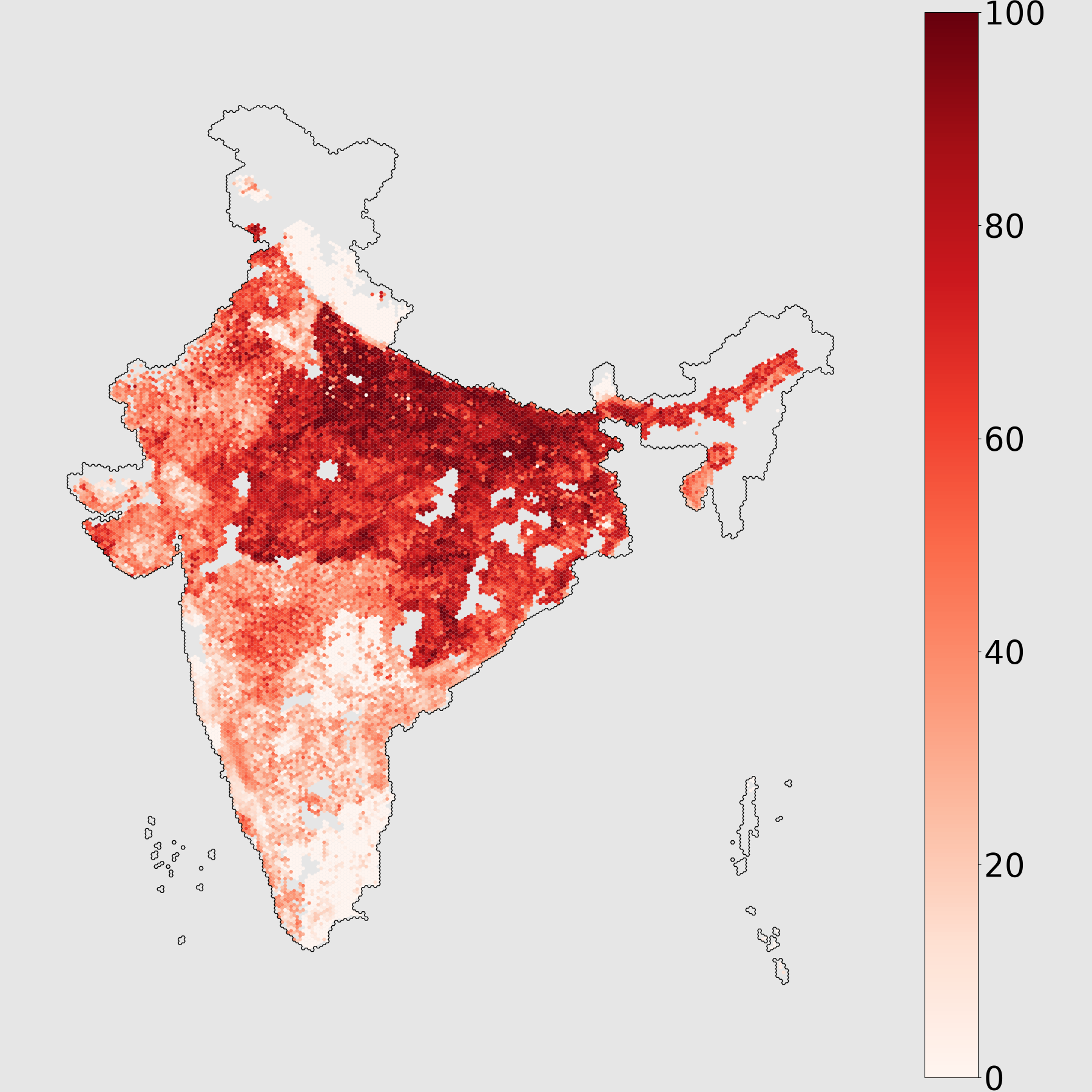}

            \captionsetup{justification=centering}
            \caption{}
            \label{fig:tube_hex_2011}

        \end{subfigure}


        \caption[]{Visualization of Decoded Full-Scale Socioeconomic Indicators (2011). The figure highlights the model's ability to reproduce broad spatial patterns while capturing fine-scale detail. Rows: (a–c) Coffee Expenditure, concentrated in southern states; (d–f) Tap Water Access, showing regional infrastructure gaps; and (g–i) Tube Well Usage, inversely related to tap water, indicating groundwater reliance in northern regions. Columns: Left—ground-truth district values; Center—predicted cluster-scale values averaged to districts; Right—final high-resolution predictions.}
        \label{fig:dec_map_2011}
    
    \end{figure}

The model's predictions for 2011 capture a significant temporal shift in coffee consumption patterns. As illustrated in the corresponding map (Fig. \ref{fig:dec_map_2011}), expenditure on coffee had undergone a notable geographic diffusion by 2011. In contrast to the pattern in 2001, where consumption was heavily concentrated in the southern states, the 2011 map reveals an expansion into other parts of the country, suggesting a broader cultural adoption of coffee as a beverage.

A second set of examples is provided by indicators related to drinking water sources, which exhibit distinct and often inverse regional patterns. For instance, the proportion of households relying on tube wells is notably high in northern states such as Uttar Pradesh, a reflection of widespread groundwater extraction practices in the region (Fig. \ref{fig:tube_org_2001}). Conversely, the use of tap water as a principal source is more prevalent in other parts of the country, exhibiting a largely inverse spatial distribution (Fig. \ref{fig:tap_org_2001}).
The model's cluster-level and aggregated predictions for both the tube well indicator (Figs. \ref{fig:tube_hex_2001}, \ref{fig:tube_agg_2001}) and the tap water indicator (Figs. \ref{fig:tap_hex_2001}, \ref{fig:tap_agg_2001}) successfully reproduce these opposing spatial trends. Crucially, the high-resolution cluster-scale maps for both indicators reveal plausible fine-scale variations within individual districts, providing a more granular view of local water infrastructure patterns than is available in the original source data.\par

The temporal analysis of drinking water sources between 2001 and 2011 reveals divergent yet complementary trends. The 2011 map for tap water as the principal drinking source (Fig. \ref{fig:dec_map_2011}) shows a marked increase in usage across numerous districts, which is indicative of an expansion of piped water infrastructure into urban and semi-urban areas.

Concurrently, tube well usage also intensified in certain regions, notably in the northern state of Uttar Pradesh, suggesting a continued and potentially growing reliance on local groundwater sources where formal supply may be insufficient. In many areas, the spatial patterns of these two indicators exhibit a complementary relationship; regions with a rising prevalence of tap water access often show a corresponding decrease in tube well dependence. This observable trend can be interpreted as a proxy for development, reflecting improved access to formal water infrastructure in those regions.\par

\subsection{Socioeconomic Indicators of Poverty}
The NSSO dataset contains numerous key indicators that provide valuable insights into regional variations in socioeconomic conditions and food security. The spatiotemporal visualization of these indicators facilitates the identification of geographic patterns, regional disparities, and temporal shifts between 2001 and 2011. While the full dataset offers a wide range of variables for diverse analytical purposes, the following discussion will focus on a selection of representative indicators to illustrate the utility of our downscaled predictions. Specifically, we examine patterns related to the type of housing structure, the primary cooking fuel used, and the number of meals consumed per day.\par

\begin{figure}[!htb]
        \centering

    
        \begin{subfigure}[!htb]{0.31\linewidth}
            \centering
            \includegraphics[width=\linewidth]{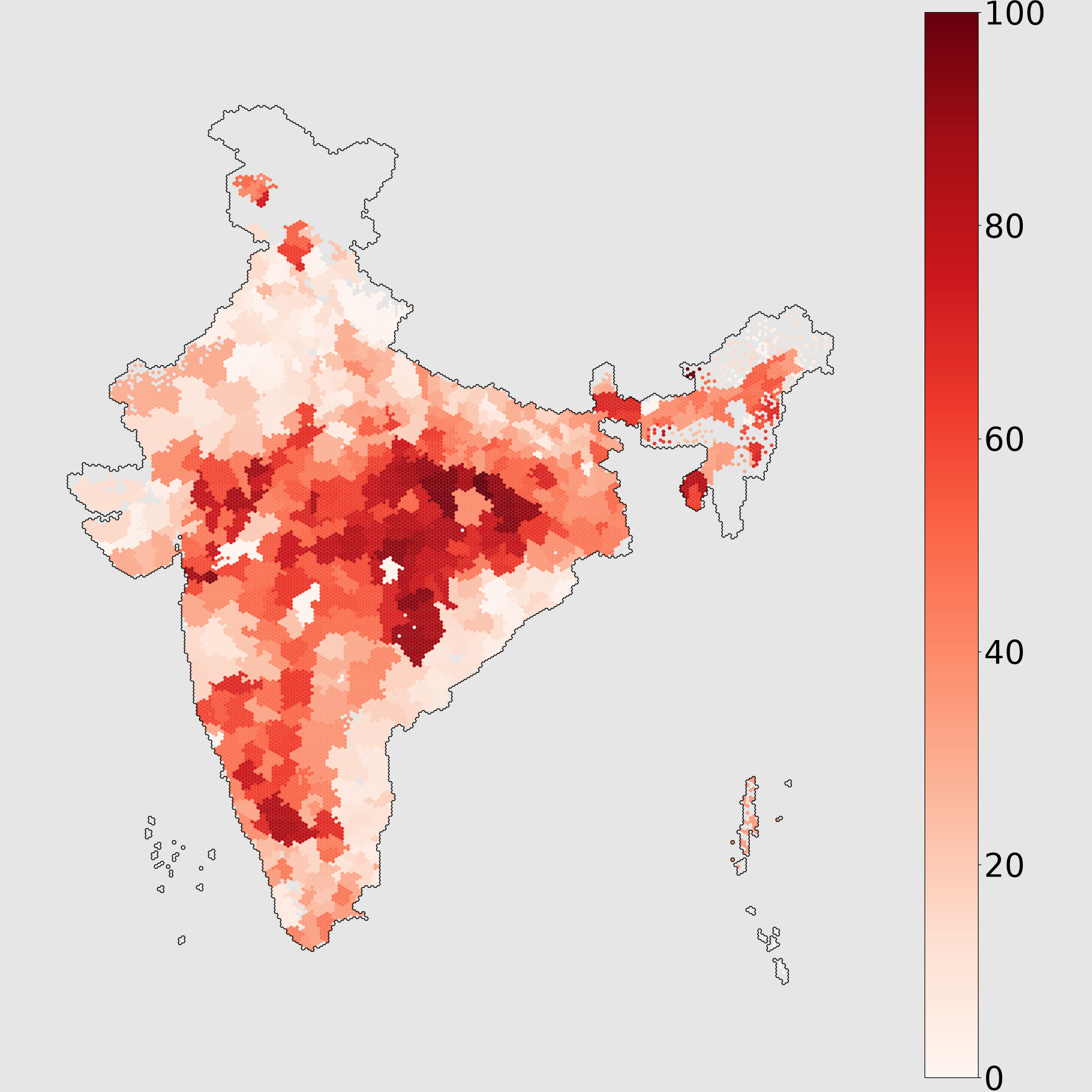}
            \captionsetup{justification=centering}
            \caption{}
            \label{fig:housebad_org_2001}

        \end{subfigure} 
        \begin{subfigure}[!htb]{0.31\linewidth}
            \centering
            \includegraphics[width=\linewidth]{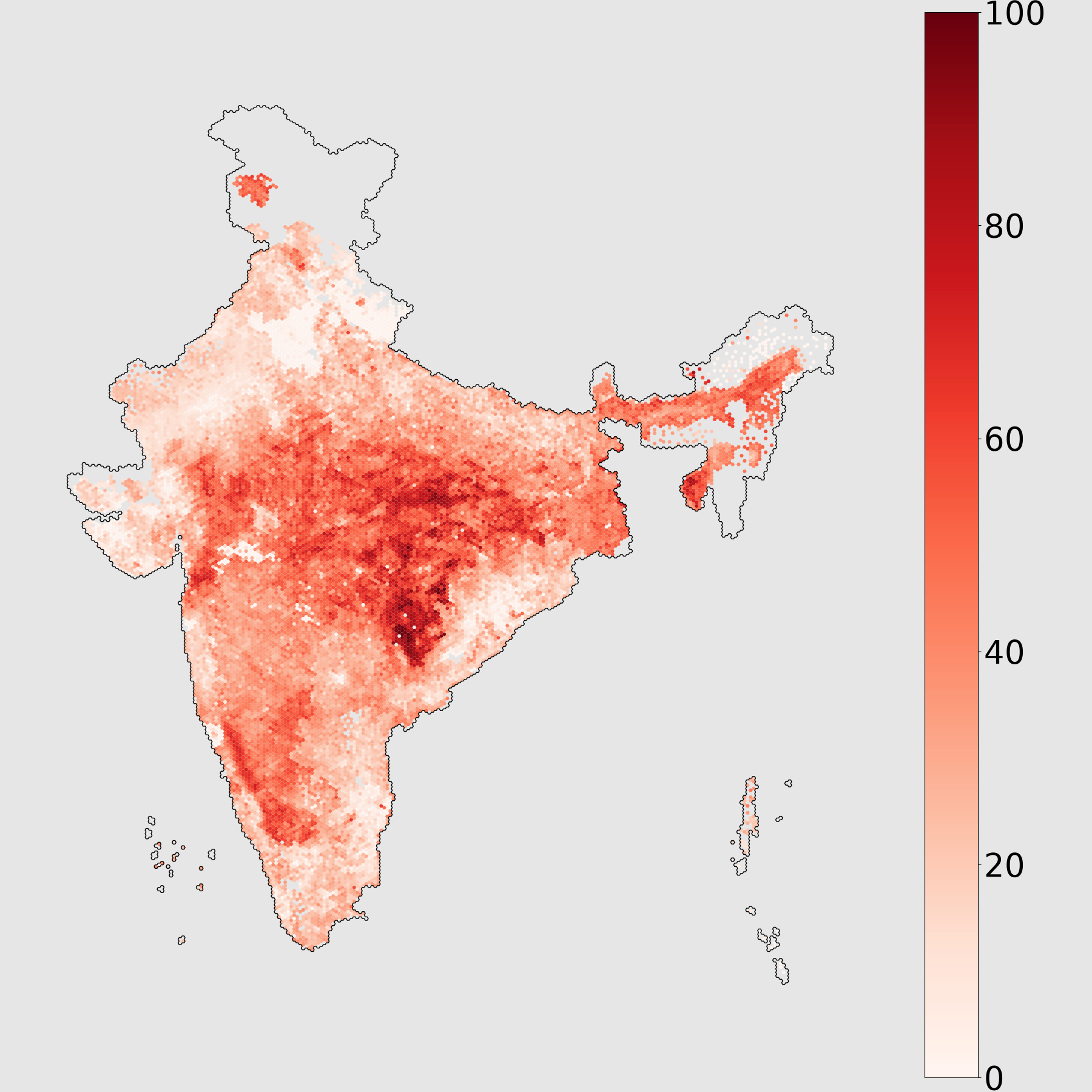}
            \captionsetup{justification=centering}
            \caption{}
            \label{fig:housebad_hex_2001}

        \end{subfigure} \\

        \begin{subfigure}[!htb]{0.31\linewidth}
            \centering
            \includegraphics[width=\linewidth]{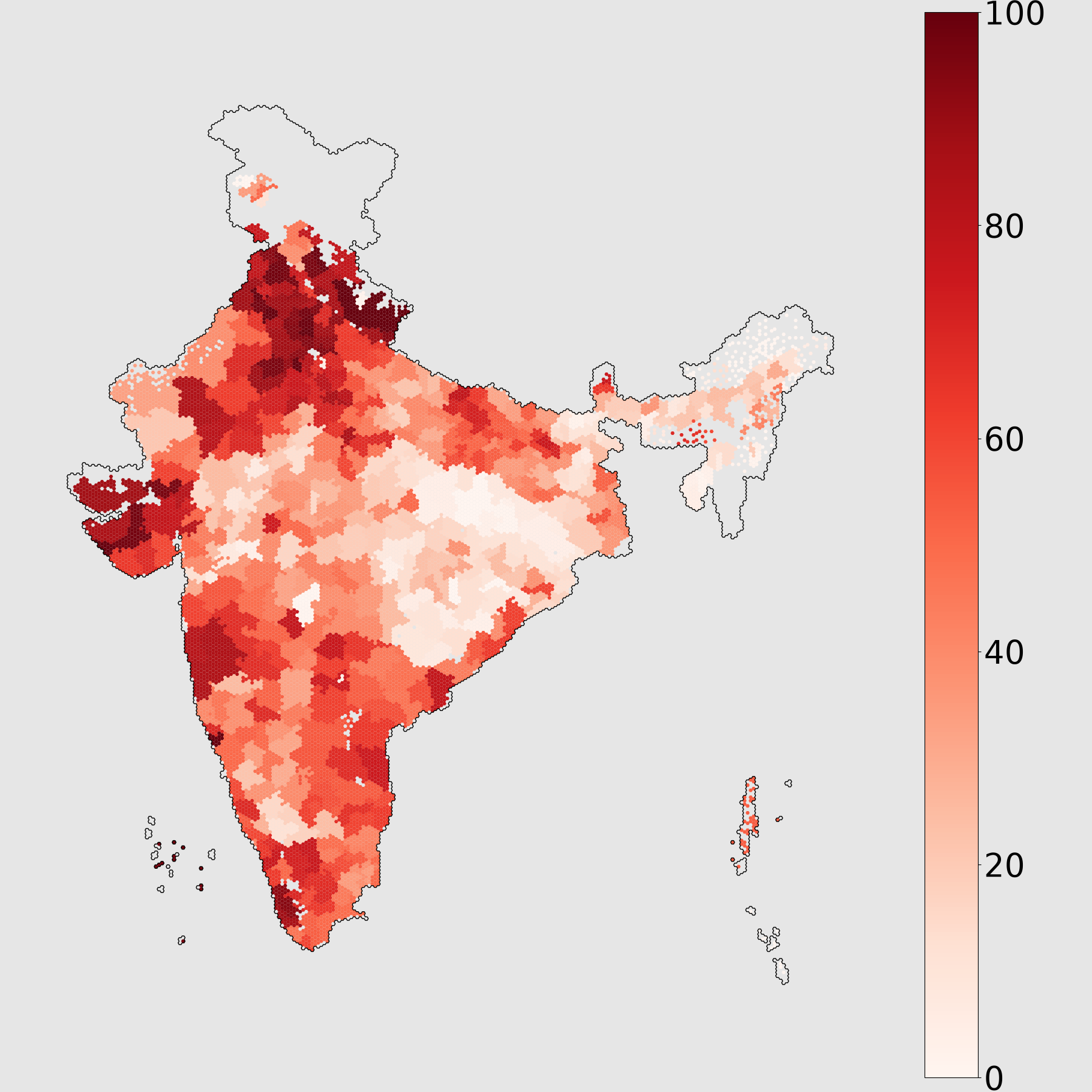}
            \captionsetup{justification=centering}
            \caption{}
            \label{fig:housegood_org_2001}

        \end{subfigure} 
        \begin{subfigure}[!htb]{0.31\linewidth}
            \centering
            \includegraphics[width=\linewidth]{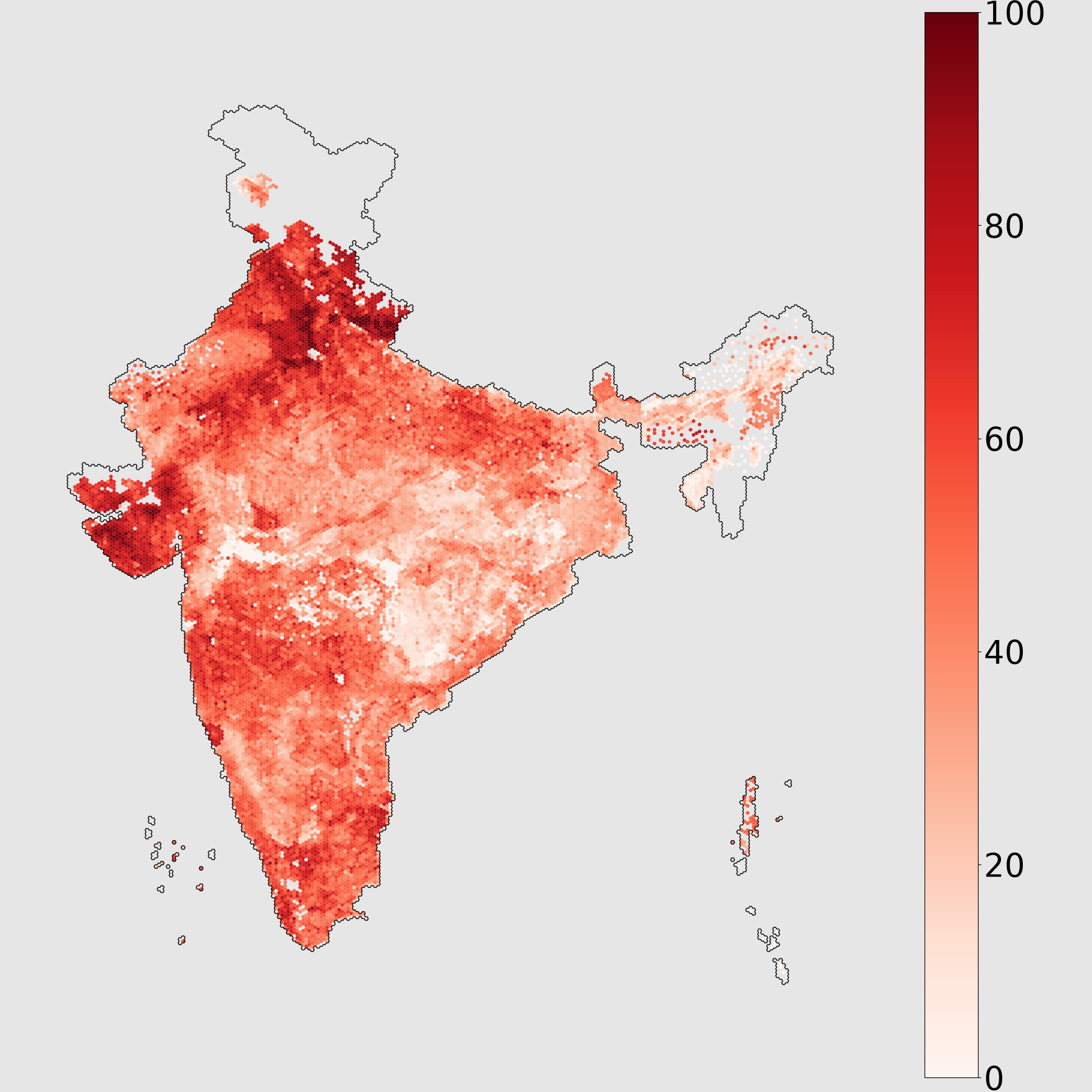}
            \captionsetup{justification=centering}
            \caption{}
            \label{fig:housegood_hex_2001}

        \end{subfigure} \\

        \caption[]{Each row shows, left: district-level ground‐truth, and right: cluster‐level predictions. (a) and (b) shows the percentage of people who live in a partially permanent structure (called a \textit{semi-pucca} house in Hindi language) with a mix of durable and temporary materials. (c) and (d) shows the percentage of people who live in a permanent structure (called a \textit{pucca} house in Hindi language) made of durable materials like concrete and brick. The data is from 2001.}
        \label{fig:house2001}
    
    \end{figure}

The first indicator examined is the `type of structure,' which categorizes housing into three classes: \textit{katcha} (structures made from temporary materials like mud and bamboo), \textit{pucca} (structures made from durable materials like brick and concrete), and \textit{semi-pucca} (a hybrid of the two). The prevalence of \textit{pucca} housing is a well-established proxy for higher socioeconomic status. Figure \ref{fig:house2001} presents the maps of the original district-level data alongside the high-resolution cluster-level predictions for the \textit{pucca} and \textit{semi-pucca} housing types for the year 2001.

A visual comparison of the maps reveals that the model successfully captures the distinct spatial distribution patterns for both housing types. For instance, regions with a high proportion of \textit{pucca} housing, such as parts of southern and western India, exhibit strong correspondence between the ground-truth data and the model's predictions. Similarly, the higher prevalence of \textit{semi-pucca} housing in some northeastern and central regions is also faithfully reproduced. Critically, the cluster-level predictions further resolve fine-scale variations within individual districts, elucidating localized housing conditions that are obscured by the coarse aggregation of the original district-level data.

\begin{figure}[!htb]
        \centering

    
        \begin{subfigure}[!htb]{0.31\linewidth}
            \centering
            \includegraphics[width=\linewidth]{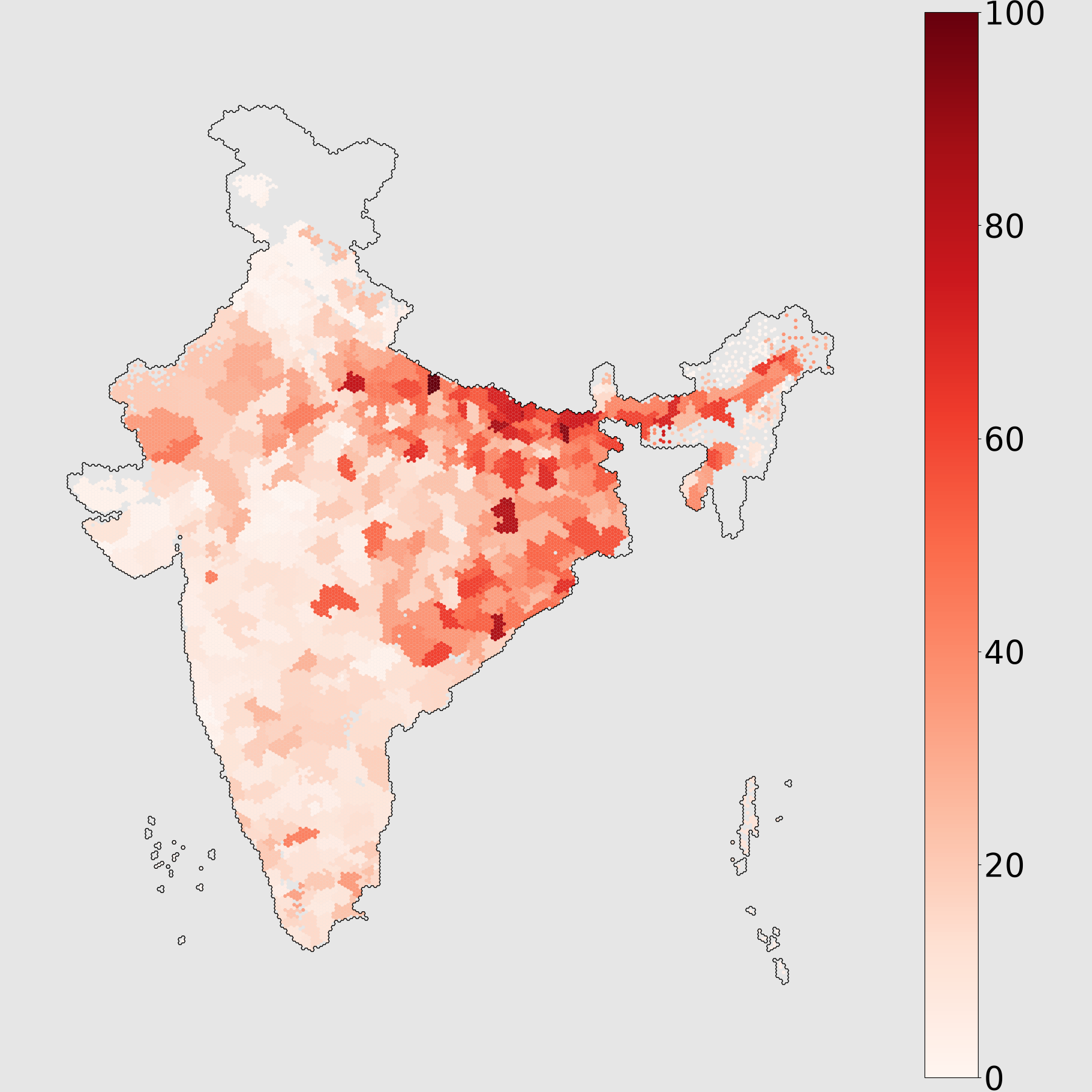}
            \captionsetup{justification=centering}
            \caption{}
            \label{fig:cooking_org_2001}

        \end{subfigure} 
        \begin{subfigure}[!htb]{0.31\linewidth}
            \centering
            \includegraphics[width=\linewidth]{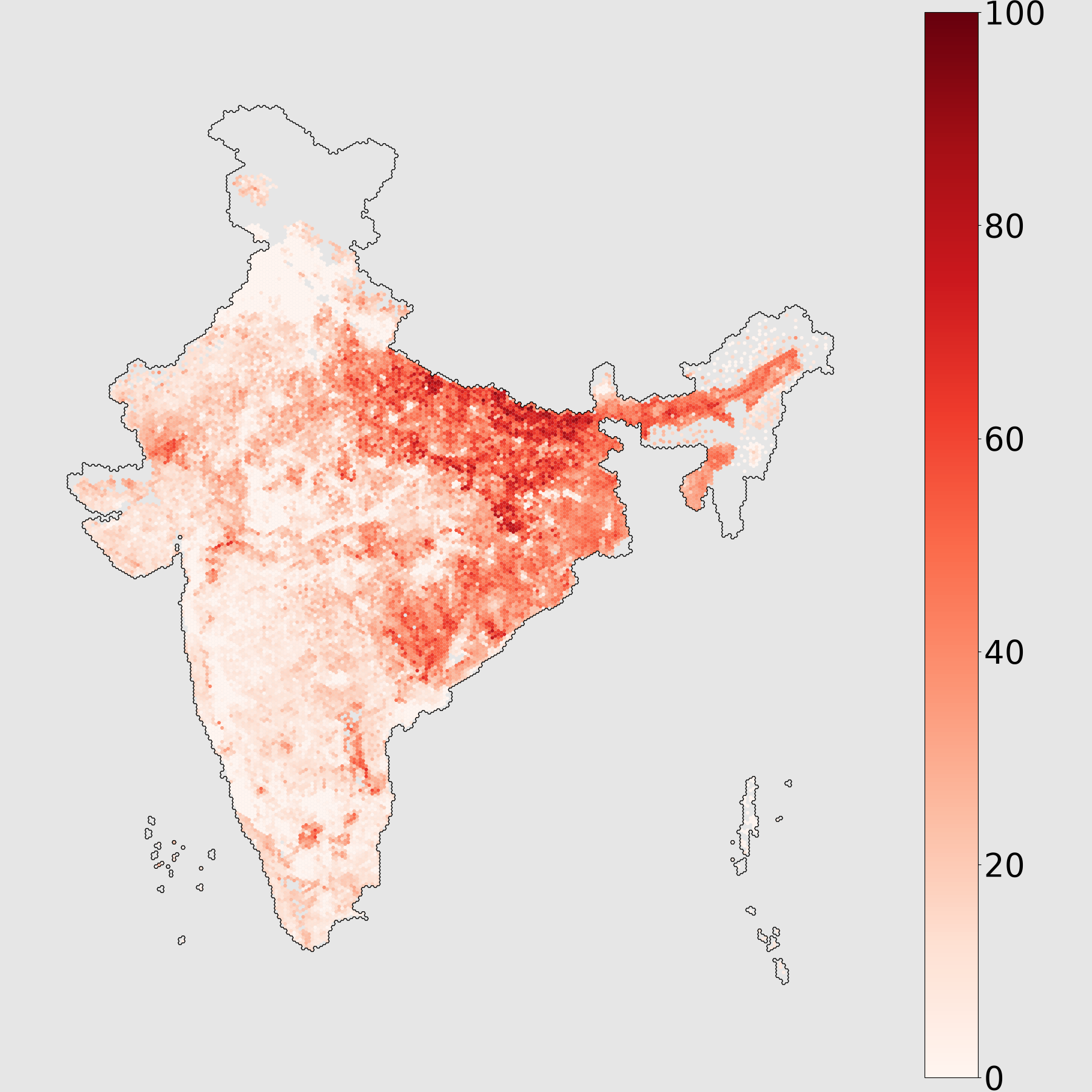}
            \captionsetup{justification=centering}
            \caption{}
            \label{fig:cooking_hex_2001}

        \end{subfigure} 

        \caption[]{The percentage of people who use kerosene as a fuel for cooking in 2001. (a) the district-level ground‐truth, (b) the cluster‐level predictions.}
        \label{fig:cooking2001}
    
    \end{figure}

The second illustrative indicator is the primary fuel used for cooking, with maps for 2001 presented in Fig. \ref{fig:cooking2001}. An analysis of the spatial pattern for kerosene usage reveals a high prevalence in specific regions that correspond to areas historically characterized by higher levels of poverty and developmental challenges. This elevated reliance on kerosene can be interpreted as a proxy for limited access to more modern cooking fuels, such as LPG or electricity, and a continued dependence on subsidized energy sources.\par

This example demonstrates the utility of the downscaled data for regional disparity analysis. The high-resolution maps can help identify localized, sub-district areas with limited access to clean fuels—a condition often correlated with underdevelopment and infrastructure gaps—thereby highlighting pockets of persistent socioeconomic vulnerability.\par

\begin{figure}[!htb]
        \centering

    
        \begin{subfigure}[!htb]{0.31\linewidth}
            \centering
            \includegraphics[width=\linewidth]{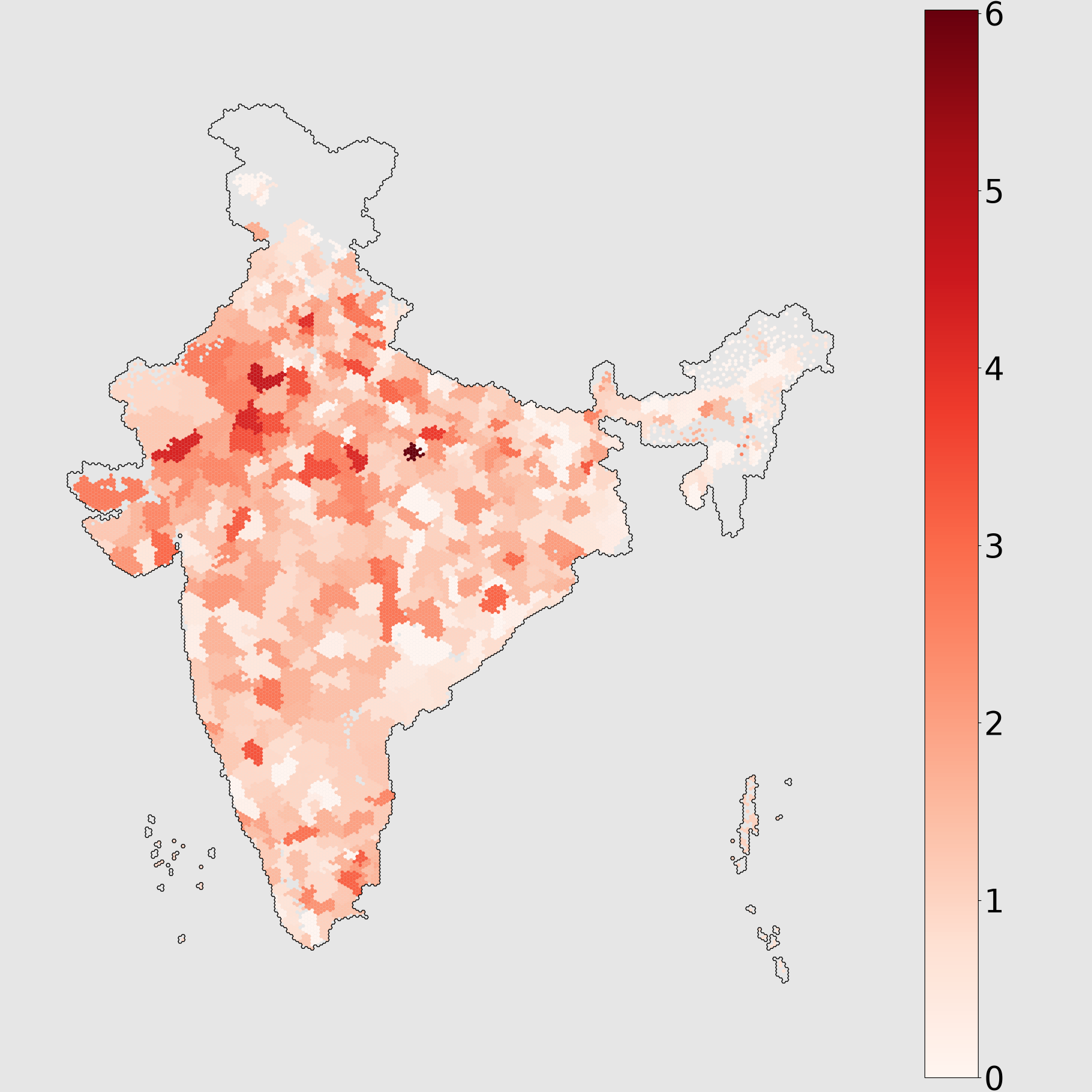}
            \captionsetup{justification=centering}
            \caption{}
            \label{fig:meal_org_2001}

        \end{subfigure} 
        \begin{subfigure}[!htb]{0.31\linewidth}
            \centering
            \includegraphics[width=\linewidth]{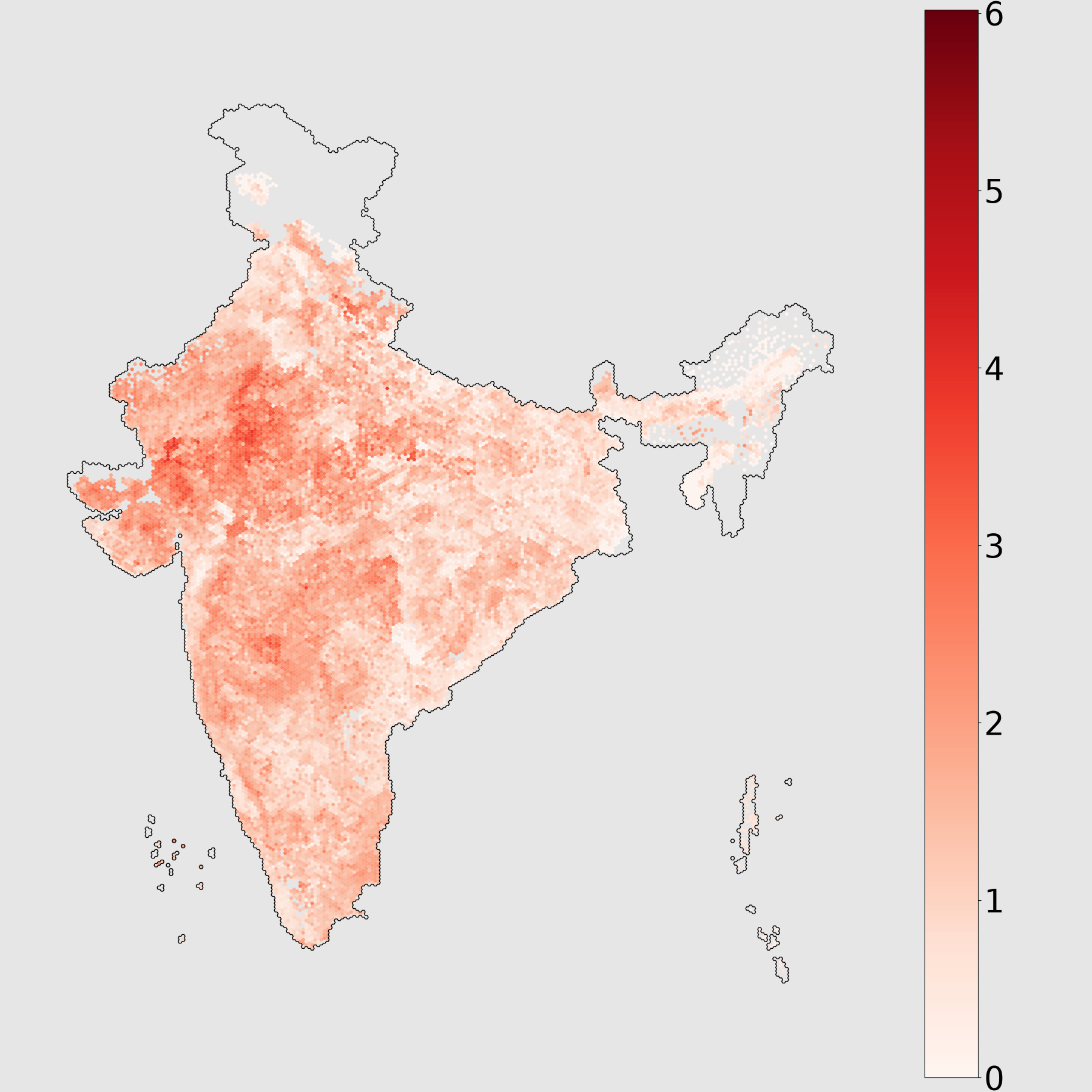}
            \captionsetup{justification=centering}
            \caption{}
            \label{fig:meal_hex_2001}

        \end{subfigure} \\

        \begin{subfigure}[!htb]{0.31\linewidth}
            \centering
            \includegraphics[width=\linewidth]{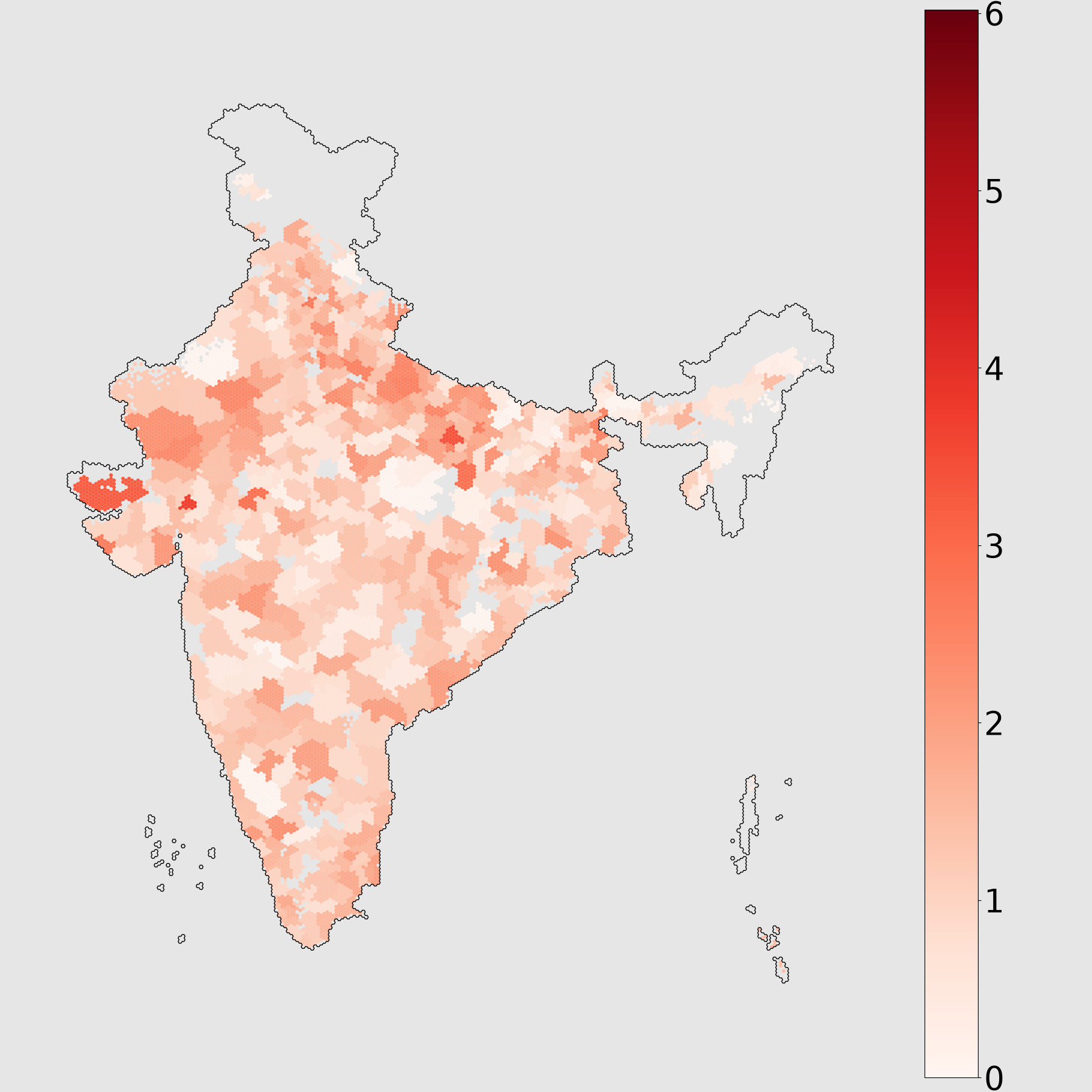}
            \captionsetup{justification=centering}
            \caption{}
            \label{fig:meal_org_2011}

        \end{subfigure} 
        \begin{subfigure}[!htb]{0.31\linewidth}
            \centering
            \includegraphics[width=\linewidth]{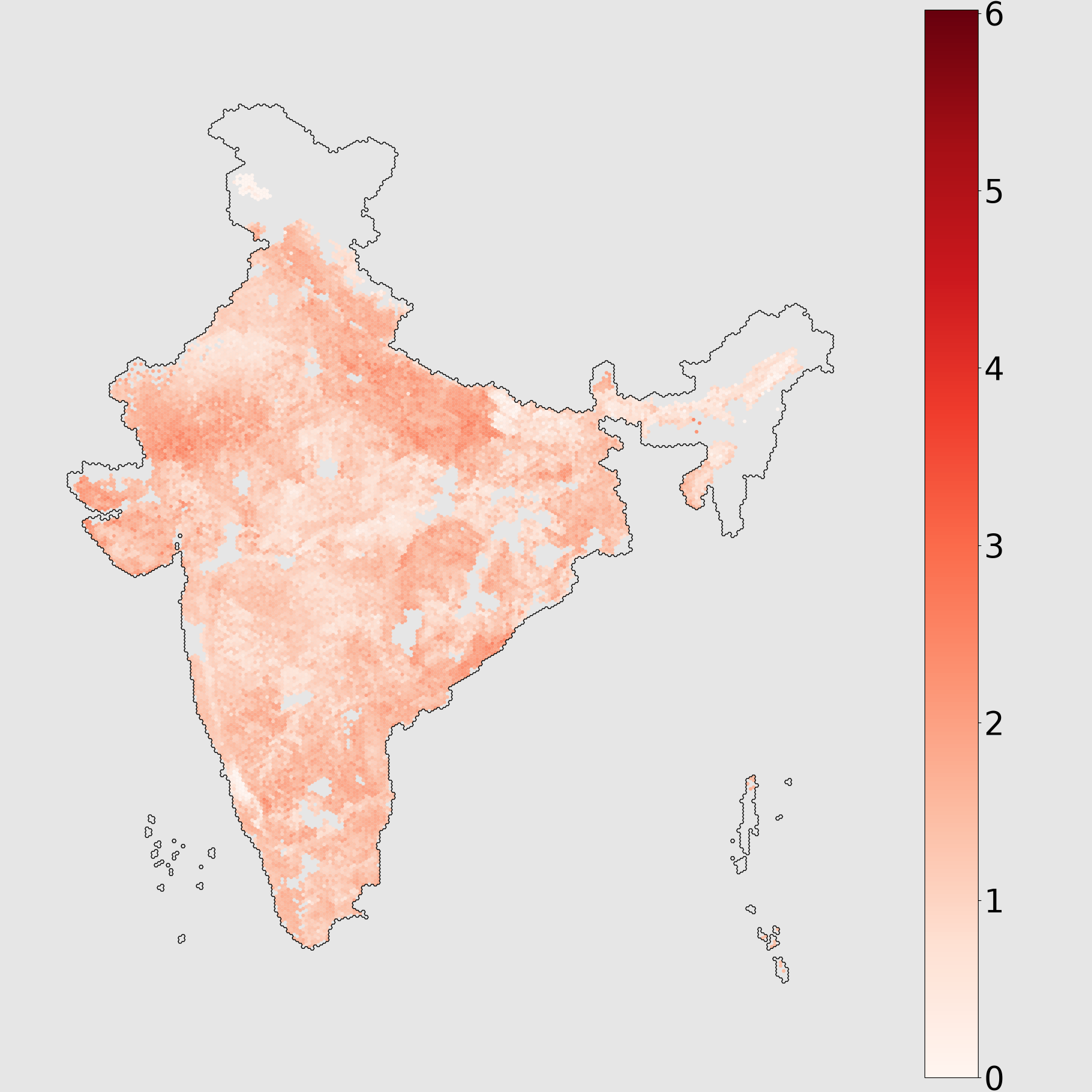}
            \captionsetup{justification=centering}
            \caption{}
            \label{fig:meal_hex_2011}

        \end{subfigure} \\

        \caption[]{The percentage of people who experienced days within the past 30 days when they could not afford even a single meal. (a) the district-level ground‐truth from 2001, (b) the cluster‐level predictions from 2001, (c) the district-level ground‐truth from 2011, and (d) the cluster‐level predictions from 2011.}
        \label{fig:meals}
    
    \end{figure}

The final illustrative indicator is a direct measure of severe food insecurity, capturing the proportion of the population that experienced one or more days in the preceding month without a single meal due to an inability to afford food. The spatial distribution of this indicator for 2001 is presented in Fig. \ref{fig:meals}, which shows both the original district-level data and the corresponding high-resolution cluster-scale prediction.\par

The spatial distribution of this indicator of food insecurity in 2001 exhibits a dispersed pattern, likely reflecting localized pockets of acute poverty in districts with limited employment opportunities. A temporal comparison reveals a marked improvement by 2011, with a significant reduction in the number of regions reporting high values for this indicator. This trend suggests an overall decline in extreme food insecurity, potentially indicating broader improvements in access to basic necessities during this period.\par
This type of spatiotemporal analysis exemplifies the utility of the downscaled dataset for temporal poverty tracking. Such an approach can highlight districts experiencing improvements in food security over time and contribute to the evaluation of social welfare programs and policy interventions.





\section{Method}
\label{sec:Method}


This section describes the JuGAAD framework in detail. JuGAAD integrates coarse socioeconomic indicators from NSSO with fine-resolution census, geographical, and state-level features to learn mappings that allow the generation of spatially detailed socioeconomic estimates.

\begin{figure}[!htb]
    \centering
    \includegraphics[width=\linewidth]{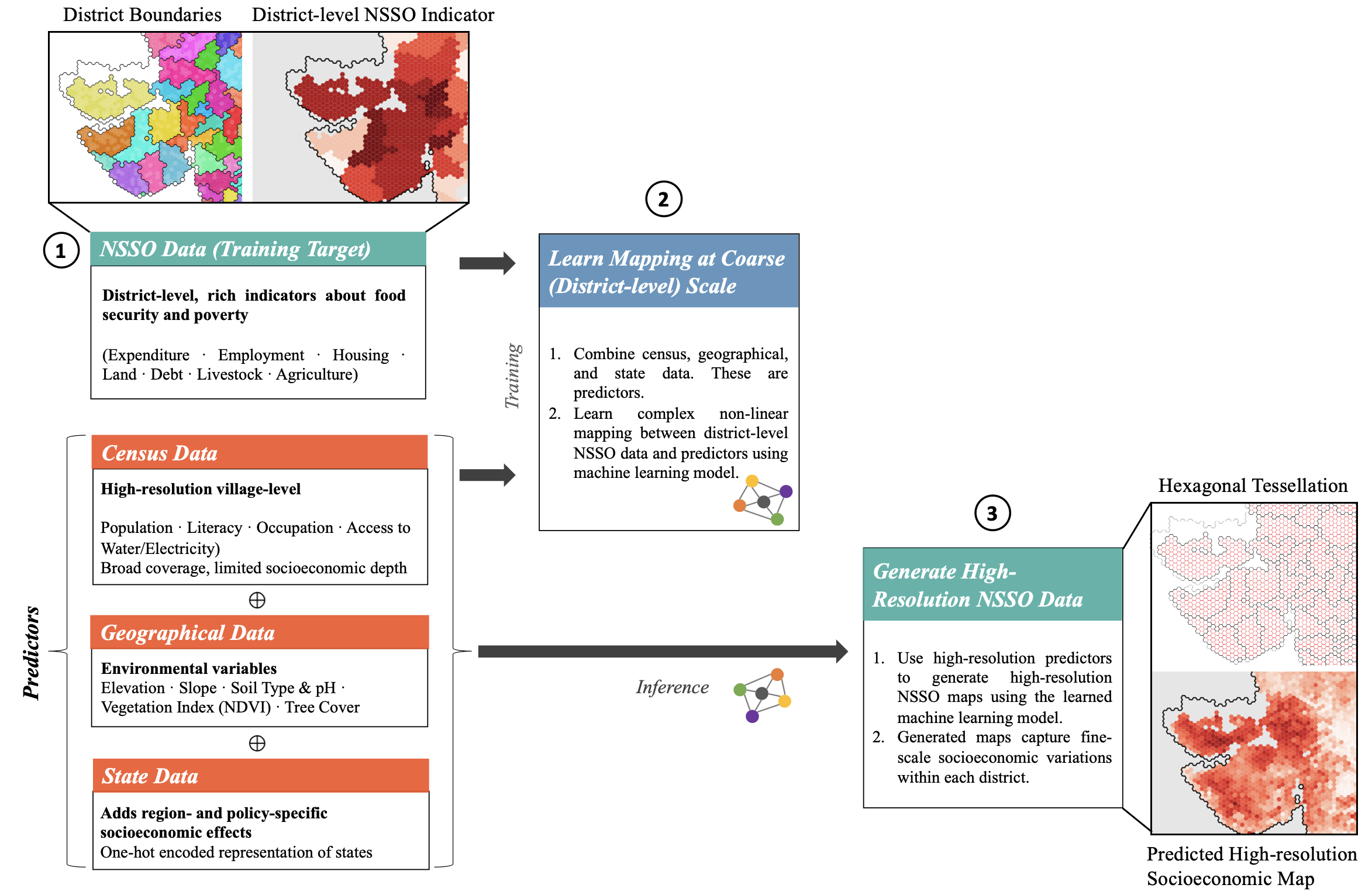}
    \caption{Conceptual overview of the proposed JuGAAD framework. The model learns relationships between coarse but detailed NSSO socioeconomic indicators and fine-scale predictors (census, geographical, and state data) at the district level and applies the learned mapping to generate high-resolution detailed socioeconomic estimates.}
    \label{fig:conceptual}
\end{figure}

The overall pipeline is summarized in Fig. \ref{fig:conceptual}. Subsequent subsections describe the data sources and the model design in detail.

\subsection{Data Sources}

\begin{figure}[!htb]
        \centering
    
        \begin{subfigure}[htb]{0.4\linewidth}
            \centering
            \includegraphics[width=\linewidth]{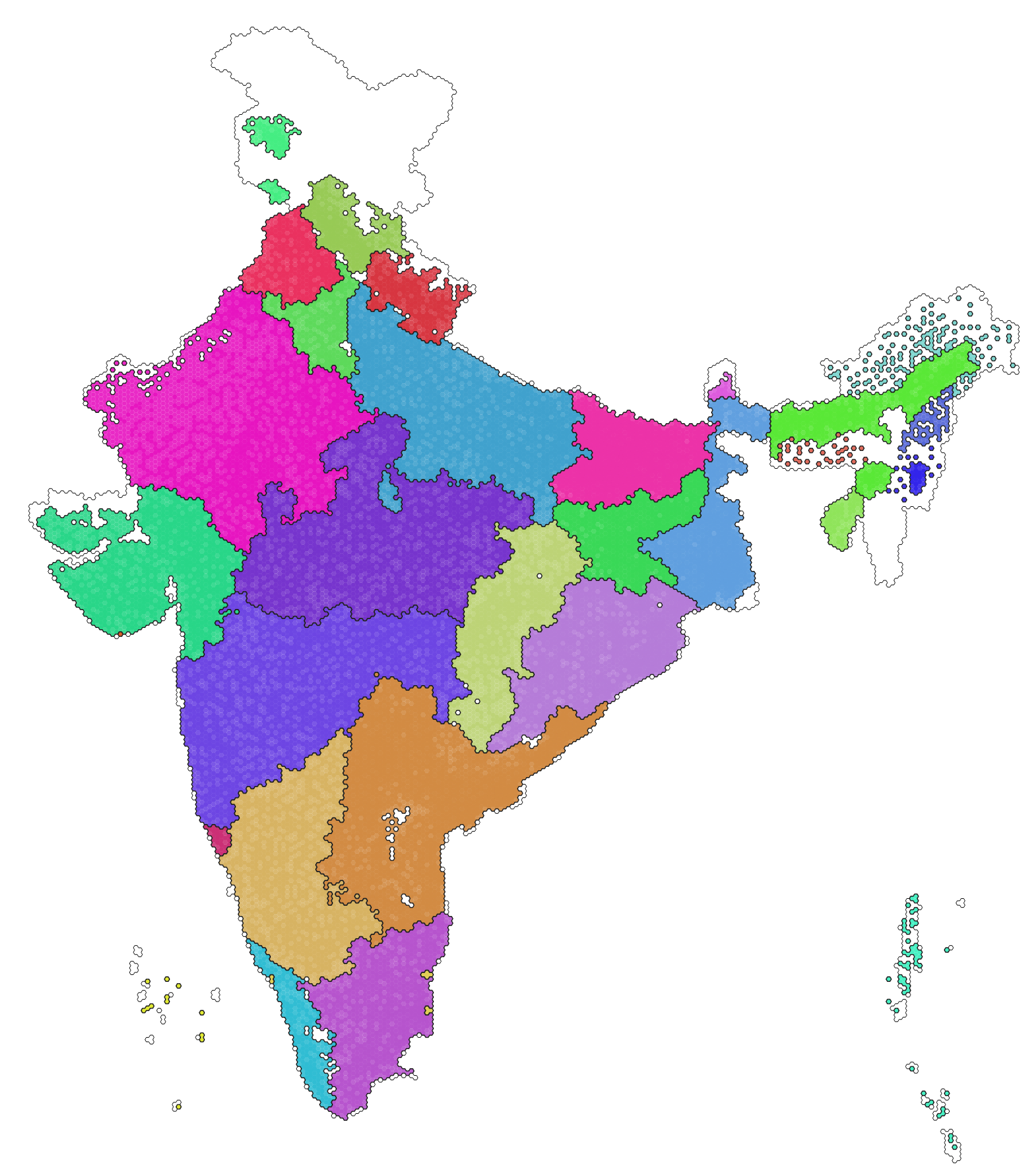}
            \captionsetup{justification=centering}
            \caption{}
            \label{fig:states}
        \end{subfigure}
        \begin{subfigure}[htb]{0.4\linewidth}
            \centering
            \includegraphics[width=\linewidth]{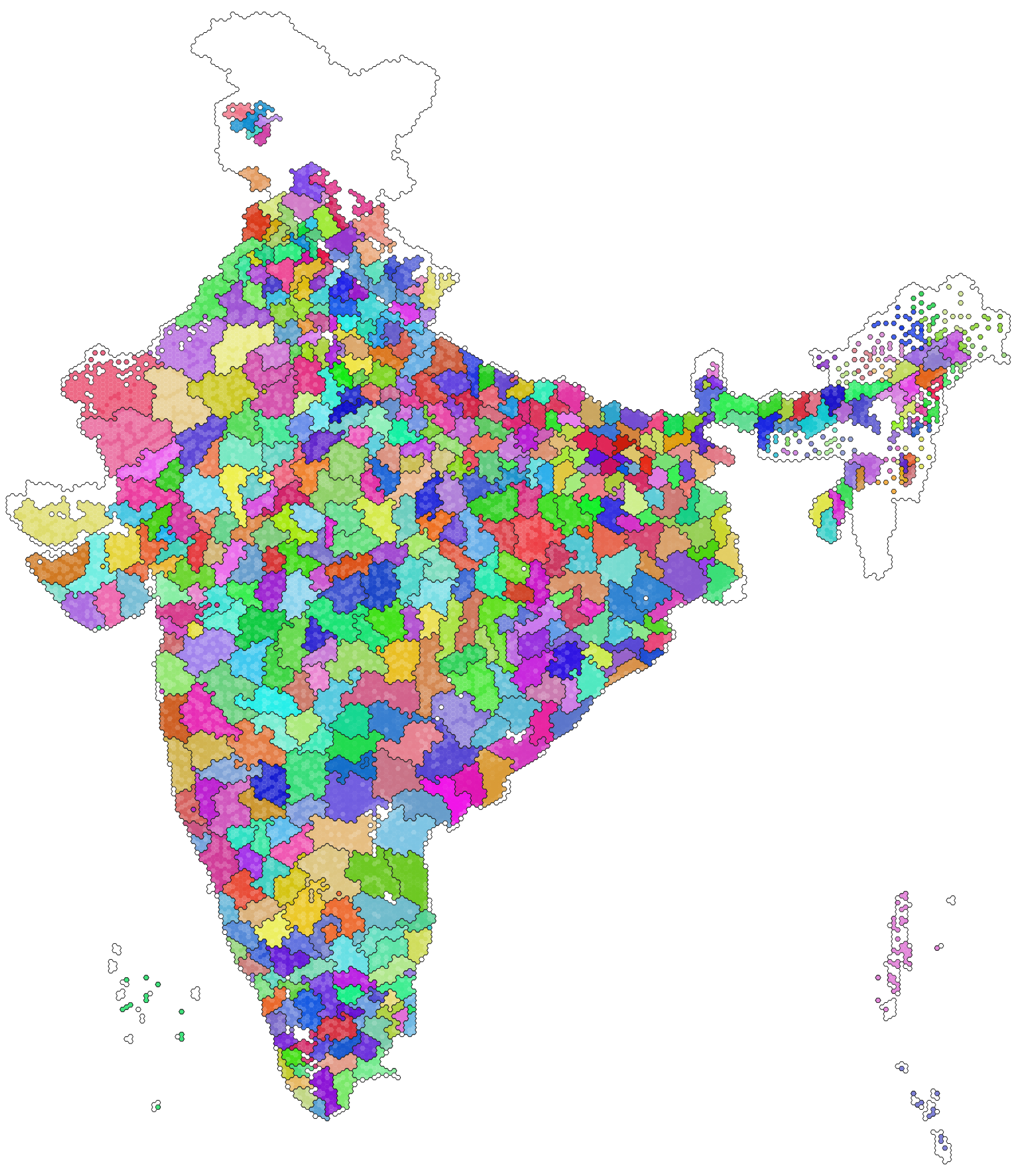}
            \captionsetup{justification=centering}
            \caption{}
            \label{fig:districts}
        \end{subfigure} 
        \begin{subfigure}[htb]{0.4\linewidth}
            \centering
            \includegraphics[width=\linewidth]{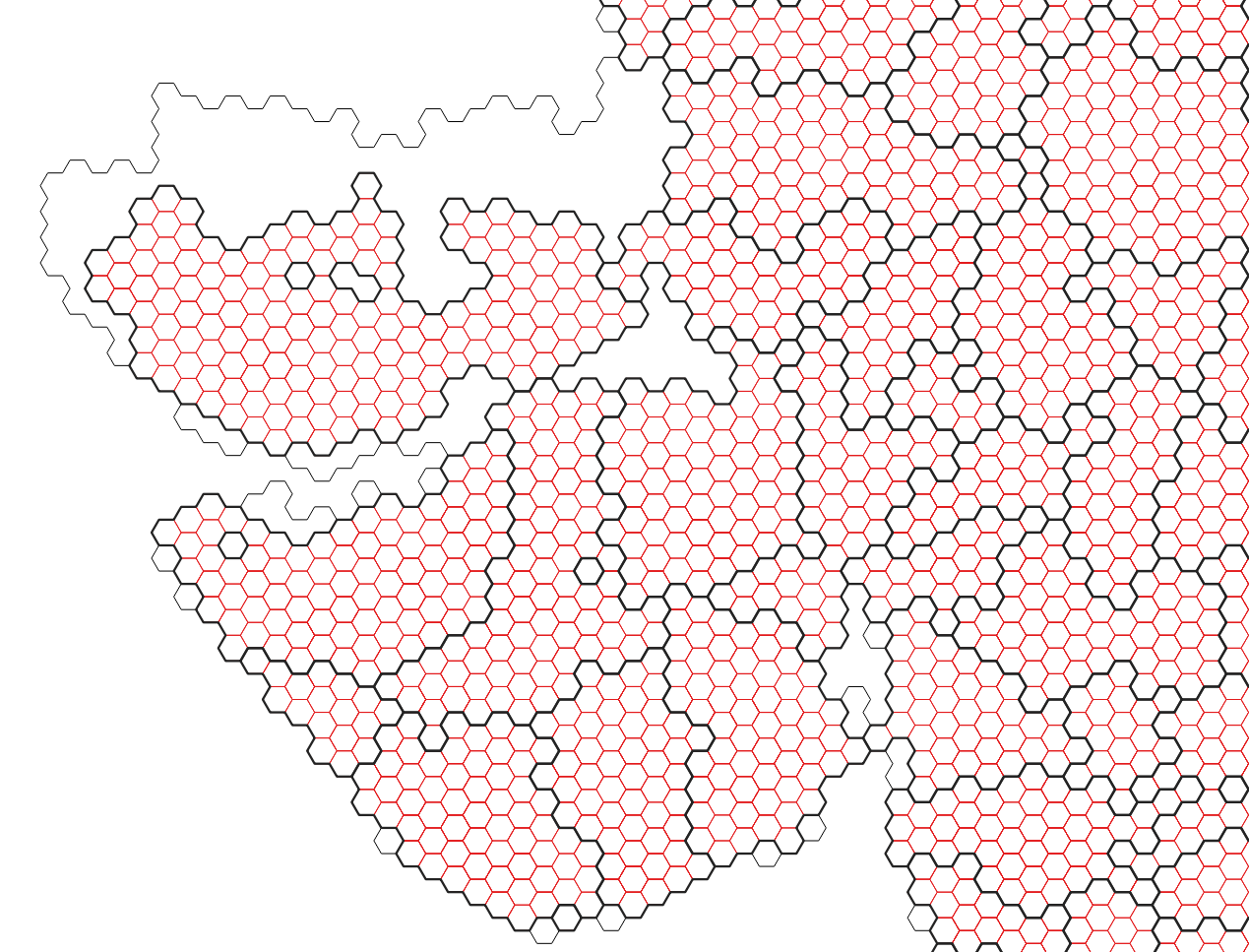}
            \captionsetup{justification=centering}
            \caption{}
            \label{fig:hexagon}
        \end{subfigure} 

        \caption[]{The map of India showing (a) states and (b) different districts in each state. (c) Uniform hexagons are used as fine-scale village cluster-scale prediction units. Each district (separated by black boundaries) contains multiple hexagons, enabling localized analysis of socioeconomic indicators.}
        \label{fig:data_maps}
    
    \end{figure}

\begin{figure}[!htb]
        \centering
    
        \begin{subfigure}[htb]{0.4\linewidth}
            \centering
            \includegraphics[width=\linewidth]{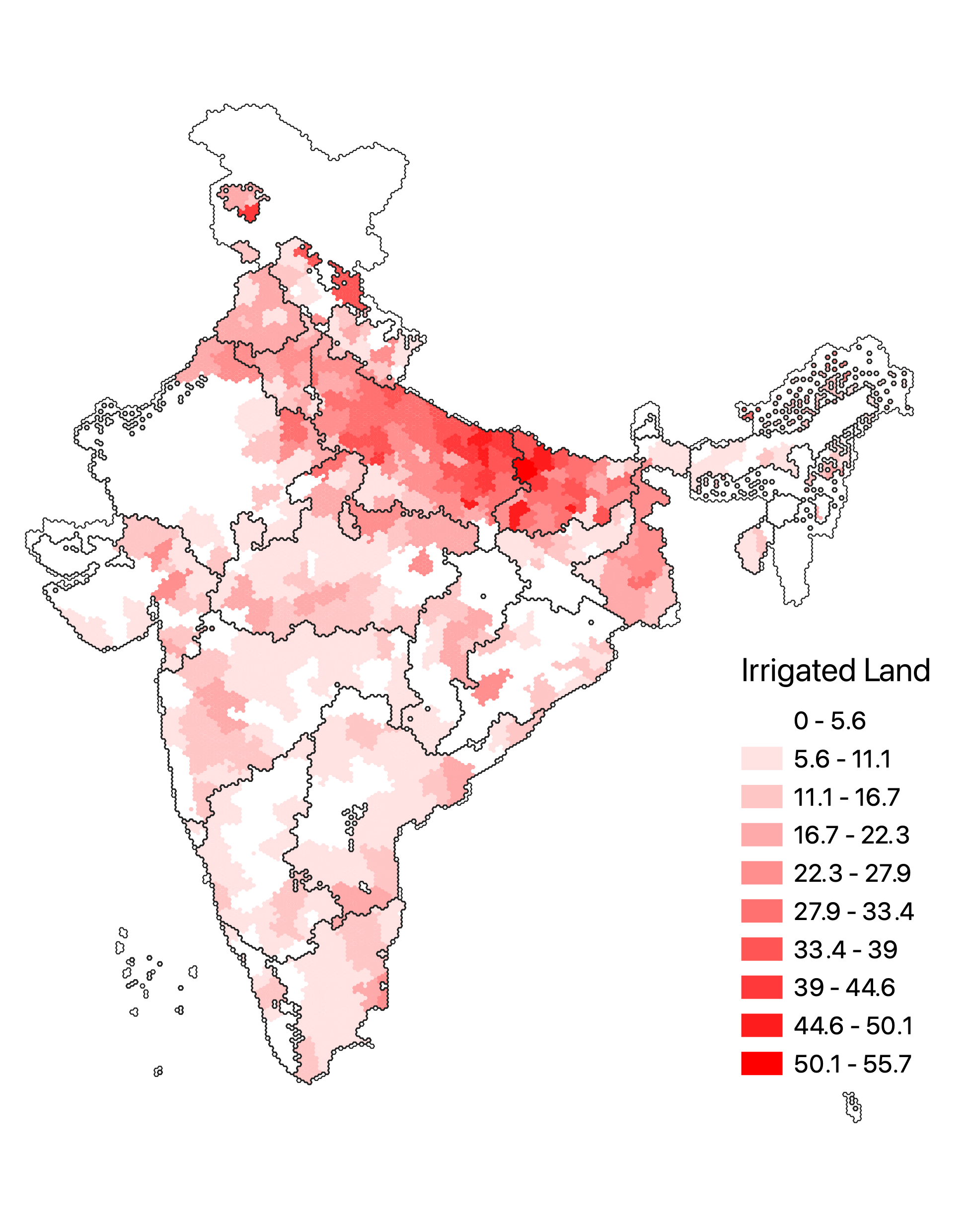}
            \captionsetup{justification=centering}
            \caption{}
            \label{fig:nsso_data_sample}
        \end{subfigure}
        \begin{subfigure}[htb]{0.4\linewidth}
            \centering
            \includegraphics[width=\linewidth]{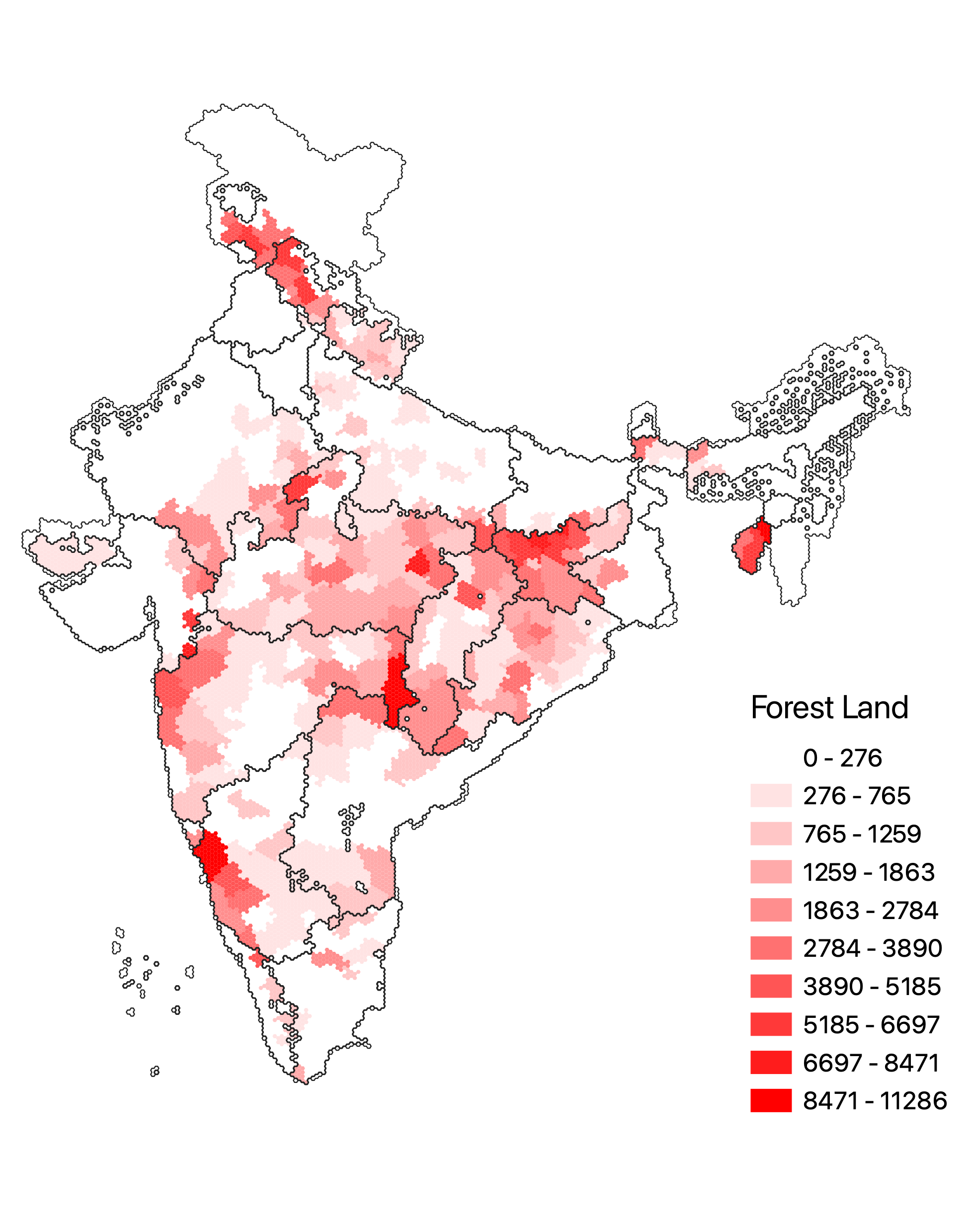}
            \captionsetup{justification=centering}
            \caption{}
            \label{fig:census_data_sample}
        \end{subfigure} 
        \begin{subfigure}[htb]{0.4\linewidth}
            \centering
            \includegraphics[width=\linewidth]{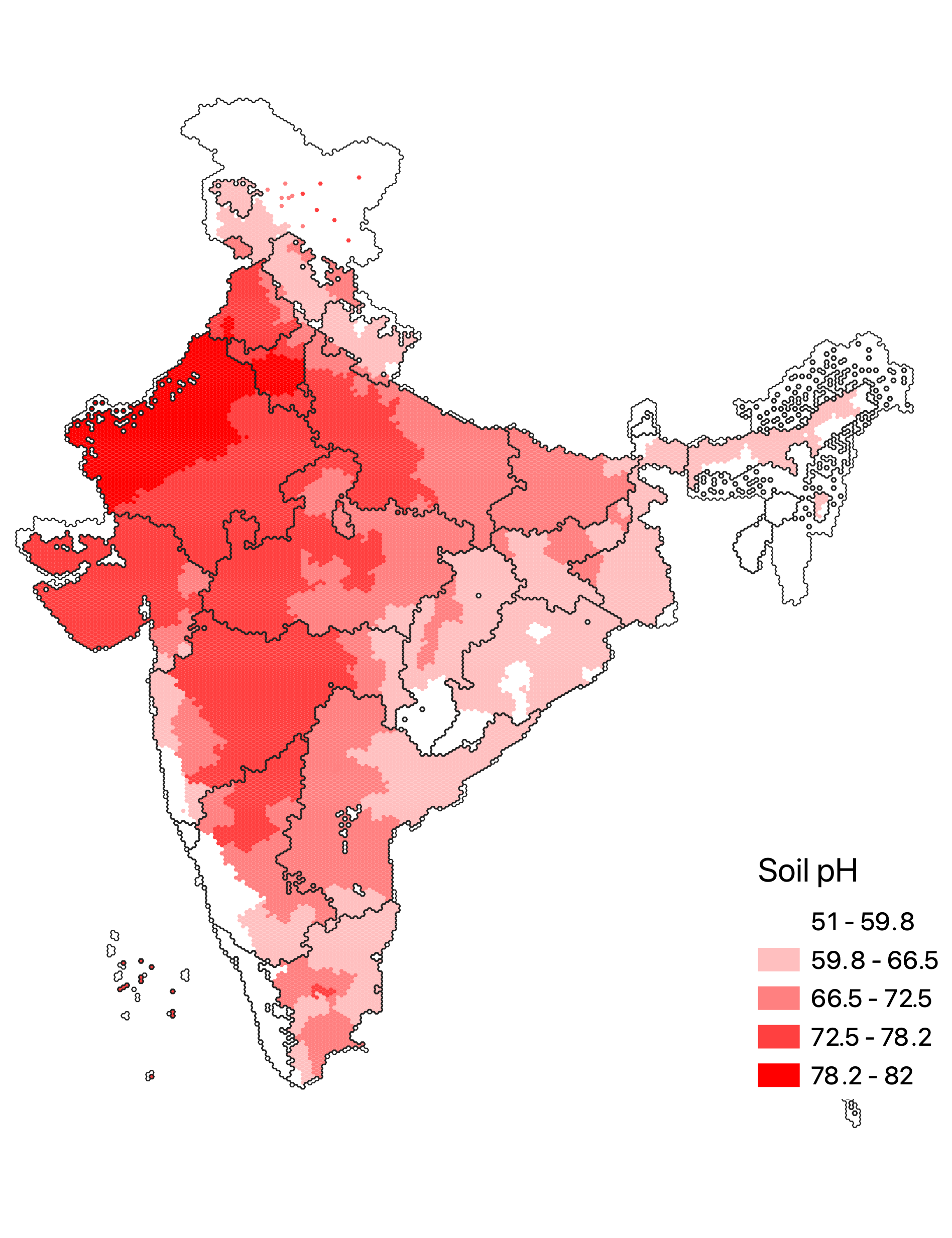}
            \captionsetup{justification=centering}
            \caption{}
            \label{fig:geo_data_sample}
        \end{subfigure} 

        \caption[]{Example of features from each data source: (a) Irrigated land map from NSSO data, (b) Forest area map from Census data, and (c) Soil pH map from geographic data. Each map illustrates the spatial distribution of the respective variable across India. All maps are based on data from $2001$.}
        \label{fig:data_samples}
    
    \end{figure}

Data Sources and Spatial Framework
This study integrates two principal sources of nationwide data from India: the National Sample Survey Office (NSSO) and the Indian Census, using datasets from both for 2001 and 2011. Our methodology is designed to leleveragesgths of each dataset to produce high-resolution, detailed socioeconomic indicators.
\begin{enumerate}
\item \textbf{NSSO Data:} 
The NSSO provides extensive data on a wide range of socioeconomic topics through large-scale household surveys. These surveys are organized into six primary categories: Consumer Expenditure, Employment, Agriculture, Housing Conditions, Land and Livestock Holdings, and Debt and Investment. The thematic richness of the NSSO data is a key strength, offering detailed indicators unavailable in the census, such as housing construction materials, specific agricultural practices, household dietary habits, and access to credit. For the purposes of this study, the NSSO data is aggregated at the district level, as illustrated by the spatial variation in irrigated land shown in Fig. ~\ref{fig:nsso_data_sample}.

\item \textbf{Census Data:} 
The Indian Census offers a comprehensive demographic and socioeconomic snapshot of the entire population. While less thematically detailed than the NSSO surveys, the Census provides data at a much finer spatial resolution, including both district and village levels (an example feature is shown in Fig. ~\ref{fig:census_data_sample}). However, the coverage and availability of village-level census data are inconsistent across different years, villages often change administrative units, or are added or removed causing gaps in the record. To establish a uniform basis for our analyses, we constructed a series of hexagonal tessellations across the entire country using an iterative methodology: we aimed at finding a size that would be large enough to allow averaging of census and geospatial data to reduce noise, while being small enough to preserve the spatial representation of districts and subdistricts (\textit{taluks/tehsils}). A vertex-to-vertex size of 15Km was found to be optimal (\textasciitilde20 villages to a hexagon), hexagons with fewer than 5 villages were dropped to avoid biasing data and extrapolating into sparsely populated areas. Districts and subdistrict IDs were assigned to each hexagon using a simple majority rule for all village centroids falling within each hexagon. Each hexagon therefore represents a village cluster and is a consistent spatial unit of analysis across years that is considerably smaller than a district and  subdistrict, and allows the capture of the hierarchical spatial structure of administrative units. This scheme also ensures that villages changing administrative boundaries or getting added/removed between survey periods do not unduly influence large-scale inference. Generating data for subsequent years is also made simple as the same aggregation scheme can be applied keeping the number of records (the number of hexagons) the same. Figure \ref{fig:data_maps} illustrates the spatial schema (see specifically Fig. \ref{fig:hexagon}). 
\end{enumerate}

A central methodological challenge arises from the disparity between the datasets: the NSSO provides rich, nuanced indicators but only at a coarse spatial resolution (districts), while the Census offers high-resolution data that lacks thematic detail. This study addresses this gap by using the downscaled Census indicators as predictors to estimate the detailed NSSO indicators at the village cluster level. The resulting predictions yield a novel, high-resolution dataset that combines the spatial granularity of the Census with the thematic depth of the NSSO, enabling a more nuanced analysis of socioeconomic landscapes and their development over the study period. The complete NSSO dataset utilized is available as supplementary material.


\subsubsection{NSSO Data Preprocessing} \label{sec:DataPrep}
A critical preprocessing phase was necessary to address two primary challenges inherent in the raw NSSO data: data completeness and feature dimensionality. First, the prevalence of missing values across numerous states and districts necessitated the implementation of a robust gap-filling procedure. Second, the dataset comprises both continuous (non-negative) and discrete (compositional categories summing to 100\%) features. Many discrete variables exhibit high dimensionality and significant sparsity, which can adversely affect model convergence and performance. To mitigate this, a dimensionality reduction strategy was applied to generate a more compact feature representation, thereby improving both computational efficiency and model efficacy. The complete preprocessing pipeline is detailed below.

\begin{enumerate}
\item \textbf{Gap-filling strategy:} 
To address the challenge of non-uniformly distributed missing data, we developed a predictive imputation methodology. A complete-case analysis was deemed unsuitable as it would result in a substantial loss of geographic coverage. Instead, our approach imputes missing values for each NSSO feature by training a model on a set of predictor variables from other thematic categories. Prior to imputation, the predictor set is refined by removing highly correlated features and standardizing the remaining variables to a zero mean and unit variance. The imputation model consists of a shallow regression neural network with a single hidden layer and a ReLU activation function on the output layer to enforce non-negativity. The network is trained using the Adam optimizer to minimize mean squared error loss. The validity of each imputation was evaluated using the coefficient of determination ($R^2$), with an acceptance threshold of $R^2 \geq 0.6$. If a feature could not be predicted with sufficient accuracy, alternative predictor categories were tested. Features that failed to meet the performance threshold across all predictor combinations were removed from the entire dataset. Finally, to ensure a consistent analytical framework, only features present across all districts post-imputation were retained for subsequent analysis.
\par



\item \textbf{Data cleaning:}
Following imputation, all the features of the various NSSO categories were consolidated into a unified analytical data set. This data set was subjected to a rigorous feature selection process to enhance the performance and interpretability of the model. Initially, we systematically removed uninformative predictors, specifically those exhibiting zero or near-zero variance (i.e., constant or quasi-constant features). To address multicollinearity, we then pruned features that were highly correlated with others. However, an exception was made for a curated set of variables directly pertinent to food security and poverty; these were deliberately retained for downstream analysis, regardless of their correlation coefficients, due to their theoretical importance.

High-dimensional categorical variables presented a distinct challenge, particularly within the Consumer Expenditure category, some of which comprised over 200 discrete classes. Such features introduce significant sparsity, which can impede the learning capacity of neural networks. Upon inspection, these high-dimensional variables were found to possess minimal variance and were consequently eliminated from the dataset based on the criterion of near-zero variance. \par

After this step, the NSSO dataset contains $475$ features, summarized in Table~\ref{table:nsso_dist}.
\end{enumerate}

\subsubsection{Additional Data Sources: Geographic Information and State Identifiers} \label{sec:AddnData}
To enhance the predictive power of the model beyond the primary Census indicators, the feature space was augmented with a comprehensive set of auxiliary geospatial data. We incorporated $48$ distinct geographical variables, including soil characteristics (\textit{e.g.}, sand fraction, organic content, pH), vegetation indices (\textit{e.g.}, tree cover, NDVI), climate data, and topographic information such as elevation. The inclusion of these variables is justified by their well-documented correlation with patterns of socioeconomic development and urbanization, thereby providing crucial geospatial context for the prediction task. Figure \ref{fig:geo_data_sample} shows the spatial variation of soil pH across India. \par

To control for unobserved heterogeneity arising from distinct administrative, economic, and political contexts at the state level, we incorporated state identifiers as categorical features in the model. These identifiers were subsequently one-hot encoded, transforming each state's numeric label,$i$, into a binary vector where the $i^{\text{th}}$ index is set to 1 and all other elements are 0. This final set of state-level fixed effects, combined with the census and geographical variables, constitutes the complete feature space used to predict the NSSO indicators. \par

The final input feature set for the prediction model is composed of $58$ census-derived variables, $48$ geographical variables, and $35$ one-hot encoded state and union territory identifiers, constituting a feature vector with a total dimensionality of $141$.

\begin{table*}[!htb]
    \centering
    \caption{Number of NSSO features per category after preprocessing.}
    \footnotesize
    \begin{tabular}{lcc}
    \toprule
    \textbf{Data Category} & \textbf{Number of Features} & \textbf{Latent space Dimensionality} \\ [0.5ex]
    \midrule
    Consumer Expenditure & $110$ & $20$\\
    Employment & $92$ & $20$\\
    Agriculture & $76$ & $16$\\
    Housing Conditions & $89$ & $20$ \\
    Land and Livestock Holdings & $45$ & $16$\\
    Debt and Investment & $83$ & $20$\\
    \midrule
    \textbf{Total} & \textbf{475} & \textbf{112}\\
    \bottomrule
    \end{tabular}
\label{table:nsso_dist}
\end{table*}


\subsection{Architecture}


The primary modeling objective is to simultaneously predict the full suite of NSSO indicators using the consolidated set of census, geographic, and state-level features. This task constitutes a multi-output regression problem. A significant methodological challenge arises from the fact that the dimensionality of the target variable space (the NSSO indicators) is considerably greater than that of the input feature space. It is well-established that the performance of such models tends to degrade as the number of target variables increases, particularly when outputs substantially outnumber inputs. This imbalance can lead to reduced predictive accuracy, lower computational efficiency, and increased prediction variance. To mitigate these issues, our approach first involves reducing the dimensionality of the target NSSO indicators themselves, for which we employ an autoencoder to learn a compact, low-dimensional representation.\par

This section outlines the proposed methodology, beginning with a formal definition of the data sources. The target NSSO data is organized into six thematic categories: Consumer Expenditure, Employment, Agriculture, Housing Conditions, Land and Livestock Holdings, and Debt and Investment. The data vector for any given category $x$ is denoted as $\mathbf{X}_{nsso}^x \in \mathbb{R}^{D^x_n}$, where $D^x_n$ represents its dimensionality. The predictor variables consist of census data, denoted as $\mathbf{X}_{cen} \in \mathbb{R}^{D_c}$, and geographic features, $\mathbf{X}_{geo} \in \mathbb{R}^{D_g}$.To prevent the model from learning a spurious ordinal relationship from arbitrary numerical assignments, state identifiers are one-hot encoded into categorical vectors. These are denoted as $\mathbf{X}_{st} \in \mathbb{R}^N$, where $N$ is the total number of states and union territories. The complete methodology is detailed in the subsections that follow.


\subsubsection{Autoencoder: Efficiently compressing the NSSO socioeconomic indicators} \label{sec:Autoencoder}
To preserve the thematic integrity of the data, a separate autoencoder was trained independently for each of the six NSSO categories. This approach ensures that the resulting latent variables are category-specific, preventing the conflation of features from disparate domains and thereby enhancing the interpretability of the learned low-dimensional representations.\par

We begin by formally defining the data at our two primary spatial scales: the district and the village cluster. At the district level, the available data for each of the six NSSO categories is denoted by $\mathbf{X}^{dist,x}_{nsso}$. The corresponding predictor variables at this scale are the census data, $\mathbf{X}_{cen}^{dist}$, and geographic data, $\mathbf{X}_{geo}^{dist}$. At the finer cluster level, only the predictor variables are available, denoted as $\mathbf{X}_{cen}^{hex}$ and $\mathbf{X}_{geo}^{hex}$, respectively.

\begin{figure}[!htb]
    \centering
    \includegraphics[width=0.9\linewidth]{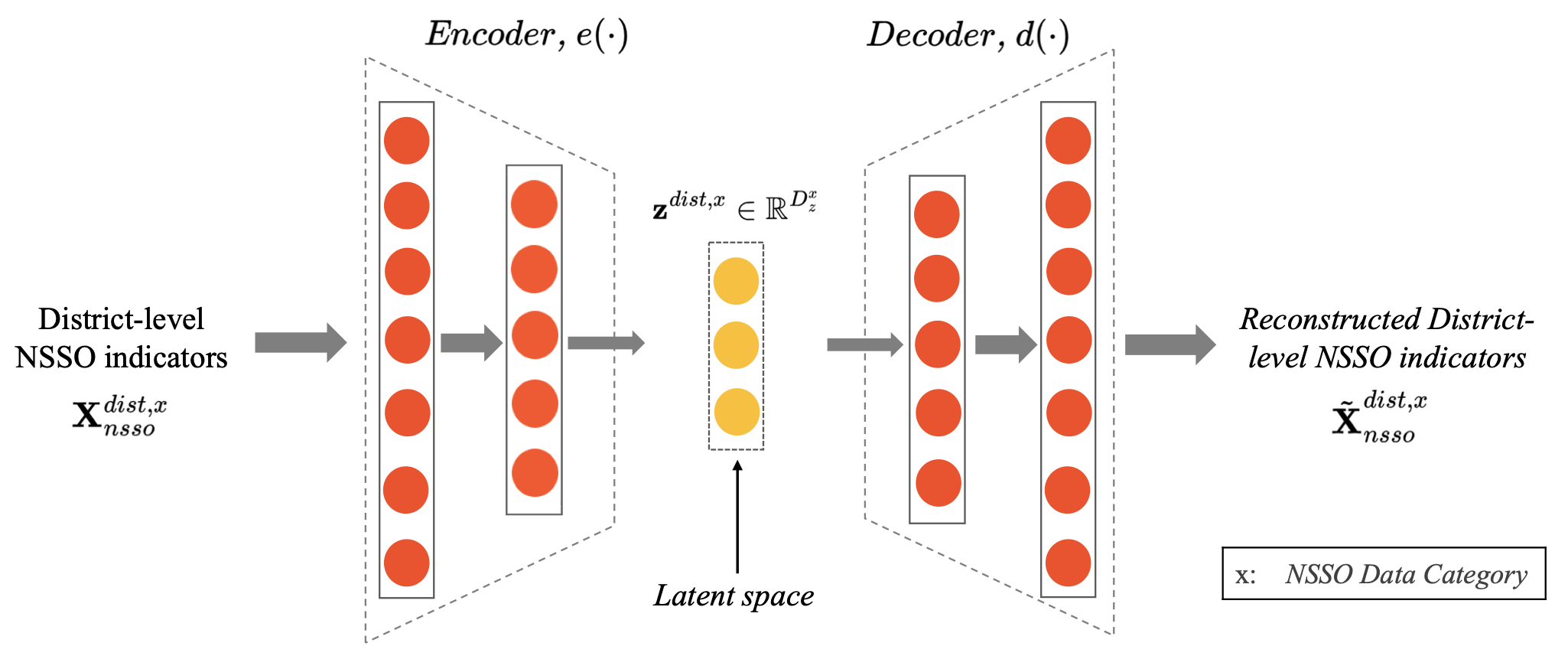}
    \caption{The autoencoder model for compressing the high-dimensional district-level NSSO features is shown.}
    \label{fig:AE}
\end{figure}

The central objective is to predict the NSSO indicators at the cluster level, for which data is unavailable. Our methodology commences by first learning a compact representation of the known district-level NSSO indicators. As illustrated in Fig. \ref{fig:AE}, a category-specific encoder, $\textbf{e}_x(\cdot)$, is employed to reduce the dimensionality of $\mathbf{X}^{dist,x}_{nsso}$. The latent space of the autoencoder is given by:\par 

\begin{equation}
\label{latent}
        \textbf{z}^{dist,x} = \textbf{e}_x\left(\textbf{X}^{dist,x}_{nsso}\right),
\end{equation}

\noindent where $x$ is the NSSO data category, $\mathbf{z}^{dist,x} \in \mathbb{R}^{D^x_z}$, $D^x_z$ is the dimensionality of the latent space, and $D^x_z < D_n^x$.

A decoder, $\textbf{d}_x(\cdot)$, is used to reconstruct the NSSO data from category $x$. The reconstructed NSSO data, $\tilde{\textbf{X}}_{nsso}^{dist,x}$, is denoted by:

\begin{equation}
    \tilde{\textbf{X}}_{nsso}^{dist,x}= \textbf{d}_x\left(\textbf{z}^{dist,x}\right),
\end{equation}

\noindent where $\tilde{\textbf{X}}_{nsso}^{dist,x}$ is the reconstructed NSSO data from the $x^{th}$ category.

\subsubsection
{Regression Model: Predicting compressed NSSO socioeconomic indicators using the census} \label{sec:CensusPredict}
The next stage of the methodology involves training a regression model to predict the compressed latent representation of the district-level NSSO data, $\textbf{z}^{dist,x}$. The input for this model is a consolidated feature vector, $\textbf{X}_{cen}^{dist}$, formed by concatenating the district-level census data, geographic features, and the one-hot encoded state identifiers:

\begin{equation}
\mathbf{X}_{inp}^{dist} = \mathbf{X}_{cen}^{dist} \oplus \mathbf{X}_{geo}^{dist} \oplus \mathbf{X}_{st},
\end{equation}

\noindent where $\oplus$ denotes the concatenation operator.

\begin{figure}[!htb]
        \centering
        {\includegraphics[width= 0.95\linewidth]{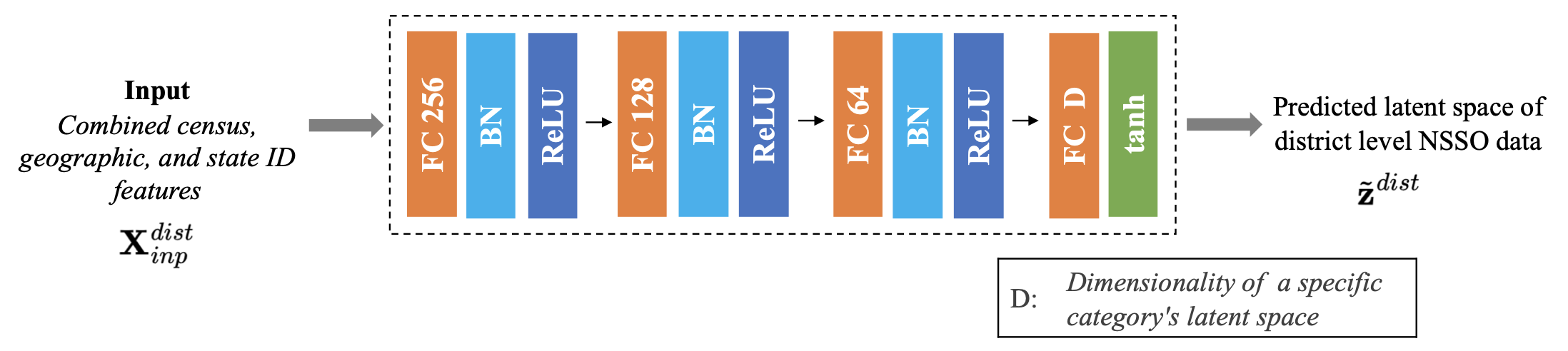} }
        \caption{The regression model for predicting the latent space of NSSO socioeconomic features is shown. The district-level census, geographical, and categorical state ID features are used as inputs to predict the latent space of district-level NSSO socioeconomic features. FC denotes fully-connected layers.}
        \label{fig:regression}
    \end{figure}

The regression model is a fully connected neural network, the architecture of which is illustrated in Fig. \ref{fig:regression}. The network consists of a series of blocks, where each block comprises a fully connected transformation, a batch normalization step, and a rectified linear unit (\textit{ReLU}) activation function. The final fully connected layer is shaped by a hyperbolic tangent (\textit{tanh}) activation function as the same activation is used for latent space in the autoencoder. The output of this network is the predicted latent space for category $x$, denoted as $\tilde{\textbf{z}}^{dist,x}$, and is formally defined in \Crefrange{eq:regression_one}{eq:regression_four}.

\begin{equation}
        \textbf{h}_{1} = \textit{ReLU}\left(BN\left(\textbf{X}_{inp}^{dist} \textbf{W}_{1} + \textbf{b}_{1} \right)\right),
    \label{eq:regression_one}
\end{equation}

\begin{equation}
        \textbf{h}_{2} = \textit{ReLU}\left(BN\left(\textbf{h}_1 \textbf{W}_{2} + \textbf{b}_{2} \right)\right),
    \label{eq:regression_two}
\end{equation}

\begin{equation}
        \textbf{h}_{3} = \textit{ReLU}\left(BN\left(\textbf{h}_2 \textbf{W}_{3} + \textbf{b}_{3} \right)\right),
    \label{eq:regression_three}
\end{equation}

\begin{equation}
        \tilde{\textbf{z}}^{dist} = tanh\left(\textbf{h}_3 \textbf{W}_{4} + \textbf{b}_{4} \right),
    \label{eq:regression_four}
\end{equation}

\noindent where $\tilde{\mathbf{z}}^{dist} \in \mathbb{R}^{D}$ is the predicted latent space of district-level NSSO data. $\textbf{W}_1$, $\textbf{W}_2$, $\textbf{W}_3$, and $\textbf{W}_4$ are the neural network weight matrices. $\textbf{b}_1$, $\textbf{b}_2$, $\textbf{b}_3$, and $\textbf{b}_4$ are the biases. $\textbf{h}_1$, $\textbf{h}_2$, and $\textbf{h}_3$ are hidden layer outputs. $BN$ is the batch normalization layer.\par

The operations in \Crefrange{eq:regression_one}{eq:regression_four} can be represented concisely using the operator, $\textbf{REG}(\cdot)$, for each category $\textit{x}$, as shown in Eq. \ref{eq:short_regressor}.

\begin{equation}
    \tilde{\textbf{z}}^{dist,x} = \textit{REG}_x\left(\textbf{X}_{inp}^{dist}\right).
    \label{eq:short_regressor}
\end{equation}


The predicted NSSO latent space of category $x$ can be reconstructed using the decoder, $\textbf{d}_x(\cdot)$. This concludes the training phase of the models.

\subsubsection{Loss Function} 
The Huber loss function was selected for training both the autoencoder and the regression models. This choice is motivated by the function's inherent robustness to outliers, a common characteristic of socioeconomic data. The Huber loss effectively combines the desirable properties of the mean squared error (MSE) for small errors with the lower outlier sensitivity of the mean absolute error (MAE) for large errors. This makes it particularly well-suited for modeling real-world datasets that often contain noise and anomalous values. The function is formally defined as:

\begin{equation}
\mathcal{L}_{\delta} = 
\begin{cases}
\frac{1}{2} \left\| \mathbf{\tilde{y}} - \mathbf{y} \right\|_2^2, & \text{if } \left\| \mathbf{\tilde{y}} - \mathbf{y} \right\| \leq \delta, \\
\delta \left( \left\| \mathbf{\tilde{y}} - \mathbf{y} \right\| - \frac{1}{2} \delta \right), & \text{otherwise},
\end{cases}
\end{equation}
\noindent 

wherein $y$ denotes the observed value, $\tilde{y}$ is the predicted value, and the hyperparameter $\delta$ modulates the function's robustness to outliers by defining the transition point between a mean-squared-error and a mean-absolute-error-based penalty.

For autoencoder training phase, $y = \mathbf{X}^{\text{dist},x}_{nsso}$, and $\tilde{y} = \tilde{\mathbf{X}}^{\text{dist},x}_{nsso}$. For the regression model training phase, $y = \mathbf{z}^{\text{dist},x}$, and $\tilde{y} = \tilde{\mathbf{z}}^{\text{dist},x}$.



\subsubsection{Inference: Predicting the high-resolution NSSO indicators} 
\label{sec:Decode}
During inference, the model is deployed to predict the NSSO indicators at the cluster level. Accordingly, the input to the regression network is formed by combining the downscaled census features, geographic attributes, and state identifiers as follows:
\begin{equation}
    \mathbf{X}_{inp}^{hex} = \mathbf{X}_{cen}^{hex} \oplus \mathbf{X}_{geo}^{hex} \oplus \mathbf{X}_{st}.
\end{equation}

This cluster-level input vector, $\textbf{X}_{inp}^{hex}$, is then passed through the trained regression network to predict the corresponding latent space for each NSSO category $x$, denoted as $\textbf{z}^{hex,x}$. This prediction is formulated as:

\begin{equation}
    \label{eq:inference_hex}
    \tilde{\textbf{z}}^{hex,x} = \textit{REG}_x\left(\textbf{X}_{inp}^{hex}\right).
\end{equation}

The predicted latent vectors are then transformed back to the original feature space via the decoder,  $\textbf{d}(\cdot)$. This reconstruction step produces the final estimate of the cluster-level NSSO indicators, formulated as:
 
\begin{equation}
    \tilde{\textbf{X}}_{nsso}^{hex,x} = \textbf{d}_x \left(\tilde{\mathbf{z}}^{hex,x} \right),
\end{equation}

\noindent where $\tilde{\textbf{X}}_{nsso}^{hex,x}$ is the cluster-level NSSO data from $x^{th}$ category. To qualitatively validate the downscaled predictions, the resulting cluster-level NSSO indicators were mapped and visually compared against the original district-level data. This comparison allows for an assessment of whether the model preserves the broad spatial patterns present in the source data while introducing plausible fine-scale variations. \par 

\subsubsection{Evaluation}
A quantitative validation was performed to assess model accuracy. To facilitate a direct comparison with the source data, the cluster-level predictions were first aggregated to the district scale via averaging. The mean squared error (MSE) was then computed between these aggregated district-level predictions and the actual district-level NSSO indicators to measure overall performance.

\subsubsection{Uncertainty Estimation}
To produce robust predictions and estimate epistemic uncertainty, an ensemble of models with identical architectures was trained. Each model was initialized with different random weights, resulting in a distribution of predictions for each output indicator. The final prediction is the mean of this distribution, while its standard deviation provides a direct measure of model uncertainty. This uncertainty metric is crucial for end-users to gauge the reliability of any given prediction.


\section{Discussion}
\label{sec:Discussion}

\textit{Principal Findings and Methodological Contribution.}
This study demonstrates the efficacy of an autoencoder-based framework for capturing and reconstructing complex socioeconomic patterns from coarse-grained survey data. The central contribution of our methodology is its ability to learn a compact, latent representation of diverse socioeconomic indicators at coarse scales and subsequently decode this representation to generate meaningful, high-resolution predictions at a finer tesselation. While the latent space itself is not directly interpretable, each dimension encodes a nonlinear combination of input features, enabling the generation of decoded maps that reflect key spatial and temporal patterns consistent with observations.\par

The most significant outcome of this spatial downscaling method is the model's capacity to reproduce not only broad regional trends but also plausible fine-scale variations within districts. Despite being trained on a single data point per district, the model's predictions exhibit significant intra-district heterogeneity, revealing subtle socioeconomic gradients that are obscured in the original coarse-scale data. This capacity for spatial decomposition allows for a more nuanced analysis of local socioeconomic dynamics than was previously possible.\par

\textit{Temporal Dynamics and Thematic Exploration.}
The utility of this approach is further underscored by temporal comparisons between the 2001 and 2011 datasets. For instance, the predicted maps of coffee consumption illustrate a clear geographic expansion of the practice from its traditional concentration in southern states like Karnataka and Kerala to broader regions of the country over the decade, reflecting a documented shift in consumer behavior. Similarly, observed increases in the use of tap water and tubewells as primary drinking sources, particularly in states like Uttar Pradesh, suggest patterns of infrastructure development and public health interventions. While caution must be exercised in causal interpretation, these temporal shifts provide valuable insights into regional progress and socioeconomic change. These examples represent only a fraction of the analytical possibilities; the rich variety of indicators within the NSSO data can be similarly explored to study topics ranging from regional disparities to evolving patterns in living conditions.\par

\textit{Uncertainty Quantification and Data Dissemination.}
In addition to the high-resolution prediction maps, our framework provides corresponding uncertainty estimates to support more informed interpretation. By employing an ensemble of models, we quantify the standard deviation across their predictions, which serves as a proxy for epistemic uncertainty arising from limitations in the model and data. This information enables end-users to assess the reliability of specific predictions and make decisions with an explicit understanding of the model's confidence. To facilitate further research and application, a cleaned version of the original district-level NSSO dataset and the associated geographic data will be made publicly available alongside our predictions.\par

\textit{Limitations.}
Several limitations of the present study must be acknowledged. First, the training dataset is constrained to a single sample per district, which limits the model's learning capacity and increases the risk of overfitting. Second, a potential distributional shift may exist between the training and inference phases; the decoder is trained exclusively on latent vectors derived from district-level data but is tasked during inference with decoding representations aggregated from hexagon-level features. The statistical properties of these two sets of latent vectors may differ, posing a generalization challenge. Third, the imputation procedure used to address missing values may have introduced noise. Finally, the source survey data itself may contain inherent biases or inaccuracies from the collection process, which would propagate through the modeling pipeline.

\textit{Conclusion and Future Directions.}
In summary, this study introduces a flexible and broadly applicable framework for modeling and decoding latent socioeconomic structures from sparse, coarse-grained data into fine-grained spatial predictions. Despite the noted limitations, the approach demonstrates how regional modeling, temporal tracking, and indicator-specific analysis can be successfully performed even in data-scarce environments. The findings point to the broader utility of representation learning in enhancing the accessibility and analytical value of socioeconomic datasets across spatial and temporal scales. \par

The framework can be readily extended to other geographical contexts where similar data structures exist. Moreover, the learned latent features hold potential as inputs for a range of downstream tasks, such as classifying regions by poverty level, clustering for policy targeting, or as auxiliary inputs to hybrid models that also incorporate satellite imagery. Ultimately, this form of representation learning offers researchers and policymakers a powerful tool to explore and interpret socioeconomic data at a finer resolution than is directly observable, unlocking new insights into the dynamics of human development. \par

\section{Conclusion}
\label{sec:Conclusion}

This study presents a deep learning framework, JuGAAD, for generating high-resolution socioeconomic maps from limited census survey data. Using features derived from Census records, geographic attributes, and state identifiers, the model predicts district-level indicators from the National Sample Survey Office (NSSO). During inference, the model enables fine-grained predictions, capturing variations even within districts and facilitating more sophisticated and detailed socioeconomic analysis. \par

The high-resolution socioeconomic indicators preserve broad spatial trends while revealing localized variations that are not captured at the district level. Temporal comparisons between 2001 and 2011 further highlight meaningful changes in key indicators, such as improvements in drinking water access and reductions in food insecurity. To assess prediction confidence, an ensemble of regression models is used. The standard deviation of their outputs is used to estimate epistemic uncertainty. This uncertainty information allows users to identify which outputs are more reliable and to interpret prediction confidence spatially. \par

The final dataset includes a cleaned version of district-level (or coarse) indicators, predicted high-resolution indicators, and uncertainty estimates. By transforming sparse and irregular data into detailed spatial representations, this work provides a foundation for regional analysis, policy evaluation, and further research in socioeconomic development, particularly in data-constrained contexts. \par

\textbf{Acknowledgments}\par
This study was supported by funding from NASA South/Southeast Asia Research Initiative grants NNX17AK75G, and 80NSSC22K1363.

\begin{appendices}

\section{NSSO Gap-filling of Indicators}\label{secA2}
The gap-filling procedure is summarised briefly in Section \ref{sec:DataPrep}. This appendix reports the performance of the imputation models across all six NSSO thematic categories.

Table~\ref{tab:gap_filling_summary} reports, for each category, the total number of variables, the number of districts with missing observations, and the mean $R^{2}$ and MSE of the accepted imputation models. Across all categories, the mean $R^{2}$ exceeds $0.94$, indicating that the cross-category predictor structure in the NSSO surveys provides sufficient information to reliably recover missing values. Only two variables, one in Agriculture and one in Housing Conditions, could not be imputed with sufficient accuracy and were removed. \par

The missing data is not uniformly distributed across categories. Housing Conditions and Employment exhibit the highest gap fractions ($0.162$ and $0.112$, respectively), indicating that these surveys had substantially lower geographic coverage, likely reflecting the administrative challenges of collecting labor and dwelling condition data across all districts. The remaining categories, Consumer Expenditure, Debt and Investment, and Land and Livestock 
Holdings show near-complete coverage, with gap fractions below  $0.003$.

\begin{table*}[!htb]
    \centering
    \caption{Summary of NSSO gap-filling performance by category. Missing Districts is the maximum number of district-level observations requiring imputation across indicators in the category. Gap Fraction is the corresponding proportion of all districts. Mean $R^{2}$ and Mean MSE are averaged over all indicators.}
    \resizebox{\linewidth}{!}{
    \begin{tabular}{lrrrcc}
        \toprule
        \textbf{Category} & \textbf{Number of Indicators} & 
        \shortstack{\textbf{Missing} \\ \textbf{Districts}} & 
        \shortstack{\textbf{Gap} \\ \textbf{Fraction}} & 
        \shortstack{\textbf{Mean CV} \\ \boldmath{$R^{2}$}} & 
        \textbf{Mean MSE} \\
        \midrule
        Agriculture        & 76  & 14  & 0.022 & 0.9452 & 0.000124 \\
        Consumer           & 110 & 2   & 0.003 & 0.9629 & 0.000665 \\
        Debt               & 83  & 1   & 0.002 & 0.9899 & 0.000159 \\
        Employment         & 92  & 72  & 0.112 & 0.9843 & 0.000191 \\
        Housing Conditions & 69  & 104 & 0.162 & 0.9686 & 0.000220 \\
        Land               & 45  & 1   & 0.002 & 0.9887 & 0.000156 \\
        \bottomrule
    \end{tabular}}
    \label{tab:gap_filling_summary}
\end{table*}

\begin{figure}[!htb]
    \centering
    \includegraphics[width=0.28\linewidth]{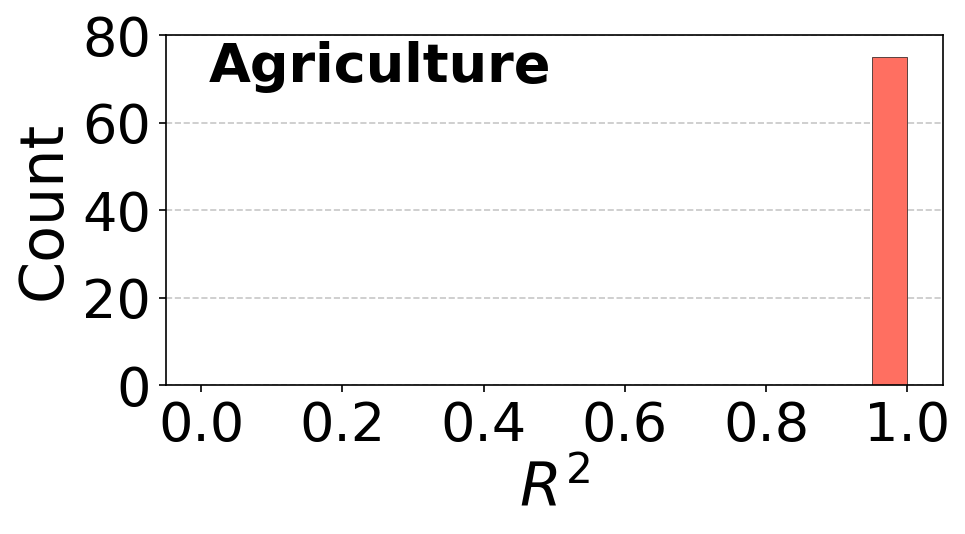}
    \includegraphics[width=0.28\linewidth]{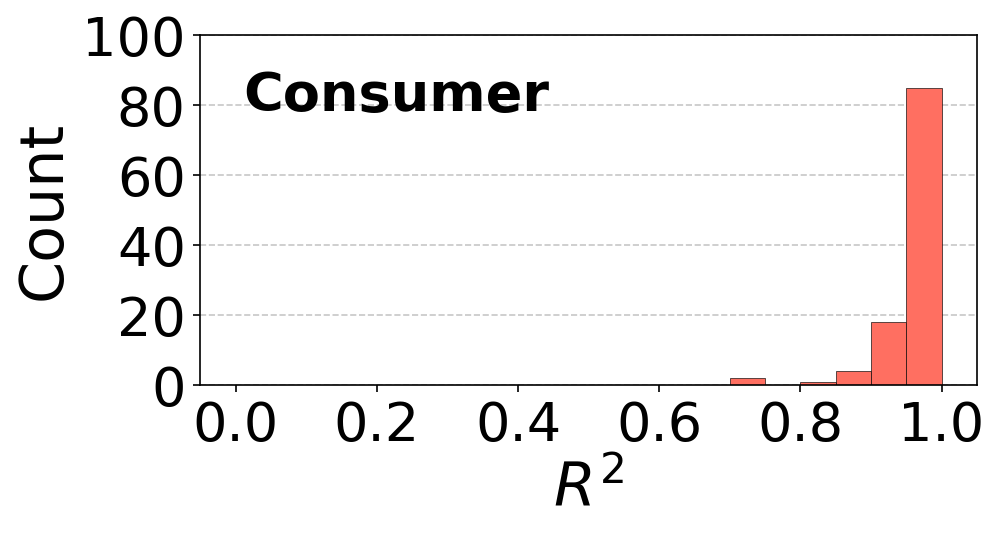}
    \includegraphics[width=0.28\linewidth]{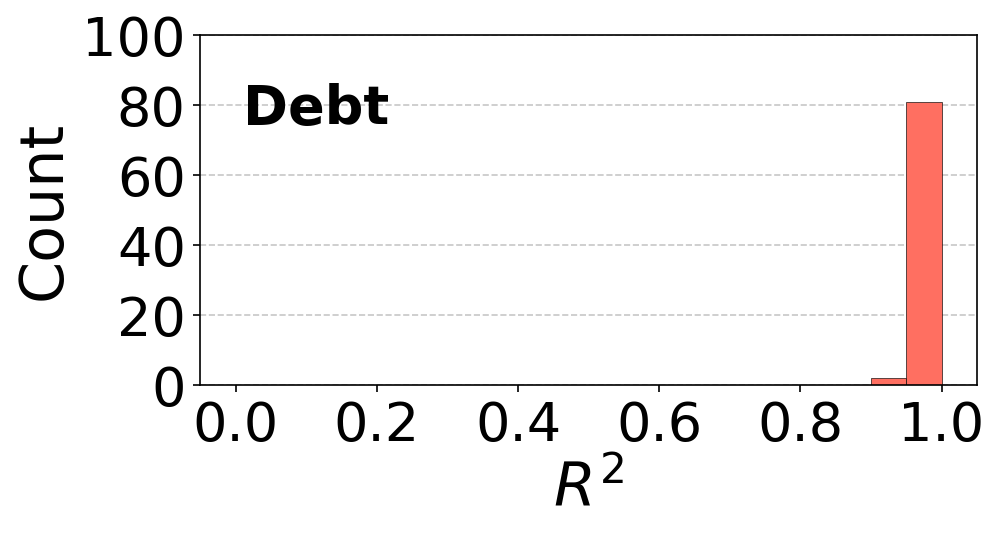}\\[6pt]
    \includegraphics[width=0.28\linewidth]{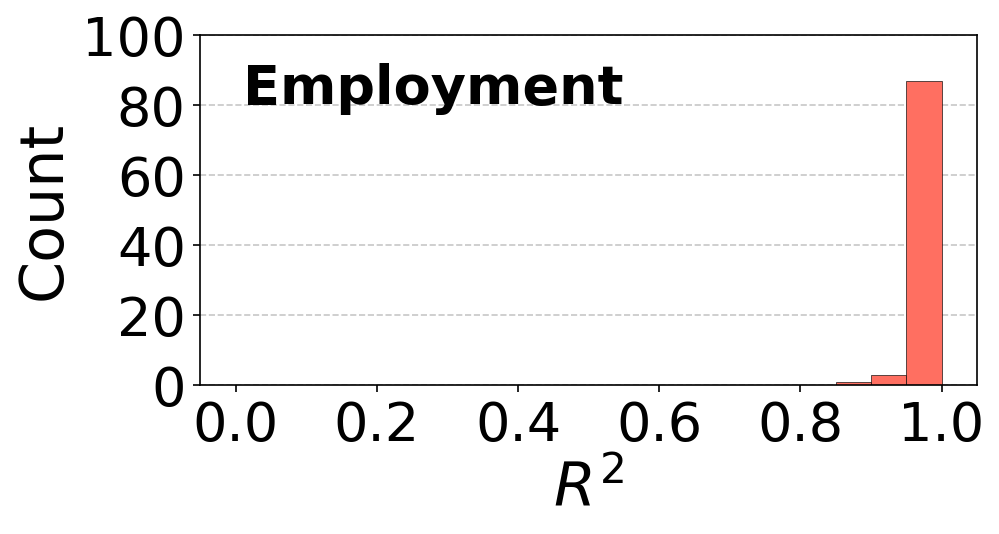}
    \includegraphics[width=0.28\linewidth]{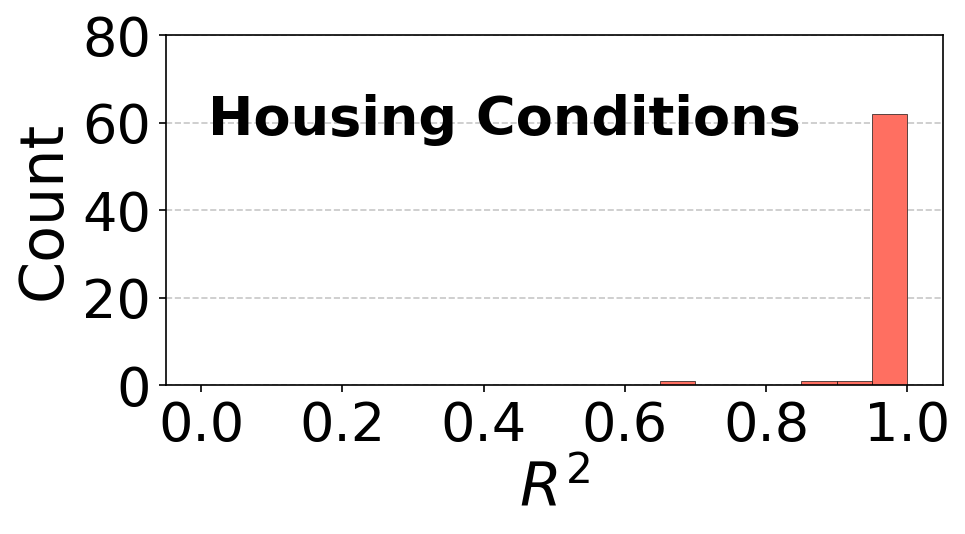}
    \includegraphics[width=0.28\linewidth]{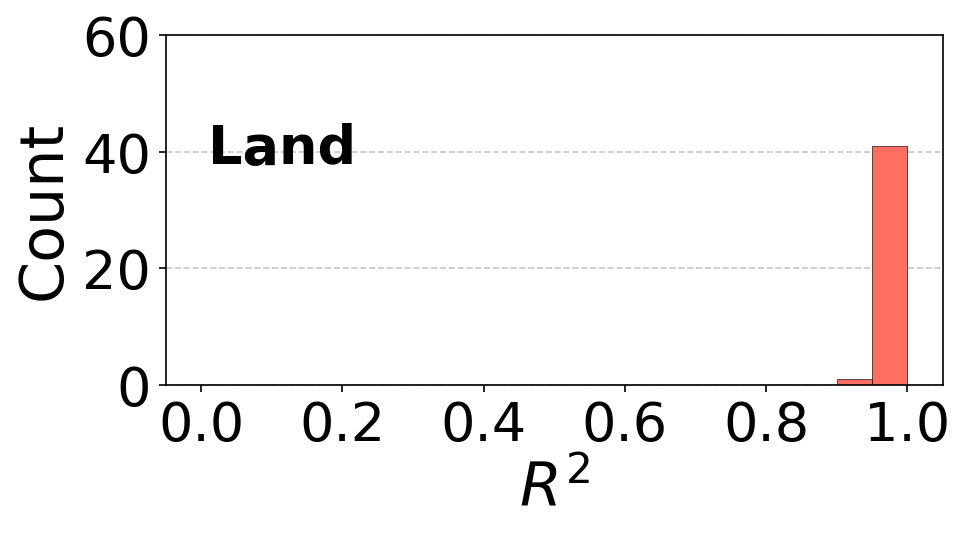}
    \caption{Distribution of cross-validated $R^{2}$ values for the 
    gap-filling imputation models, shown separately for each NSSO 
    thematic category. Each bar represents the number of variables 
    whose imputation model achieved the corresponding $R^{2}$ value. 
    The heavy concentration near 1.0 across all categories indicates 
    strong and reliable imputation performance.}
    \label{fig:gap_filling_r2}
\end{figure}
Figure~\ref{fig:gap_filling_r2} shows the distribution of 
cross-validated $R^{2}$ values for each category. In all cases, the vast majority of variables achieve $R^{2} > 0.95$, with the distribution heavily concentrated near 1.0. Consumer Expenditure shows the widest spread, with some variables in the 0.7--0.9 range, reflecting the greater heterogeneity across districts in expenditure patterns. Employment, Land, and Debt and Investment show the most concentrated distributions, with nearly all variables exceeding $R^{2} = 0.94$.

\end{appendices}

\bibliography{References}

\end{document}